\documentclass{article}

\usepackage[final, nonatbib]{neurips_2023}

\usepackage[utf8]{inputenc} 
\usepackage[T1]{fontenc}    
\usepackage{hyperref}       
\usepackage{url}            
\usepackage{amsfonts}       
\usepackage{nicefrac}       
\usepackage{microtype}      
\usepackage[dvipsnames]{xcolor}

\usepackage{caption}
\usepackage{soul}
\usepackage{graphicx}
\usepackage{booktabs}
\usepackage{pifont}
\usepackage{enumitem}
\usepackage{multirow}
\usepackage{wrapfig}

\title{HOH: Markerless Multimodal Human-Object-Human Handover Dataset with Large Object Count}

\author{%
  Noah Wiederhold \\
  Clarkson University\\
  \texttt{wiedern@clarkson.edu} \\
  \And
  Ava Megyeri \\
  Clarkson University\\
  \texttt{megyeram@clarkson.edu} \\
  \And
  DiMaggio Paris \\
  Clarkson University\\
  \texttt{parisda@clarkson.edu} \\
  \AND
  Sean Banerjee \\
  Clarkson University\\
  \texttt{sbanerje@clarkson.edu} \\
  \And
  Natasha Kholgade Banerjee \\
  Clarkson University\\
  \texttt{nbanerje@clarkson.edu} \\
  }

\begin{document}

\maketitle

\begin{figure}[h!]
    \centering
    \includegraphics[width=\linewidth]{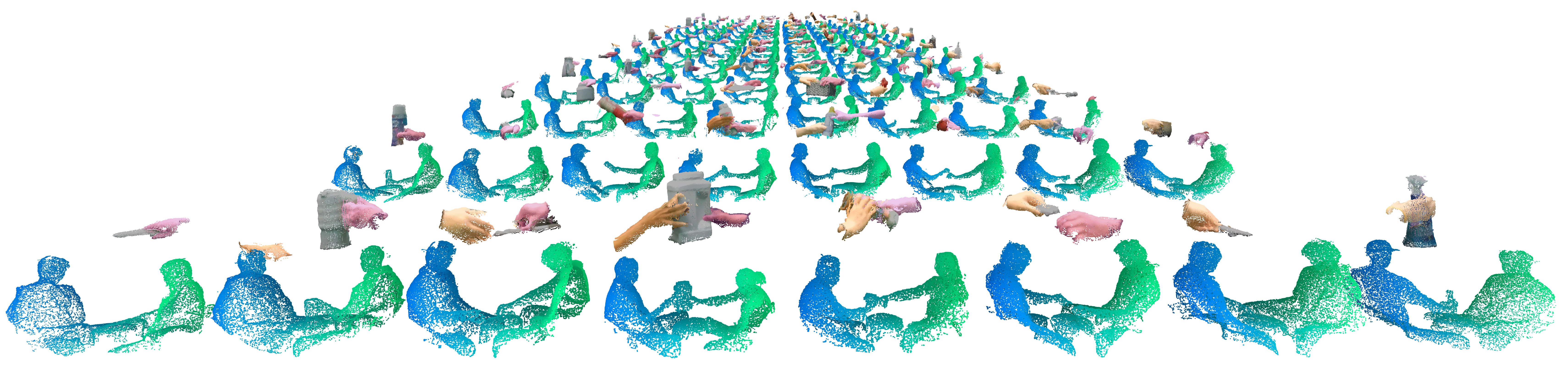}
    \caption{HOH is a markerless 3D multimodal dataset on human-human handovers with 136 objects and 20 participant pairs, 40 accounting for role-reversal. We show 3D point clouds of the upper body, giver (pink) and receiver (yellow) hands, and objects (colorized) fused from 4 Kinects from various time points of handover and across various participant pairs and objects. The dataset demonstrates diversity of object geometry, participant posture, interaction approaches, and grasp types. Shown are right-seated givers. The dataset also consists of left-seated givers.}
    \label{fig:teaser}
\end{figure}

\begin{abstract}

We present the \textbf{HOH} (\textbf{H}uman-\textbf{O}bject-\textbf{H}uman) Handover Dataset, a large object count dataset with 136 objects, to accelerate data-driven research on handover studies, human-robot handover implementation, and artificial intelligence (AI) on handover parameter estimation from 2D and 3D data of two-person interactions. HOH contains multi-view RGB and depth data, skeletons, fused point clouds, grasp type and handedness labels, object, giver hand, and receiver hand 2D and 3D segmentations, giver and receiver comfort ratings, and paired object metadata and aligned 3D models for 2,720 handover interactions spanning 136 objects and 20 giver-receiver pairs\textemdash{}40 with role-reversal\textemdash{}organized from 40 participants. We also show experimental results of neural networks trained using HOH to perform grasp, orientation, and trajectory prediction. As the only fully markerless handover capture dataset, HOH represents natural human-human handover interactions, overcoming challenges with markered datasets that require specific suiting for body tracking, and lack high-resolution hand tracking. To date, HOH is the largest handover dataset in terms of object count, participant count, pairs with role reversal accounted for, and total interactions captured.

\end{abstract}

\section{Introduction}
\label{sec:intro}

Human-human handover of objects is a complex process that has been highly studied due to its role in enabling fluent human-robot interaction (HRI) in collaborative operations. Researchers have investigated a wide range of parameters underlying handover, including physical factors such as grip force~\cite{mason2005grip,chan2012grip,dohring2020grip,controzzi2018humans,khanna2023multimodal}, interpersonal distance~\cite{basili2009investigating,hansen2017human}, object orientation~\cite{chan2015characterization,chan2020affordance}, object mass~\cite{hansen2017human}, hand movements~\cite{rasch2017understanding,parastegari2017modeling}, and reaction times~\cite{glasauer2010interacting}, as well as cognitive parameters such as giver/receiver intent communication~\cite{lee2011predictability}, gaze variation and joint attention for shared goals~\cite{strabala2012learning,moon2014meet,zheng2015impacts}, movement adaptation~\cite{huang2015adaptive}, and affordance preferences~\cite{cini2019choice}. The study of these parameters has given rise to a plethora of work in data-driven human-robot handover interactions~\cite{ortenzi2021object,kshirsagar2019specifying,khanna2022human}, and has shown potential for learning-based prediction of handover parameters such as receiver grasp~\cite{ye2021h2o}. Given the often non-verbal aspects of handover and its dependence on inter-personal coordination for shared goal success~\cite{kopnarski2023systematic}, there is a strong interest within the cognitive science community~\cite{kopnarski2023systematic} to understand ``What Does a Handover Tell''~\cite{bekemeier2019does}, e.g., sensitivity of motion kinematics to social intention~\cite{becchio2008case}, potential individuality of short-range handover trajectories~\cite{bekemeier2019does}, and inter-person coordination. The study of handover also has a societal benefit, of providing robots that engage in patient-centric caregiving and productivity-aware collaboration.

The large-scale propagation of human-human handover research has been hindered by two challenges. The first challenge is that, so far, human-human handover studies have used \textbf{small object counts}, typically 2-5, a few with 10-20~\cite{chan2020affordance,carfi2019multi,kshirsagar2023dataset}, and to-date no more than 30~\cite{ye2021h2o}. Given the vast diversity of objects likely to be interacted with by robots in consumer spaces\textemdash{}tools, kitchen utensils, containers, toys, fruit, bathroom items, office supplies, and electronic items to name a few, conclusions on physical and cognitive parameters in small object count studies cannot be generalized to objects in the wild, without fully studying the impact of variation in object properties such as size, shape, mass, functionality, and presence of protrusions or affordances. 

The second challenge is that publicly available datasets~\cite{chan2020affordance,carfi2019multi,khanna2023multimodal,kshirsagar2023dataset}, apart from being few and using small object counts, use \textbf{marker-based motion capture (mocap)} with only a single marker on the wrist and 1-5 markers on the object. Markered body mocap suffers from known limitations such as use of form-fitting suits that prevents clothing diversity, and lack of high-resolution hand geometry and object structure that prevents analysis of spatial affordance during grasp and object transfer~\cite{cini2019choice} to enable safe human-robot handover. Attempts have been made to use public datasets to develop human-inspired robotic controllers~\cite{kshirsagar2019specifying,faibish2022human,khanna2022human}, demonstrating their value. However, the constrained markered setup and small object count hinder their use in studying parameters of natural handover or use in developing learning algorithms for robotics that rely on large data with high degree of diversity. 

We contribute \textbf{HOH} (\textbf{H}uman-\textbf{O}bject-\textbf{H}uman), the first markerless, high object count human-human handover dataset that is publicly available. HOH contains 2,720 interactions performed by 40 participants organized in 20 role-reversing giver-receiver pairs covering a total of 136 objects\textemdash{}116 store-bought and 20 3D printed\textemdash{}spanning 17 form/function categories and 8 everyday use classes. We adopt a markerless approach to capture natural real-world motions and clothing. We use a multi-camera setup of 4 30FPS Kinect RGB-D sensors and 4 60FPS FLIR Point Grey cameras to perform 360$^\circ$ allocentric (non-body-mounted) capture of human-human handover. We record post-handover giver and receiver perceptions of comfort. We provide the following contributions in HOH.

\begin{enumerate}[noitemsep,topsep=0pt,leftmargin=*]
    \item Kinect depth video for 2,720 handovers from 20 participant pairs, and Kinect and Point Grey color video for 2,448 handovers from 18 pairs who consented to identifiable color (IC) data release.
    \item Manual ground truth (GT) annotations of key events on first giver grasp on object, object transfer, and last receiver contact, giver hand, object, and receiver hand masks assisted by Segment Anything Model (SAM)~\cite{kirillov2023segany}, and giver and receiver handedness and grasp type at object transfer using the taxonomy of Cini et al.~\cite{cini2019choice}. 
    \item Processed data in the form of full 360$^\circ$ point clouds (color mapped for IC pairs), OpenPose~\cite{cao2021openpose} skeletons, tracked hand and object masks, and color-mapped hand and object point clouds over from first to last key event frame in all 2,720 interactions. 
    \item Information on the object used in each interaction, including mass, class, category, and 3D model.
    \item Object 3D model GT alignments to frames ranging from the first key event to the last key event, providing GT 6DOF object pose. 
    \item Analysis of object, trajectory, and grasp properties.
    \item Experimental results of neural networks trained to predict giver grasp and transfer orientation using object point clouds, and receiver grasp and trajectories using object and giver data.
\end{enumerate}

\textbf{Intended Use Cases.} The dataset facilitates studies on human-human handover interaction with a broad range of objects to inform cognitive psychology and human-robot handover research. The dataset enables investigation of parameters such as giver and receiver kinematics, object motions, giver/receiver hand and upper body coordination for shared goal accomplishment, grasp type, multi-person handedness, and key event timing. Investigations can include per-parameter statistical analyses, cross-parameter relationships, and relationships with respect to factors such as object geometry, participant demographics, and subjective ratings of comfort. The benefit of 3D markerless capture for HRI is to develop learning-based algorithms that inform robotic manipulators connected with RGB-D sensors on estimating where to grasp to enable safe human-robot handover using point clouds backprojected from RGD-D data. Robot givers can learn from the behavior of human givers where to preferentially hold objects, how to move during giving, and how to orient the object at the transfer point when handing the object to a receiver. Robot receivers can use the behavior of human receivers to learn where to grasp objects handed to them by a giver, and what trajectory to navigate in space to remain safe. By containing full upper-body capture and associated skeleton estimation, the dataset enables the study of relationships such as hand and body articulation, and head pose and eye gaze at various key points of the handover. 
These analyses are motivated by prior human-robot handover implementations~\cite{huang2015adaptive,kshirsagar2019specifying,khanna2022human,moon2014meet,zheng2015impacts,rasch2017understanding,parastegari2017modeling} and cognitive studies~\cite{becchio2008case,bekemeier2019does} based upon similar human-human handover studies on smaller object sets. Study of simultaneous giver and receiver comfort perceptions is expected to enable understanding of alignment between the giver and receiver, facilitating study of inter-person coordination important to shared goal accomplishment~\cite{kovacs2020accessing}. 

\section{Related Work}

\paragraph{Human-Human Handover Datasets.} To the best of our knowledge, 4 publicly available human-human handover datasets exist at this time~\cite{carfi2019multi,chan2020affordance,khanna2023multimodal,kshirsagar2023dataset}. As discussed in Section~\ref{sec:intro}, they are hindered by the limited object quantity, and low-resolution data and unnatural interaction setting due to use of markered mocap. Khanna et al.~\cite{khanna2023multimodal} focus on grip force measurement, and use a single unweighted and weighted force-sensor instrumented baton to gather mocap data. Their data prevents understanding handover dependence on object geometry. Carfi et al.~\cite{carfi2019multi} and Kshirsagar et al.~\cite{kshirsagar2023dataset} provide RGB-D data from 1 and 2 views respectively. The Kinect FOV makes fewer than 4 views insufficient for full 360$^\circ$ coverage, e.g., to capture hand-object surfaces occluded from the camera view. Though Chan et al.~\cite{chan2020affordance} provide raw color data from their Vicon cameras, they use 8 views, lack depth data, use a wider-out setup, and have participants wear black body suits. The low view count, lack of texture, and high distance is unlikely to yield success in using multi-view stereo for 3D reconstruction. Their color data is in a proprietary Vicon format requiring a license purchase.

We have identified two more multi-object dataset collections~\cite{cini2019choice,ye2021h2o} that are not publicly available. Cini et al.~\cite{cini2019choice} perform grasp taxonomy analysis by recording mocap using IR reflectors on the hand dorsum and object, with one video per interaction. Their data suffers from similar concerns as the public datasets. Ye et al.~\cite{ye2021h2o} acquire mocap using 6 magnetic markers per hand (5 at fingertips and one at wrist) and 3 optical markers per object, as well as RGB-D video using 5 30FPS allocentric cameras. They provide deep networks to hand-object pose estimation and giver-conditioned receiver pose prediction. The 5-fingertip capture provides higher detail than other datasets, however, knuckle articulation is absent. Though the RGB-D cameras can provide dense detail, the magnetic sensor cables occlude the hand dorsum. The RGB-D views lack face and upper body data, important to analyze gaze~\cite{moon2014meet,faibish2022human} and arm extension. To ensure comprehensive natural grasp and structure capture, we use a markerless allocentric setup with 360$^\circ$ coverage of the upper body of both participants. 

We summarize key properties of multi-object datasets in comparison to HOH in Table~\ref{tab:comparison}. HOH is the largest multi-object dataset in terms of object counts, participants recruited, pairs with role-reversal (RR) accounted for (matching the non-public H2O dataset), and number of cameras (matching the count in Chan et al.~\cite{chan2020affordance}). Our object count at 136 is 6.8$\times$ the count of Chan et al., the public dataset with the next largest count. 10 participant pairs (20 with RR) in HOH interact with 68 objects and the other 10 pairs (20 with RR) interact with a separate set of 68 objects, resulting in multi-pair interactions for the same object set with 3.4$\times$ the object count in Chan et al. HOH is the only dataset with simultaneous giver and receiver comfort ratings.

\setlength{\tabcolsep}{4pt}
\begin{table}[t!]
    \caption{Comparison of HOH (last column) versus prior multi-object human-human handover datasets. RR = role reversal, i.e., giver becomes receiver and vice versa, and is only applicable for some datasets. SB = Single-blind ($^\star$experimenter is one participant), DB = Double-blind (both participants are recruited). $^\dag$from Vicon X2D. **Obtained by placing markers on object, rather than by model alignment.}
    \centering
    {\footnotesize\begin{tabular}{|l|c|c|c|c|c|c|c|c|}
    \hline        
        & Cini~\cite{cini2019choice} &  Chan~\cite{chan2020affordance} & \multicolumn{2}{c|}{Carfi~\cite{carfi2019multi}} & \multicolumn{2}{c|}{Kshirsagar~\cite{kshirsagar2023dataset}} & H2O~\cite{ye2021h2o} & HOH\\
        \hline
         &  &  & SB & DB & Biman. & Uniman. &  & (Ours) \\
         \hline
         \hline
        \# Interactions & 1,734 & 1,200 & 799 & 288 & 240 & 120 & 1,200 & \textbf{2,720}\\
        \hline
        \# Objects & 17 & 20 & 3 & 7 & 10 & 5 & 30 & \textbf{136}\\
        \hline
        \# Participants & 34 & 20 & 18 & 18 & 24 & 24 & 15 & \textbf{40}\\
        \hline
        \# Pairs & 17 & 10 & 18$^\star$ & 9 & 12 & 12 & \textbf{40} & 20\\
        \hline
        \# Pairs with RR & 17 & 10 & 36$^\star$ & 18 & 24 & 24 & \textbf{40} & \textbf{40}\\
        \hline
        \# Cameras & 1 & \textbf{8} & 1 & 1 & 2 & 2 & 5 & \textbf{8}\\
        \hline
        \hline
        Markerless? & \textcolor{red}{$\times$} & \textcolor{red}{$\times$} & \textcolor{red}{$\times$} & \textcolor{red}{$\times$} & \textcolor{red}{$\times$} & \textcolor{red}{$\times$} & \textcolor{red}{$\times$} & \textcolor{LimeGreen}{$\checkmark$}\\
        \hline
        Ratings? & \textcolor{red}{$\times$} & \textcolor{red}{$\times$} & \textcolor{LimeGreen}{$\checkmark$} & \textcolor{red}{$\times$} & \textcolor{red}{$\times$} & \textcolor{red}{$\times$} & \textcolor{red}{$\times$} & \textcolor{LimeGreen}{$\checkmark$}\\
        \hline
        Color? & \textcolor{LimeGreen}{$\checkmark$} & \textcolor{LimeGreen}{$\checkmark$}$^\dag$ & \textcolor{LimeGreen}{$\checkmark$} & \textcolor{LimeGreen}{$\checkmark$} & \textcolor{LimeGreen}{$\checkmark$} & \textcolor{LimeGreen}{$\checkmark$} & \textcolor{LimeGreen}{$\checkmark$} & \textcolor{LimeGreen}{$\checkmark$}\\
        \hline
        Depth? & \textcolor{red}{$\times$} & \textcolor{red}{$\times$} & \textcolor{red}{$\times$} & \textcolor{red}{$\times$} & \textcolor{LimeGreen}{$\checkmark$} & \textcolor{LimeGreen}{$\checkmark$} & \textcolor{LimeGreen}{$\checkmark$} & \textcolor{LimeGreen}{$\checkmark$}\\
        \hline
        Point Clouds? & \textcolor{red}{$\times$} & \textcolor{red}{$\times$} & \textcolor{red}{$\times$} & \textcolor{red}{$\times$} & \textcolor{red}{$\times$} & \textcolor{red}{$\times$} & \textcolor{red}{$\times$} & \textcolor{LimeGreen}{$\checkmark$}\\
        \hline
        3D Object Model? & \textcolor{red}{$\times$} & \textcolor{red}{$\times$} & \textcolor{red}{$\times$} & \textcolor{red}{$\times$} & \textcolor{red}{$\times$} & \textcolor{red}{$\times$} & \textcolor{LimeGreen}{$\checkmark$} & \textcolor{LimeGreen}{$\checkmark$}\\        \hline
        6DOF Object Pose? & \textcolor{red}{$\times$} & \textcolor{LimeGreen}{$\checkmark$}** & \textcolor{red}{$\times$} & \textcolor{red}{$\times$} & \textcolor{LimeGreen}{$\checkmark$}** & \textcolor{LimeGreen}{$\checkmark$}** & \textcolor{LimeGreen}{$\checkmark$}** & \textcolor{LimeGreen}{$\checkmark$}\\
        \hline
        Public? & \textcolor{LimeGreen}{$\checkmark$} & \textcolor{LimeGreen}{$\checkmark$} & \textcolor{LimeGreen}{$\checkmark$} & \textcolor{LimeGreen}{$\checkmark$} & \textcolor{LimeGreen}{$\checkmark$} & \textcolor{LimeGreen}{$\checkmark$} & \textcolor{red}{$\times$} & \textcolor{LimeGreen}{$\checkmark$}\\
        \hline        
    \end{tabular}}
    \label{tab:comparison}
\end{table}

\paragraph{Hand-Object Datasets.} In consisting of grasp-based hand-object interactions, human-human handover shares features with the large body of recent work in hand-object interaction, focused on, e.g., AI-driven hand and object pose estimation in the presence of hand-object occlusions, and even locations for safe robotic grasp in human-to-robot handover~\cite{chao2021dexycb}. Much recent work has been invested in the collection of \textbf{single-person hand-object interaction datasets}~\cite{zimmermann2019freihand,brahmbhatt2020contactpose,taheri2020grab,chao2021dexycb,bhatnagar2022behave,huang2022intercap,liu20224d,hampali2020honnotate,kwon2021h2o,hampali2022keypoint,sener2022assembly101,fan2023arctic}. 
AI algorithms and research findings from single-person hand-object interaction datasets, even when bimanual, cannot be directly transferred to handover interactions involving \textit{two different people}. Inter-participant differences introduce variations in hand geometry and appearance. Spatial kinematics vary across the two types of interactions. Handovers may cover larger translations during reach and transfer, making egocentric setups~\cite{kwon2021h2o,liu20224d} infeasible due to the constrained view field. Single-person bimanual interactions involve mirrored right and left hands. The preponderance of right-handed individuals means that handovers with unimanual interactions are likely to be dominated by non-mirrored right-right hand interactions at the giver-to-receiver object transfer phase. These concerns justify the need for a distinct handover dataset such as HOH, in which over 75\% of interactions occur using non-mirrored hands. Several interactions in HOH have 3+ hands, e.g., unimanual giver grasp to bimanual receiver grasp. HOH exceeds all but the egocentric-only HOI4D dataset~\cite{liu20224d} in object count. Though Assembly101, with 101 objects, approaches HOH, Assembly101 lacks paired and aligned 3D scans. 

\section{HOH Dataset}
\label{sec:datasetcollection}
\subsection{Dataset Design Philosophy} 

Motivated by the goal of 360$^\circ$ capture of natural handover, we used a markerless allocentric capture approach and recorded giver-to-receiver handover in a seated pose to minimize fatigue. Seated handover interactions are common in social settings, e.g., at a coffee shop or in a collaborative work environments.
 We employ the standardized nomenclature of Kopnarski et al.~\cite{kopnarski2023systematic} who define handover as consisting of four phases, \textit{reach and grasp}, where the giver applies enough grip force for object hold, \textit{transport} where the object is moved to reach the shared transfer point, \textit{object transfer} during which the object is handed over from giver to receiver, and \textit{end of handover}, during which the receiver acquires full object possession and ``uses the object in line with their intention''~\cite{kopnarski2023systematic}. In our study, end of handover involves placing the object on the table.

Some prior studies inform participants to consider intended use, or specifically request bimanual or unimanual grasp~\cite{kshirsagar2023dataset}, or giver/receiver consideration~\cite{chan2020affordance}. We provide no prompting regarding interaction, except that the giver maintain grasp during transport. Our goal of unprompted interaction was to acquire natural giver reach and grasp, similar to the \textit{natural} condition of Chan et al.~\cite{chan2020affordance}, to enable analysis of grasp type and hand combinations in open-ended handover, and to avoid compromising data diversity, a concern for learning-based algorithms. Given the unstructured nature of the interaction, a giver could be inconsiderate of the receiver's comfort, or in thinking of the receiver could compromise their own comfort. We obtained a 7-point Likert scale rating after each interaction to gauge giver and receiver comfort.

\begin{figure}
    \centering
    \includegraphics[width=\linewidth]{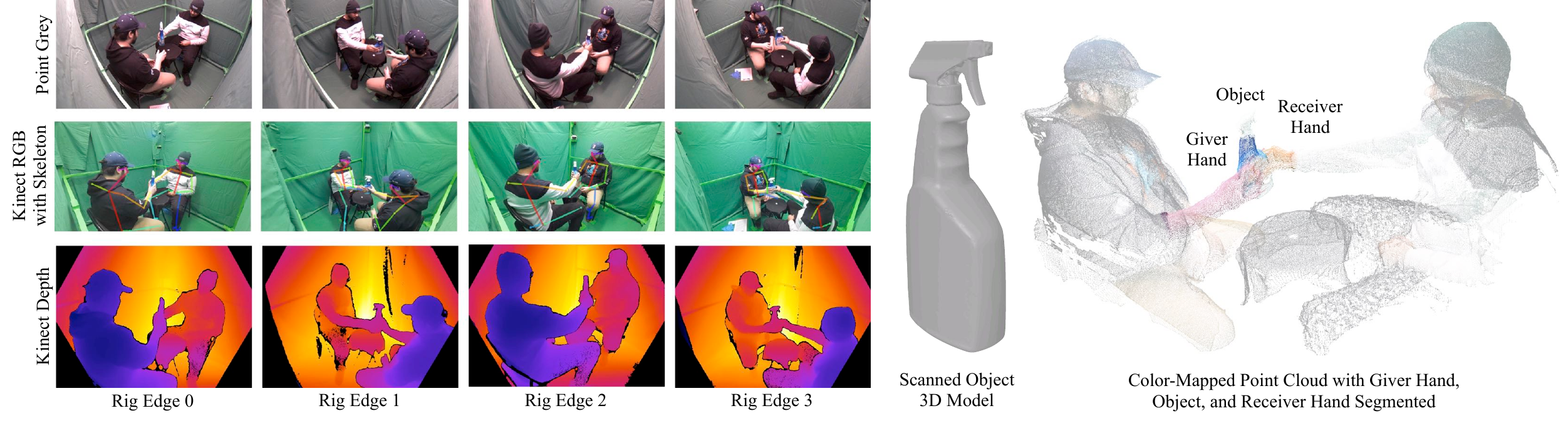}
    \caption{Left: Synchronized color and depth data collected in HOH from 4 viewpoints for a handover interaction, with OpenPose skeletons show in in Kinect color images. Center: Object 3D model used in interaction. Right: Color-mapped 3D point cloud with segmented giver, object, and receiver.}
    \label{fig:data}
\end{figure}

Our data is captured in a 1.7m $\times$ 1.7m $\times$ 2.0m green-screened T-slot frame rig using 4 Azure Kinect RGB-D sensors and 4 FLIR Point Grey BlackFly S high-speed color cameras, as shown by the images in Figure~\ref{fig:data}. The supplementary covers the setup, computing, connection, and networking setup and details to conduct simultaneous multi-sensor capture, spatial geometric camera calibration, and post-capture synchronization via an overhead light. In 160 of the 2,720 interactions (4 per pair), we requested the participants to wear blue gloves with the intention of acquiring high-fidelity tracking using color-segmentation. The Segment Anything Model (SAM)~\cite{kirillov2023segany} now facilitates tasks such as background-region removal, and foreground extraction as performed for this dataset. However, we collected HOH data prior to SAM, when reliable off-the-shelf easy-to-use segmentation tools were unavailable. We opted for a close-confined setup over spread-out to acquire color and depth data at the highest resolution feasible using Kinects and Point Grey cameras.

\subsection{Dataset Collection}

\textbf{Dataset-in-a-Nutshell.} HOH contains multimodal data on handover interactions from a total of 20 participant pairs, or 40 pairs with giver-receiver role-reversal (RR), interacting with a total of 136 objects while seated in our multi-camera capture space. 10 pairs interact with 68 and the other 10 pairs with the remaining 68. Table~\ref{tab:data} provides a summary of HOH data. We pair HOH interactions with a 136-object dataset containing 3D models and metadata.

\setlength{\tabcolsep}{3.5pt}
\newcommand{\STAB}[1]{\begin{tabular}{@{}c@{}}#1\end{tabular}}
\begin{table}[]
\caption{HOH Dataset Summary. GT = Ground Truth Annotation, Pre = pre-GT processing,  post = post-GT processing to propagate annotations. $^\star$Currently all skeleton estimation, color-mapping, segmentation, GT, and post-processing has been conducted for Kinect images.}
\centering
{\scriptsize\begin{tabular}{@{}|c|l|p{2.5in}|l|l|@{}}
\hline
& \textbf{Modality} & \textbf{Modality Description} & \textbf{Interactions} & \textbf{Count} \\
\hline
\hline
{\multirow{6}{*}{\rotatebox[origin=c]{90}{Pre}}} & Kinect Color & 4-viewpoint 30FPS RGB Video, 1920x1080 & 2,448 & 1.4M \\
 \cline{2-5}
& Kinect Depth & 4-viewpoint 30FPS Depth Video & 2,720 & 1.6M \\
\cline{2-5}
& FLIR Point Grey Color & 4-viewpoint 60FPS RGB Video & 2,448 & 2.8M \\
\cline{2-5}
& Skeletons$^\star$ & Estimated using OpenPose~\cite{cao2021openpose} in Kinect color images & 2,720 & 1.6M \\
\cline{2-5}
& Fused Point Clouds & Fused from depth images using multi-camera calibration & 2,720 & 250K \\
\cline{2-5}
 & Fused Color Point Clouds$^\star$ & Colorized using Kinect color images & 2,448 & 240K \\
\cline{2-5}
& Full Color Segmentation Masks$^\star$ & Extracted using SAM~\cite{kirillov2023segany} & 2,720 &  \\
\hline
 \hline
{\multirow{4}{*}{\rotatebox[origin=c]{90}{GT$^\star$}}} & Key Event Annotations & Frames for first giver-object contact (G),  giver/receiver grasp at transfer (T), and last receiver-object contact (R) & 2,720 & 8,160 \\
\cline{2-5}
& Giver Hand / Receiver Hand / & Manually isolated from SAM segmentation for giver hand from G & 2,720 & 8,513 / 8,879 / \\
& Object Segmentation Masks & and T, object from all G, T, R, and receiver hand from T & & 21,224 \\
& & and R, done for as many viewpoints where entity is visible  &  & \\
\hline
 \hline
{\multirow{6}{*}{\rotatebox[origin=c]{90}{Post$^\star$}}} & Tracked Giver Hand / Receiver & Tracked using Track Anything from G to R  & 2,720 & 710K / 780K \\
& Hand / Object Masks & & 2,720  & 1M \\
\cline{2-5}
& Giver Hand / Receiver Hand / & Segmented from fused point clouds using  tracked masks & 2,720 & 240K / 260K \\
& Object Point Clouds &  &  2,720 & 290K \\
\cline{2-5}
& 3D Model \& 6DOF Object Pose & Aligned using iterative closest point (ICP)~\cite{besl1992method} from 3D model to G, G to G+1, G+1 to G+2, $\cdots$ R-2 to R-1, and R-1 to R. &  2,720 & 290K\\
\hline
\end{tabular}}
\label{tab:data}
\end{table}

\begin{figure}
    \centering
    \includegraphics[width=\linewidth]{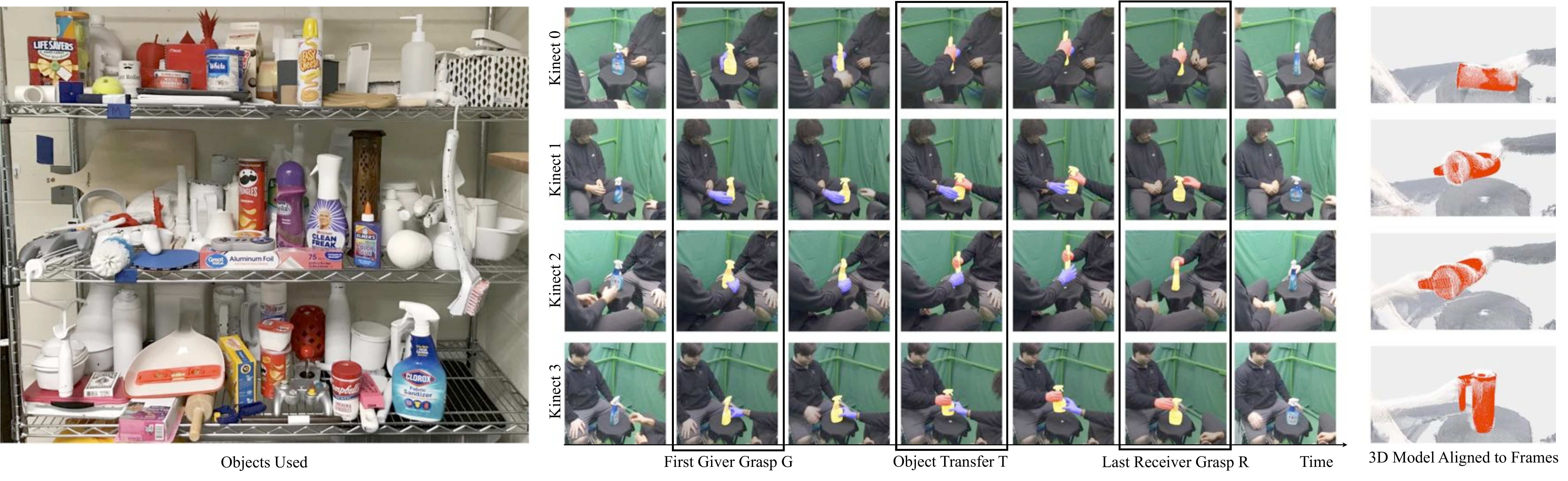}
    \caption{Left: Objects used in HOH. Center: GT annotation of key events, and SAM and GT-aided tracks of giver hand (purple), object (yellow), and receiver hand (red) segments. Right: 6DOF pose estimation via ICP-alignment of 3D model (red) to frames.}
    \label{fig:objann}
\end{figure}

\textbf{Object Dataset.} HOH's diverse 136-object dataset, shown in Figure~\ref{fig:objann}, spans 8 everyday use classes\textemdash{}toys (14), mugs (11), food and drink items (13), cooking utensils (25), tools (15), office items (11), household items (27), and 3D printed items (20). Objects are organized into 17 categories based on aspect ratio and functionality, with 8 objects per category. 15 categories span 3 aspect ratio (or form) bins\textemdash{}1:1-2:1, 2:1-3:1, and >3:1\textemdash{}and 5 bins that describe whether the object has (H) or lacks (NH) a handle, has (F) or lacks (NF) an end used for a function such as cutting, drinking, peeling, or screwing, and is found in a vertical (V) or horizontal (Z) standard orientation. The 5 bins are NFNHZ, NFNHV, FNHZ, FNHV, and FHV. Category 16 consists of 1:1 aspect ratio objects in FHZ, and category 17 consists of objects outside categories 1 to 16. We found that FHZ objects with a larger aspect ratio are uncommon. Each pair of participants interacted with 4 objects per category to cover their allocated 68 objects while ensuring full category spanning. The supplementary provides details on object preparation and acquisition.  

\textbf{Participant Recruitment.} We recruited 40 participants from the local university population, through message posting on online collaborative work spaces, after receiving approval from the university's Institutional Review Board (IRB). Participants were not compensated for providing data. The research did not involve contract work or crowd-sourcing. We had 34 male and 6 female participants, organized into 6 female/male pairs, and 14 male/male pairs. No two pairs had the same participant. 2 participants declined to share their age. Ages of remaining participants ranged from 19 to 51, with mean of 24.8$\pm$7.4. Height ranged from 1.55m to 1.96m with mean of 1.8m$\pm$0.1m. 38 of our participants reported writing right-handed, only 1 reported being a left-handed writer, and 1 reported being ambidextrous. 2 participants have not given consent for sharing identifiable color (IC) information. For safety, at this time, we exclude all color frames for all their interactions in public release, i.e., color data from 2 pairs, even if some color frames do not identify their face. 

\textbf{Experiment Procedure.} Participants were informed through the Informed Consent form that the experiment was minimal risk, participation was voluntary, participants could wear masks for COVID-19 safety and must consent to mask wearing if their partner wished it, no identifiers would be collected, and participants could request that IC images be publicly inaccessible. Participants filled a demographics questionnaire, were assigned a random 5-digit ID, and were introduced to their partner. Participants were assigned giver and receiver roles. Upon sitting at the setup, the participants performed handovers with 68 objects randomly ordered. Participants wore gloves for a random set of 4 handovers. The experimenter placed the object at tabletop center in a random orientation. Apart from committing to grasp and receiver placing object on the table, participants were free to perform interactions as they wished. Both participants filled out post-handover 7-point Likert rating on comfort using a clipboard. Participants then reversed roles, and re-performed handovers with the same 68 objects in a different order. In all but two cases, participants returned to the same seats, resulting in 22 and 18 RR pairs with left- and right-seated givers respectively.

\textbf{Data Pre-Processing.} We processed recorded camera data to obtain time-synchronized color and depth images discussed in Table~\ref{tab:data} over all views, ensuring that dropped frames are accounted for if any. We backprojected the depth pixels into 3D using the depth image intensity values. We use the transformations between the depth cameras to the reference camera 0 to fuse backprojected pixels into a single point cloud per frame number oriented in camera 0. We used depth-to-color transforms to color-map the fused point clouds. We ran OpenPose~\cite{cao2021openpose} and SAM over all Kinect color images to acquire upper body skeletons and complete image segmentation masks.

\textbf{GT Annotation and Verification.} As shown in Figure~\ref{fig:objann}, we manually annotated three frames\textemdash{}G, T, and R\textemdash{}marking key events. G contains first giver contact marking the grab portion of reach and grab, T contains simultaneous giver and receiver grasp marking the middle region of  object transfer, and R contains last receiver contact on the object marking the end of handover. We marked SAM masks for three entities \textemdash{}giver hand, object, and receiver hand\textemdash{}in G and T for giver, G, T, and R for object, and T and R for receiver. Masks were marked when present and multiple masks per entity are fused into a single mask. We verified and corrected all SAM masks. We checked all OpenPose skeletons and found that only 4.6\% had either missing or inaccurate joints. We store OpenPose confidence values. We labeled handedness and grasp type using the 28-class grasp taxonomy of Cini et al.~\cite{cini2019choice}.

\textbf{Post-Processing.} We used Track Anything~\cite{yang2023track} to track annotator-segmented SAM masks from frames G to R. We backprojected tracked masks via the camera calibration to generate giver, object, and receiver point clouds. We denoised full, giver, object, and receiver point clouds to remove outliers, and use median filtering based on the top \textit{k} nearest neighbors per point to smooth the colorization, with \textit{k} being 10. We save both pre- and post-cleaned point clouds. 

\textbf{6DOF Object Pose.} We provide GT object 6DOF pose by aligning the object 3D model to all frames from G to R for all handover interactions. First, we conduct a frame-to-frame ICP~\cite{besl1992method} alignment of the object point cloud to estimate inter-frame cloud-to-cloud transformations of the object point cloud for all frames, linking all frames from G to R. Next, we conduct an automated alignment of the 3D model to the object point cloud in the G frame. We sample a diverse range of orientations to ensure exhaustive $SO(3)$ coverage. We transform each 3D model using each orientation and fine-tune the transformed model's alignment to the object point cloud using ICP. We choose the 3D model alignment with the smallest ICP distance. We use the inter-frame transformations to align the 3D model to all frames from G to R. Example frames from the alignment are shown in Figure~\ref{fig:objann}.

\begin{figure}
    \centering
    \includegraphics[width=\linewidth]{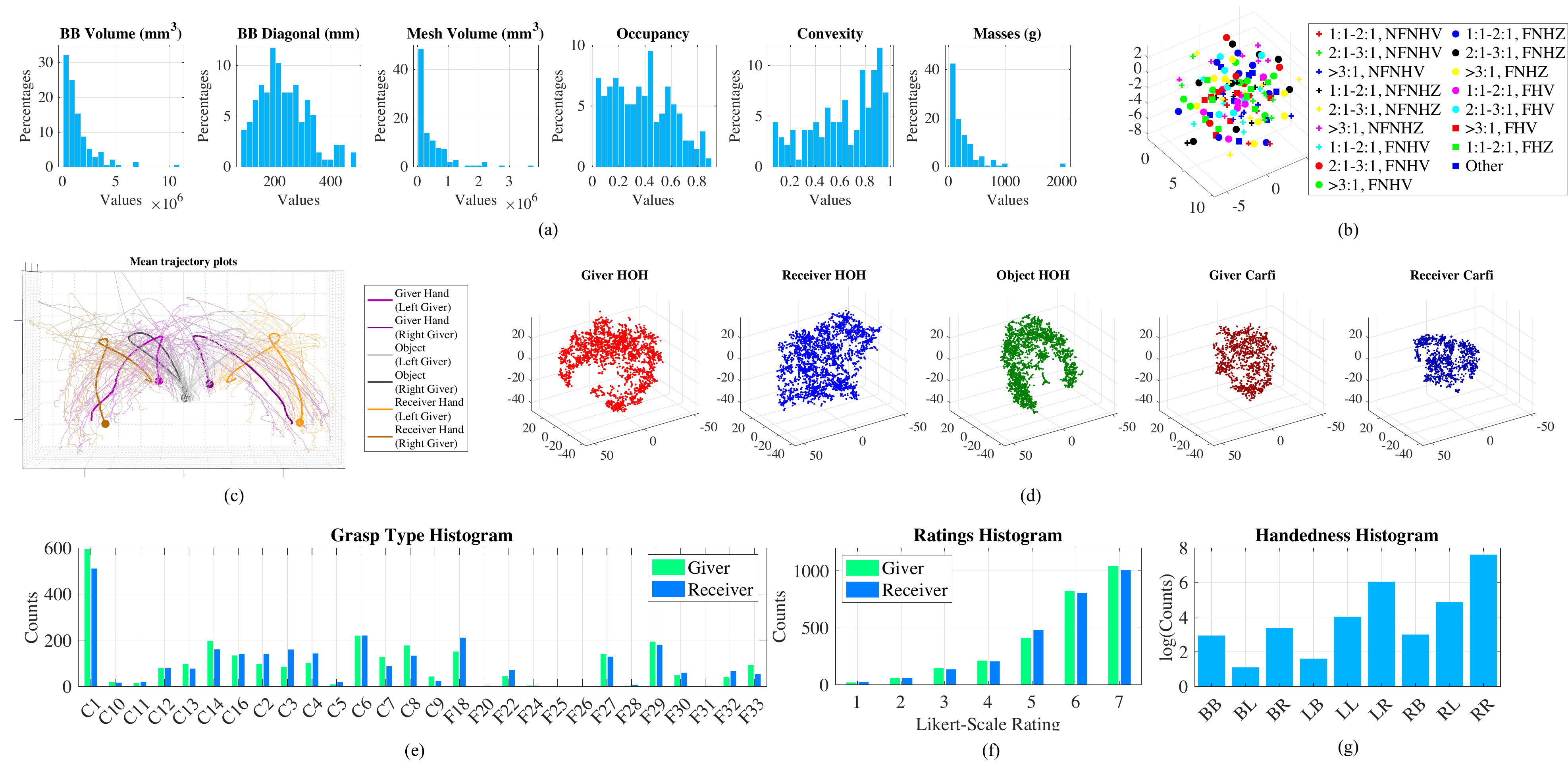}
    \caption{Dataset Analysis: (a)~histograms of object properties, (b)~t-SNE for object models, (c)~Overall (dark) and per-pair (light) mean trajectory plots with start points marked using spheres, (d)~t-SNE of trajectories for HOH compared to Carfi et al.~\cite{carfi2019multi}, (e)~histogram of grasp types per Cini et al.~\cite{cini2019choice},(f)~histogram of grasp ratings, and (g)~log histogram of handedness counts at transfer.}
    \label{fig:analysis}
\end{figure}

\subsection{Dataset Analysis}
\label{subsec:analysis}

\textbf{Objects.} Figure~\ref{fig:analysis} provides a summary of multi-modal analysis of our dataset. Object histograms in Figure~\ref{fig:analysis}(a) show that while our objects are skewed toward lower bounding box volume (BBV), mesh volume (MV), and mass (medians of $9.73\times10^6$mm$^3$, $2.54\times10^6$mm$^3$, and 167g, and skewness of 2.83, 2.77, and 3.47 respectively), the bounding box diagonal is normally distributed with mean of 233.58mm$\pm$88.50mm, and 17 objects are over 0.5kg. The bounding box occupancy (MV-to-BBV ratio) is slightly skewed toward lower values (median 0.37, skewness 0.27) indicating that HOH has high count of non-full objects, e.g., the spray bottle in Figure~\ref{fig:data}. The histogram of convexity, computed as MV-to-convex-hull ratio~\cite{attene2008hierarchical}, shows though our dataset tends toward convex objects (median 0.67, skewness -0.51), 42 objects have a convexity < 0.5, i.e., embody concavities. The spread of coefficients t-SNE plot of object geometry, in Figure~\ref{fig:analysis}(b), computed using 8,000 3D point samples per 3D object model surface, confirms the object diversity in our dataset.

\textbf{Trajectories.} Figure~\ref{fig:analysis}(c) shows overall (dark) and per-pair (light) mean trajectories for giver hand, object, and receiver hand. Means are obtained by resampling trajectories to have 100 samples, and averaging within left-seated and right-seated givers. Trajectory means show transport and object transfer phases for giver hand, inclination of object toward receiver indicative of giver intention to participate in collaborative handover, giver retraction after transfer, and looped arc for receiver toward end of handover. Left- and right-seated givers mirror each other. Spread of per-pair means demonstrates trajectory diversity, echoing prior findings of individual dependence in handover trajectory~\cite{bekemeier2019does}. HOH trajectory diversity is  confirmed by the t-SNE plot in Figure~\ref{fig:analysis}(d) where the coefficients show a higher spread than the trajectories of Carfi et al.~\cite{carfi2019multi}. To ensure comparability across HOH and Carfi trajectories, we align the trajectories within each entity and dataset using Procrustes prior to t-SNE computation, since participants may be spatially displaced in the Carfi dataset. We find trajectory lengths follow a normal distribution with giver, object, and receiver means of 3.26s$\pm$0.97s, 3.60s$\pm$0.94s, and 3.49s$\pm$0.95s. Object trajectories are longer since the receiver may react slightly later than the giver, while the giver may complete their motion before the receiver's end of handover. 

\textbf{Grasp.} The distribution of grasp types in Figure~\ref{fig:analysis}(e) organized by the taxonomy in Cini et al.~\cite{cini2019choice} shows a majority of power grasps especially near C1, or larger diameter grasps apt for HOH object sizes. High precision grasp counts are found for the giver, e.g., C6/C8 thumb-finger and C14 tripod grasp, common for rod-like or handled-equipped items. For the receiver, we see high counts for C6, F18 extension-type grasp on thin flat items, and F29 or stick grasp likely during receipt of objects with protrusion affordances. The supplementary details taxonomy nomenclature. The comfort rating histogram in Figure~\ref{fig:analysis}(f) indicates skew toward higher comfort levels (giver comfort skewness of -1.23 with 16.21\% over 4, receiver comfort skewness of -1.21 with 15.7\% over 4). We observe a high overall giver-receiver Pearson correlation of 0.38. As shown by the log histogram for handedness in Figure~\ref{fig:analysis}(g), for unimanual grasps the giver and receiver use the same hand in 76.8\% of the interactions, and opposite hands in 20.4\%. 76 interactions have bimanual grasp.

\section{Experimental Results Showing Use Case}
\label{sec:results}

The primary use of our dataset is to drive AI research in HRI. Unfortunately, no off-the-shelf algorithms exist to directly evaluate multi-person data such as HOH for robotic understanding of handover parameters. The only approach for learning-driven grasp lacks public code and data~\cite{ye2021h2o} and requires access to articulated hand pose, making it unusable with HOH. Full-fledged HOH-data-driven robotic control is outside the scope of this paper (and in fact consists of components spanning a wide range of future work). In this section, we create and evaluate \textbf{deep neural networks} to show the benefit of HOH data for \textbf{four tasks of relevance to the robotic manipulation pipeline:} (1)~Use \underline{o}bject point cloud to predict human \underline{g}iver \underline{g}rasp point cloud or \texttt{o2gg}, (2)~Use \underline{o}bject point cloud to predict object \underline{or}ientation at transfer point or \texttt{o2or}, (3)~Use object and \underline{g}iver point cloud to predict human \underline{r}eceiver \underline{g}rasp point cloud or \texttt{g2rg}, and (4)~use human \underline{g}iver hand motion to predict \underline{r}eceiver motion \underline{t}rajectory or \texttt{g2rt}. \texttt{o2gg} and \texttt{g2rg} enable robotic givers to bias grasp near preferred human giver grasp, and robot receivers to bias grasp away from / close to preferred human giver / receiver grasp. \texttt{o2or} enables robot manipulators to present objects at transfer in human-preferred poses. \texttt{g2rt} enables robot receiver motion planning in response to human giver motion.

\textbf{Implementation.} We adapt PoinTr for \texttt{o2gg} and \texttt{g2rg} and Informer for \texttt{g2rt}. We use PointNet~\cite{qi2016pointnet} with 2 dense layers to generate quaternion orientation for \texttt{o2or}. We obtain input point clouds for \texttt{o2gg} and \texttt{o2or} from Gpre, and for \texttt{g2rg} from Tpre, where Gpre and Tpre are a few frames prior to G and T, containing object only and object+giver hand respectively. Outputs for \texttt{o2gg}, \texttt{g2rg}, and \textit{o2or} consist of giver hand point cloud at G, receiver hand point cloud at T, and rotation from 6DOF pose at T represented as a quaternion. We train two networks each for \texttt{o2gg}, \texttt{o2or}, and \texttt{g2rg}, one that uses \textbf{complete} input point clouds enabling handover parameter analysis with  360$^\circ$ access to geometry through, e.g., multi-view fusion, and one that uses \textbf{partial} input point clouds, emulating single-viewpoint RGB-D sensors. We generate partial data by rendering each scene from 6 randomly generated viewpoints per scene. We pre-register all GT input and output point clouds for the complete networks to the Gpre frame, in order to assess pre-movement parameter estimation. Partial point clouds are left in the rendered viewpoint and object orientation at the appropriate handover timepoint to emulate real time behavior. Since the proposed \texttt{g2rt} focuses on point trajectories rather than objects, we use the trajectory centroids from Section~\ref{subsec:analysis} to train a single version of \texttt{g2rt}. We manually pre-transform GT to set the origin at the table center, \textit{xz}-plane as ground, and right-seated givers aligned with left-seated.

\textbf{Training.} We randomly divide the 8 objects in each form/function bin into sets A, B, and C with 3, 3, and 2 objects per bin or 51, 51, and 34 total objects to be used in train only, test only, and train+test respectively. The training set uses random 75\% of data from sets A and C. The test set uses the remaining 25\% of data from set C, and 100\% from set B. For \texttt{g2rt}, given the potential for time-varying person movement to be a behavior signature, we keep train and test participant pairs mutually exclusive. We use data from 11 pairs (22 with RR) in the training set, and the remaining 9 (18 with RR) in the test set. Hyperparameter choices and computing details are included in the supplementary.

\textbf{Evaluation Metrics.} We provide GT evaluation metrics, particularly Chamfer Distance (CD) for \texttt{o2gg} and \texttt{g2rg}, 
mean Euler angle error (MEAE) for \texttt{o2or}, and mean absolute error (MAE) for \texttt{g2rt}. As plausible parameters may not correspond to GT, we report best GT metrics to all test object instances (accounting for symmetry), to increase the chance of finding similar parameters. We report percentage overlap of giver hand with object (\%OLGO) for \texttt{o2gg} and receiver hand with giver hand (\%OLRG) for \texttt{g2rg} to gauge affordances for robot givers and receivers for safe handover.

\setlength{\tabcolsep}{6pt}
\begin{table}[t!]
    \caption{Metrics for experimental results.}
    \centering
    {\small\begin{tabular}{@{}|l|l|c|c|c|c|@{}}
    \hline
    & & \multicolumn{2}{c|}{\textbf{Complete}} & \multicolumn{2}{c|}{\textbf{Partial}}\\
    \hline
    \textbf{Task} & \textbf{Metric} & \textbf{GT} & \textbf{Best to Object} & \textbf{GT} & \textbf{Best to Object}\\
    \hline
    \hline
    \texttt{o2gg} & CD & 0.483$\pm$0.187 & 0.135$\pm$0.078 & 0.135$\pm$0.052 & 0.277$\pm$0.188\\
     \hline
     \texttt{o2or} & MEAE & 0.851$\pm$0.253 & 0.365$\pm$0.153 & 0.843$\pm$0.227 & 0.411$\pm$0.136\\
    \hline
    \texttt{g2rg} & CD & 0.539$\pm$0.648 & 0.147$\pm$0.182 & 0.206$\pm$0.121 & 0.288$\pm$0.275\\
    \hline
    \texttt{g2rt} & MAE & 0.160$\pm$0.070 & 0.107$\pm$0.037 & N/A & N/A\\
    \hline
    \multicolumn{6}{c}{}\\
    \hline
    \textbf{Task} & \textbf{Metric} & \multicolumn{2}{c|}{\textbf{Complete}} & \multicolumn{2}{c|}{\textbf{Partial}}\\
    \hline
    \hline
    \texttt{o2gg} & \%OLGO & \multicolumn{2}{c|}{17.39\%$\pm$10.37\%} & \multicolumn{2}{c|}{58.12\%$\pm$19.46\% }\\
    \hline
    \cline{2-6}
     \texttt{g2rg} & \%OLRG & \multicolumn{2}{c|}{1.44\%$\pm$4.73\%} & \multicolumn{2}{c|}{1.13\%$\pm$3.87\%}\\    
    \hline
    \end{tabular}}
    \label{tab:results}
\end{table}

\begin{figure}[t!]
    \centering
    \includegraphics[width=\linewidth]{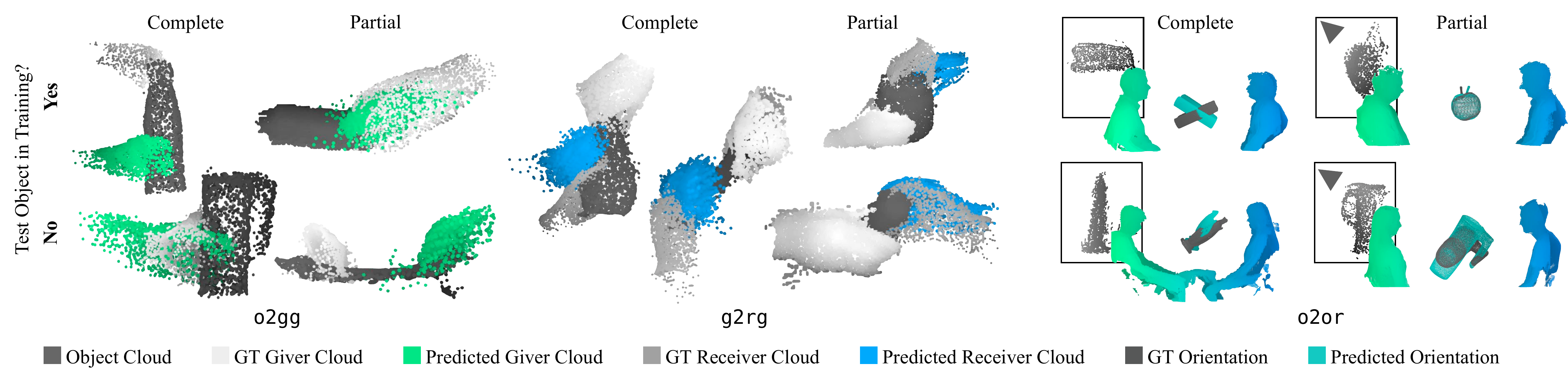}
    \caption{Visualization of outputs from \texttt{o2gg}, \texttt{g2rg}, and \texttt{o2or}. Insets for \texttt{o2or} show input point cloud. Camera angle shown for partial clouds.}
    \label{fig:results}
\end{figure}

\textbf{Results.} Table~\ref{tab:results} shows the results of the evaluation metrics. Figures~\ref{fig:results} shows qualitative results of predictions for \texttt{o2gg}, \texttt{g2rg}, \texttt{o2or}, and \texttt{g2rt}, the latter two contextualized within the handover providing the source input. Additional results are provided in the supplementary. When predictions diverge from GT, they correspond to plausible outputs, e.g., though the actual giver grasp for the grill brush at the bottom left of Figure~\ref{fig:results} for \texttt{o2gg} is on the handle, the predicted grasp is on the brush. Variability receiver hand orientation is observed for \texttt{g2rg}. For \texttt{o2or}, we notice that the alignment of the predicted object when deviating substantially from GT, corresponds to a plausible extension direction during transfer. Outputs of \texttt{g2rt} show that, even without object structure, simple trajectory prediction enables the receiver to meet the giver trajectory near the shared space. From Table~\ref{tab:results}, we notice that the mean best-to-object metrics show a drop compared to the GT, indicating that a more likely candidate for each parameter is found in the dataset, and demonstrating the dataset's diversity in representation of handover. \%OLGO in Table~\ref{tab:results}, indicates  affordance availability, especially using complete data. \%OLRG demonstrates that giver-receiver overlap is negligible, showing ability of \texttt{g2rg} to predict away from giver hand and enabling its use for robot receiver grasp biasing.

\begin{figure}[t!]
    \centering
    \includegraphics[width=\linewidth]{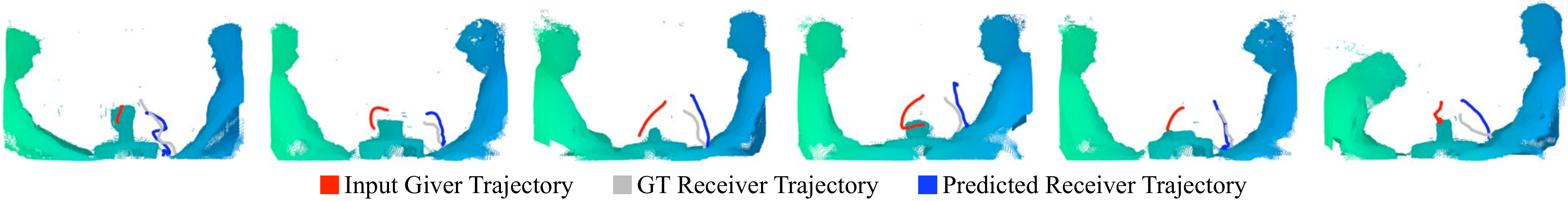}
    \caption{Visualization of outputs from \texttt{g2rt} for various pairs and objects.}
    \label{fig:resultstraj}
\end{figure}

\section{Discussion} 
\label{sec:limitationsimpacts}

\textbf{Future Work.} The experimental results presented in Section~\ref{sec:results} provide a starting point for algorithm development in HRI using HOH. The diverse, multimodal, and richly annotated data in HOH greatly opens the scope for large-scale AI algorithms in HRI. Algorithms can be developed to leverage comfort ratings to rank grasp, pose, and path prediction outputs, conduct realtime object-dependent robotic giver motion prediction and proactive generation of robotic receipt motion by giver motion forecasting, and enhance the robotics pipeline with object segmentation and tracking algorithms in the presence of multi-person occlusions due to handover. Transfer position estimation can be conducted by using upper body point clouds and/or skeletons to establish spatial relationships. The analysis in Section~\ref{subsec:analysis} provides an overall view of HOH. A large scope exists for future studies involving detailed analyses of alignment between giver and receiver comfort, trajectory velocity and timing, and coordination, with respect to grasp type, object categories, and participant pairs.

\textbf{Limitations.} HOH lacks grip force due to the use of instrumentation-free setup. Currently, grip force is collected using heavily-designed batons~\cite{mason2005grip,chan2012grip,controzzi2018humans,khanna2022human} that constrain grasp type, are difficult to control for weight, and lack geometric diversity. The scope remains to instrument everyday objects with reliable minimally-invasive grip force units. Our participants occupy a narrow range over age and ability. We recognize the challenges in recruiting participants such as children, older adults, and individuals with different abilities,
health and social concerns toward mutual interactions with unfamiliar partners, and potential cognitive barriers toward providing informed consent. Future collections can benefit from best practices on population-specific recruitment~\cite{mody2008recruitment,campbell2008their,banas2019recruiting}. Though HOH has 6DOF object pose, it currently lacks GT hand pose annotation. Full manual annotation of hand pose in markerless data in the presence of occlusions and motion is a daunting task. 
Future work includes adapting HOnnotate~\cite{hampali2020honnotate} to operate with multi-person hand interactions.

\textbf{Societal Impacts.} HOH provides the \textbf{societal benefit} of informing social robots on how to perform safe handover, important toward the development and enhancement of trust and collaborative goal accomplishment~\cite{hamacher2016believing,pan2018exploration}. Social robots aware of object-dependent handover improve fluidity of post-handover operations involving multi-object use such as assembly and activities of daily living. Improved trustworthy social robots, engaging in safe handover of objects have the potential to fill current shortages of in-home aides for older adults~\cite{caregivershortage0} and individuals with disabilities, expected to be of special concern in the light of a declining caregiving workforce~\cite{caregivershortage1} and increased demand~\cite{usbls2022}. Joint attention on shared objects, a component of handover, has been shown to encourage social closeness~\cite{wolf2016joint}. Long-term impact of human-robot bonding on older adult isolation or childhood development remains understudied. 
Social robots developed without coupling short-range focused data such as ours with studies on longer-range attachment behavior are likely to introduce \textbf{negative societal consequences}. If a robot on which an individual has developed a dependence malfunctions or has to be discontinued, it may introduce anxiety and regression of collaborative goals within the individual, akin to the loss of a loved one. We share full details of our setup, capture, computing, and code in supplementary to promote continued wider-scale data collections for societally-aware HRI.

\section*{Acknowledgments}
This work was funded by National Science Foundation grant IIS-2023998. We thank Mingjun Li and Nikolas Lamb for insightful discussions on algorithm development and sensor calibration. We also thank Priyo Ranjan Kundu Prosun, Thomas Dubay, Ben Molloy, Irfan Yaqoob, Numan Zafar, Jyothinadh Minnekanti, Sichao Li, Houchao Gan, Xinchao Song, Kun Han, Alaina Tulskie, Holly Rossmann, Rosalina Delwiche, Cameron Hood, Odin Kohler, Alexander Cohen, Christian Soucy, and Gianna Voce for assistance with data collection and data annotation. 

{\small
\bibliographystyle{plain}
\bibliography{neurips_2023}
}

\end{document}


\maketitle

\section{Dataset Link and Password}
A landing page for access to the data has been created as follows:
\begin{enumerate}
    \item Dataset Page: \url{https://hohdataset.github.io/}
    \item Dataset Access: email first author (see landing page)
\end{enumerate}

\section{Dataset Information}
\label{sec:overview}

The HOH dataset contains multimodal data from a variety of cameras. This data has been processed to include skeletons, point clouds, and segmentation masks. A summary of the included data is provided in Figure~\ref{fig:data_card}, as well as section 3 in the main paper.

\newlength{\NFwidth}
\setlength{\NFwidth}{4.5in}

\NewDocumentCommand{\NFelement}{mmm}{\normalsize\textbf{#1} #2\hfill #3}
\NewDocumentCommand{\NFline}{O{l}m}{\footnotesize\makebox[\NFwidth][#1]{#2}}

\NewDocumentCommand{\NFentry}{sm}{%
  \makebox[.5\NFwidth][l]{\normalsize
    \IfBooleanT{#1}{\makebox[0pt][r]{\textbullet\ }}%
    #2}\ignorespaces}
\NewDocumentCommand{\NFtext}{+m}
 {\parbox{\NFwidth}{\raggedright#1}}

\newcommand{\NFtitle}{\multicolumn{1}{c}{\huge\bfseries HOH Dataset Facts}}

\newcommand{\NFRULE}{\midrule[5pt]}
\newcommand{\NFRule}{\midrule[2pt]}
\newcommand{\NFrule}{\midrule}

\begin{figure}
    \centering
    \resizebox{11cm}{!}{
    
    \sffamily
{
\fbox{%
\begin{tabular}{@{}p{\NFwidth}@{}}

\NFtitle\\
\NFrule
\NFtext{\textbf{Dataset} Human-Object-Human Handover}\\

\NFRULE
\NFline{Motivation}\\
\NFrule
\NFelement{Summary}{}{A markerless 3D multimodal dataset on human-human handovers with 20 participant pairs using 136 objects spanning 8 everyday-use classes}\\
\NFelement{Example Use Cases}{  Human-robot handover research, cognitive psychology research, exploration of automated human and object pose estimation algorithms}\\
\NFelement{Original Authors}{}{N. Wiederhold, A. Megyeri, D. Paris,}\\
\NFelement{}{}{S. Banerjee, N. K. Banerjee}\\

\NFRULE
\NFline{Metadata}\\
\NFrule
\NFelement{URL}{}{\url{https://hohdataset.github.io}}\\
\NFelement{Released}{}{October 27, 2023}\\

\NFRULE
\NFline{Sensors}\\
\NFrule
\NFelement{Azure Kinect Color}{}{4}\\
\NFelement{Azure Kinect Depth}{}{4}\\
\NFelement{FLIR PointGrey Blackfly S High Speed Color}{}{4}\\

\NFRule
\NFline{Object Classes}\\
\NFrule
\NFelement{Total Objects}{}{136}\\
\NFelement{Toys}{}{19}\\
\NFelement{Food/Drink}{}{19}\\
\NFelement{Cooking}{}{24}\\
\NFelement{Tool}{}{15}\\
\NFelement{Mug}{}{12}\\
\NFelement{Office}{}{11}\\
\NFelement{Household}{}{36}\\

\NFRULE
\NFline{Participants}\\
\NFrule
\NFelement{Total participants }{}{40}\\
\NFelement{Total Pairs}{}{20}\\
\NFelement{M/M Pairs}{}{16}\\
\NFelement{M/F Pairs}{}{4}\\
\NFelement{Gender}{}{34M, 6F}\\
\NFelement{Age}{}{24.8$\pm$7.4}\\

\NFRULE
\NFline{Data Size}\\
\NFrule
\NFelement{Total Size}{}{9.51 TB}\\
\NFelement{Azure}{}{195.80 GB}\\
\NFelement{Mask Tracking}{}{2.37 GB}\\
\NFelement{OpenPose}{}{927.38 MB}\\
\NFelement{PCFiltered}{}{94.34 GB}\\
\NFelement{PCFull}{}{4.54 TB}\\
\NFelement{PointGrey}{}{4.66 TB}\\
\NFelement{3D Model Alignments}{}{26.7 MB}\\
\end{tabular}}

\par} 
    
    }
    \caption{A dataset informational card for HOH}
    \label{fig:data_card}
\end{figure}

\paragraph{Data Format}
All capture data is saved in 178 directories that represent the recording of multiple interactions. The number of interactions differs in every directory due to a 10-minute maximum recording time. Some recordings were stopped before 10 minutes had elapsed in order to redo a mistake made by the participants, e.g. the receiver accidentally picked up the object after the role swap. This resulted in some recordings having very few interactions. The naming format for these 178 directories is \textit{<giver ID>-<receiver ID>-S<starting interaction number>}. For example, the directory for the first recording of giver 01638 and receiver 46157 is named "01638-46157-S1". In each of these 178 directories, there are 6 sub-directories:
\begin{enumerate}
    \item \textit{Azure} - Contains Azure Kinect color videos and Azure Kinect depth videos for each of the 4 cameras. Color videos are named [NUM].mp4, depth videos [NUM].mkv, and viewable depth videos as [NUM]\_depth.mp4, where [NUM] represents a number from 0 to a maximum of 18, based upon the quantity of interactions that would have fit in the 10-minute duration.
    \item \textit{MaskTracking} - Contains a zip file with all tracked masks within the directory. The zip folder [RECORDING]\_mask.zip contains directories for each of the 4 cameras. In each camera folder are all interaction object and hand mask files saved as Python NPZ format. The NPZ file contains a stacked numpy array of masks, one for each frame in the interaction. The zip folder [RECORDING]\_maskcorrected.zip has a similar structure, though each NPZ file contains an individual mask for the frame that was fixed.
    \item \textit{OpenPose} - Contains a zip file with all skeletons in JSON format, following the [NUM] notation, one [NUM] folder per interaction.
    \item \textit{PCFiltered} - Contains a zip file with all object and hand point clouds, following the [NUM] notation. Cleaned versions of the point clouds are also available in the \textit{Cleaned} folder for each interaction.
    \item \textit{PCFull} - Contains a zip file with all full scene point clouds, following the [NUM] notation.
    \item \textit{PointGrey} - Contains a zip file with all Point Grey images.
    \item \textit{3dModelAlignments} - Contains a zip file with transformations that align the 3D model of the object used in each handover to the object in the scene point cloud for each timestep.
\end{enumerate}

Within each directory, point clouds, videos, and masks are generated between the giver and receiver contact frames (G and R) inclusive, and where trackable for masks and segments. Other than the 178 data directories, there exist 4 directories, called \textit{Objects}, \textit{Code}, \textit{Calibration}, and \textit{ParticipantInfo}:

The \textit{Objects} directory contains 3D models and metadata for all 136 objects. All of the 3D models are stored in a sub-directory called \textit{3d\_models}. Inside of \textit{3d\_models}, there is one directory corresponding to each object, named according to object ID. For all objects excluding 116 and 120, multiple 3D models are present which are discussed in Section~\ref{sec:ObjectDataset}.

The \textit{Code} directory contains all code used to collect and process the data. The \textit{Code} directory contains three sub-directories, named \textit{Acquisition}, \textit{Experiments}, and \textit{Processing}. The \textit{Acquisition} directory contains all code used for data acquisition, sorted into two sub-directories by language used: C\# and Python. The \textit{Experiments} directory contains all code used for the experiments described in Section~\ref{sec:ComputingAndTrainingDetails}, broken down into \textit{Grasp}, \textit{Orientation}, and \textit{Trajectory} sub-directories. The \textit{Processing} directory contains all code used for data processing. All code will be published on GitHub for public use upon acceptance, along with documentation in README files. 

The \textit{ParticipantInfo} directory contains the following:
\begin{enumerate}
    \item \textit{demographics\_responses.csv} - The answers submitted by the participants for the demographics questionnaire.
    \item \textit{grasp\_handedness.csv} - Grasp handedness labels for each interaction.
    \item \textit{grasp\_taxonomy.csv} - Grasp taxonomy labels for each interaction.
    \item \textit{participant\_seating.json} - Participant seating arrangements, organized by whether the giver is sitting on the left or right of the capture environment.
    \item \textit{Participant\_Form\_Responses} - A directory that contains all data collected from the digitization of the participant and experimenter forms, as detailed in Paragraph~\ref{text:paper-form-digitization}. The naming format inside this directory follows the convention: \textit{<giver ID>-<receiver ID>.csv}.
\end{enumerate}

The \textit{Calibration} directory contains the following:
\begin{enumerate}
    \item \textit{group1\_calib} - A directory containing calibration intrinsic and extrinsic parameters for sessions that use object set 1 as detailed in Section~\ref{text:sensor_calib}.
    \item \textit{group2\_calib} - A directory containing calibration intrinsic and extrinsic parameters for sessions that use object set 2 as detailed in Section~\ref{text:sensor_calib}.
    \item \textit{fine\_tuned\_transforms} - A directory containing the fine-tuned transformations for all sessions as detailed in Section~\ref{text:sensor_calib}.
\end{enumerate}

As shown by the data card in Figure~\ref{fig:data_card}, the dataset is nearly 9.51 TB, with the main high-file-size component being the full point clouds. Azure data is available in the form of videos. Though Point Grey data is currently shared as images, we plan to compress them into video files and expect the file sizes to be approximately twice the size of the Azure color videos, thereby greatly improving compression. Filtered point clouds for the objects and hands occupy a considerably reduced size on disk due to their small point count. We plan to provide down-sampled options for full point clouds. 

\begin{figure}
    \centering
    \includegraphics[width=\linewidth]{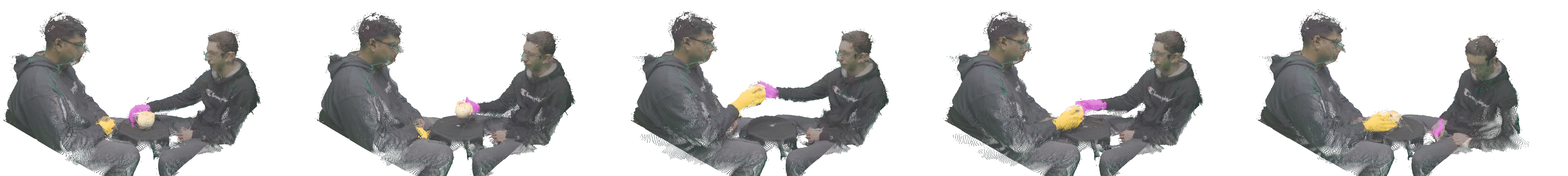}
    \includegraphics[width=\linewidth]{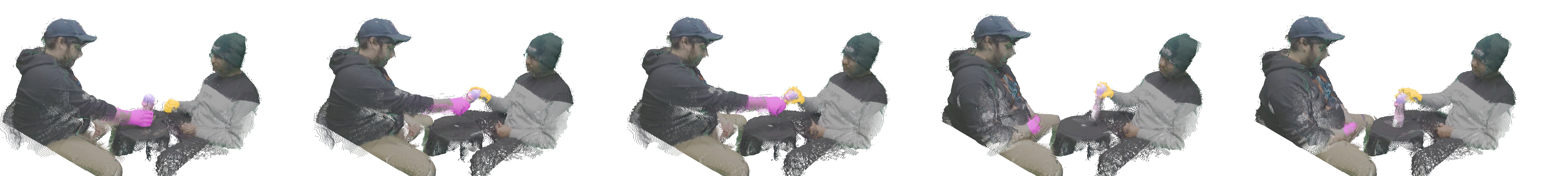}
    \includegraphics[width=\linewidth]{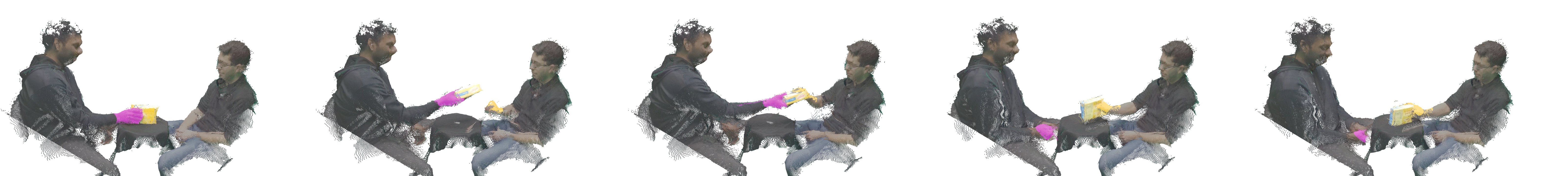}
    \includegraphics[width=\linewidth]{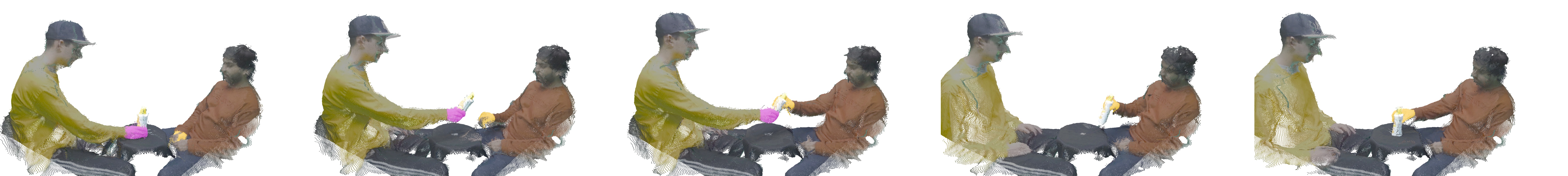}
    \includegraphics[width=\linewidth]{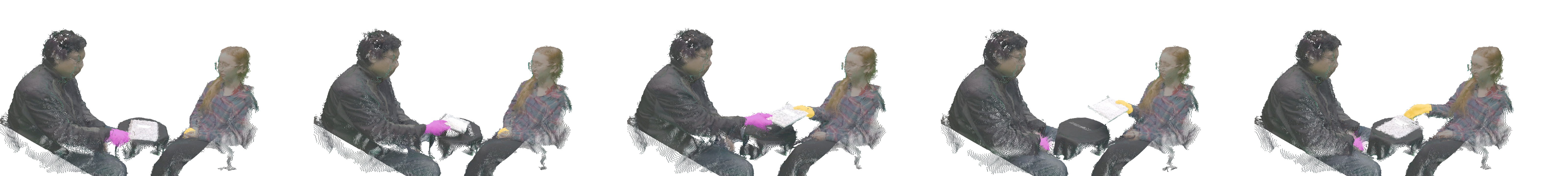}
    \includegraphics[width=\linewidth]{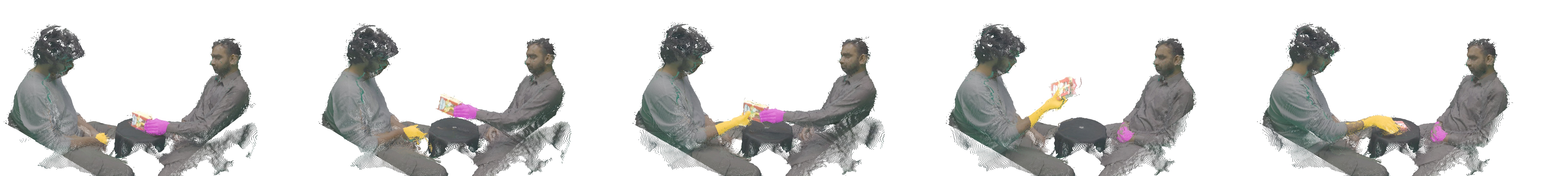}
    \includegraphics[width=\linewidth]{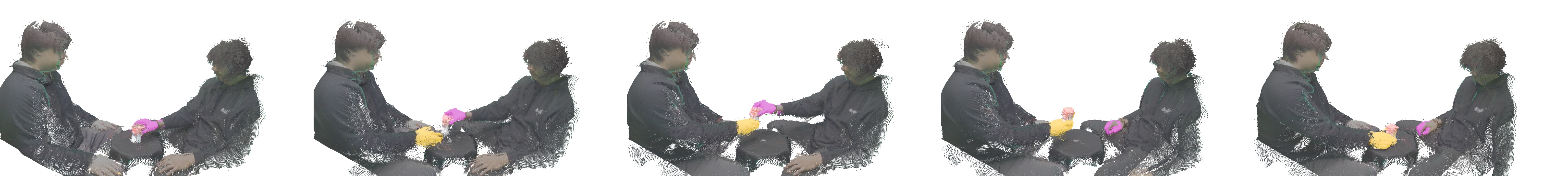}
    \includegraphics[width=\linewidth]{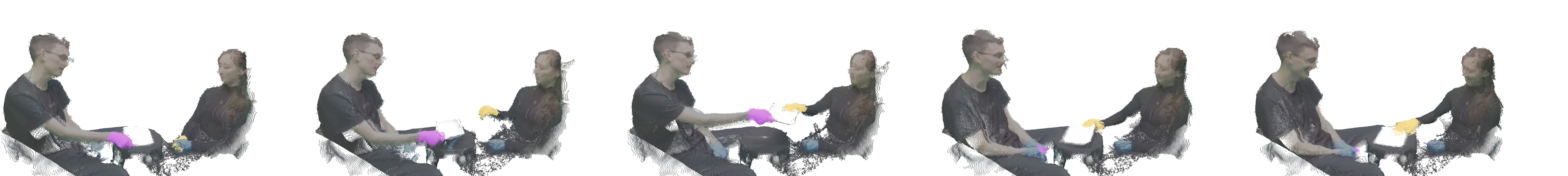}
    \includegraphics[width=\linewidth]{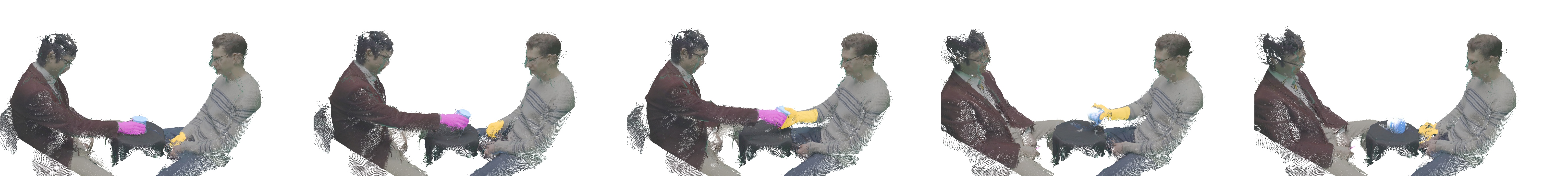}
    \caption{Example 3D visualizations of full scene point clouds at 5 time points during a handover interaction, with Frame G (point of first giver contact) in the leftmost column, Frame T (point of transfer) in the center column, and Frame R (point of last receiver contact) in the rightmost column. The giver hand is highlighted magenta and the receiver hand is highlighted gold.}
    \label{fig:full_3d_vis}
\end{figure}

Example 3D visualizations including full scene point clouds and isolated giver hand, object, and receiver hand point clouds, are shown in Figure~\ref{fig:full_3d_vis}.

\paragraph{License Information.}
\label{text:license}
We license all new assets in the dataset, including but not limited to the color and depth images, all versions of object models, manual annotations, all varieties of point clouds, segmentation masks, body skeletons, and participant demographic and comfort data, under the Creative Commons Attribution-NonCommercial 4.0 International (CC BY-NC 4.0) License (\url{https://creativecommons.org/licenses/by-nc/4.0/}), with the only exceptions being those object models in Table~\ref{tab:license}. We make use of 14 3D models from Thingiverse, for which licensing information is displayed in Table~\ref{tab:license}, and attribution information in Table~\ref{tab:thingiverse_attribution}. We do not release altered meshes for objects 116 and 120. We assign them an ID that is compliant with our naming and categorization scheme for objects. All code publicly released with this dataset, including code which allows for loading, modification, and application of the data, is licensed under the MIT License (\url{https://opensource.org/license/mit/}).

\begin{table}[htbp]
    \centering
    \caption{Licensing information for object models acquired from Thingiverse.}
    \small
    \begin{tabular}{|c|l|}
    \hline
    \textbf{Object ID} & \textbf{License}\\
    \hline
    115 & Creative Commons - Attribution License \\
    116 & Creative Commons - Attribution - Non-Commercial - No Derivatives License \\
    118 & Creative Commons - Attribution License \\
    120 & Creative Commons - Attribution - Non-Commercial - No Derivatives License \\
    121 & Creative Commons - Attribution License \\
    122 & Creative Commons - Attribution - Non-Commercial - Share Alike License \\
    127 & Creative Commons - Attribution - Non-Commercial License \\
    128 & Creative Commons - Public Domain Dedication License \\
    129 & Creative Commons - Attribution License \\
    221 & Creative Commons - Attribution - Share Alike License \\
    233 & Creative Commons - Attribution - Non-Commercial - Share Alike License \\
    234 & Creative Commons - Attribution License \\
    235 & Creative Commons - Attribution - Non-Commercial License \\
    236 & Creative Commons - Attribution License \\
    \hline
    \end{tabular}
    \label{tab:license}
\end{table}

\begin{table}[htbp]
    \centering
    \caption{Licensing attribution for object models acquired from Thingiverse.}
    \small
    \begin{tabular}{|c|l|l|}
    \hline
    \textbf{Object ID} & \textbf{Thingiverse Author} & \textbf{Link To Thingiverse Page}\\
    \hline
    115 & @RandomUser23447274 & \url{https://www.thingiverse.com/thing:4694553} \\
    116 & @Clms31 & \url{https://www.thingiverse.com/thing:4690097} \\
    118 & @ertugrulozarozar & \url{https://www.thingiverse.com/thing:4715797} \\
    120 & @MarVin\_Miniatures & \url{https://www.thingiverse.com/thing:4038181} \\
    121 & @bert\_lz & \url{https://www.thingiverse.com/thing:4688251} \\
    122 & @stratosvasilas & \url{https://www.thingiverse.com/thing:4694905} \\
    127 & @riskable & \url{https://www.thingiverse.com/thing:2173745} \\
    129 & @Cool3DModel & \url{https://www.thingiverse.com/thing:2445539} \\
    221 & @david4974 & \url{https://www.thingiverse.com/thing:1617958} \\
    233 & @sffubs & \url{https://www.thingiverse.com/thing:4684367} \\
    234 & @tobymerritt & \url{https://www.thingiverse.com/thing:4695393} \\
    235 & @Onil\_Creations & \url{https://www.thingiverse.com/thing:4700386} \\
    236 & @clanmcfadden & \url{https://www.thingiverse.com/thing:4688105} \\
    \hline
    \end{tabular}
    \label{tab:thingiverse_attribution}
\end{table}

\paragraph{Author Statement of Responsibility.}
The authors confirm all responsibility in case of violation of rights and confirm the license associated with the dataset and code.

\paragraph{Dataset Accessibility and Long-Term Preservation Plan.}
\label{text:access_preservation_plan}
Upon acceptance, we plan to host the full dataset on our local datacenter and make it available through the project webpage. We plan to host a compressed version of the dataset on Google Drive associated with our institution. Code will be hosted on GitHub. The project page will be hosted on GitHub to ensure that the data remains accessible.

\paragraph{Dataset Identifier.}
Access the persistent landing page for the dataset here: \url{https://tinyurl.com/hohdataset}

\subsection{Datasheets For Datasets}
We follow the framework of Datasheets for Datasets~\cite{gebru2018datasheet} for our dataset documentation and intended uses. 

\begin{enumerate}
    \item \textbf{Motivation} 
    \begin{enumerate}
        \item \textbf{For what purpose was the dataset created?}
        \newline To accelerate data-driven research on handover studies, human-robot handover implementation, and artificial intelligence on handover parameter estimation from reality-representative 2D and 3D data of natural person interactions.
        \item \textbf{Who created the dataset (e.g., which team, research group) and on behalf of which entity (e.g., company, institution, organization)?} 
        \newline Terascale All-sensing Research Studio at Clarkson University.
        \item \textbf{Who funded the creation of the dataset?}
        \newline This work was funded by National Science Foundation grant IIS-2023998.
    \end{enumerate}
    \item \textbf{Composition} 
    \begin{enumerate}
        \item \textbf{What do the instances that comprise the dataset represent (e.g., documents, photos, people, countries)?} 
        \newline The dataset is comprised of images (.png, .jpg), videos (.mp4,mkv), texture-mapped point clouds (.ply), segmentation masks (.npz), skeletons (.json), object models (.obj), and spreadsheets (.csv).
        \item \textbf{How many instances are there in total (of each type, if appropriate)?} 
        \newline See Figure~\ref{fig:data_card} and main paper Table 2 for detailed breakdown.
        \item \textbf{Does the dataset contain all possible instances or is it a sample (not necessarily random) of instances from a larger set?} 
        \newline The dataset contains all possible instances.
        \item \textbf{What data does each instance consist of?} 
        \newline Each data instance is a single handover interaction, including color and depth images, point clouds, segmentations, skeletons, and annotations.
        \item \textbf{Is there a label or target associated with each instance?} 
        \newline Each handover has an object 3D model associated with it. Also, each handover is linked to rich metadata including participant comfort ratings, object metadata, participant demographics information, and experimenter notes. Each handover is named according to the ID numbers of the participants involved and a serial number denoting the place of the interaction in the overall sequence.
        \item \textbf{Is any information missing from individual instances?} 
        \newline A few Azure Kinect images were dropped during data collection. Full scene point clouds, skeletons, and masks are missing for these dropped frames.
        \item \textbf{Are relationships between individual instances made explicit (e.g., users’ movie ratings, social network links)?} 
        \newline All instances are explicitly grouped by the participant dyad involved.
        \item \textbf{Are there recommended data splits (e.g., training, development/validation, testing)?} 
        \newline Not at this time. Users of this dataset are encouraged to experiment and divide the dataset as it suits their applications.
        \item \textbf{Are there any errors, sources of noise, or redundancies in the dataset?} 
        \newline An extraction error occurred that affects <1\% of the Point Grey color images. The depth data can be somewhat noisy when recording from as far away as the depth sensors are in our system. This is mitigated through 2D segmentation and 3D-based noise and outlier removal. No redundancies exist based on the knowledge of the authors.
        \item \textbf{Is the dataset self-contained, or does it link to or otherwise rely on external resources (e.g., websites, tweets, other datasets)?} 
        \newline The dataset is self-contained, though it includes 14 other 3D models for objects. See Table~\ref{tab:license} for details.
        \item \textbf{Does the dataset contain data that might be considered confidential (e.g., data that is protected by legal privilege or by doctor patient confidentiality, data that includes the content of individuals’ non-public communications)?} 
        \newline No.
        \item \textbf{Does the dataset contain data that, if viewed directly, might be offensive, insulting, threatening, or might otherwise cause anxiety?}
        \newline No.
        \item \textbf{Does the dataset identify any subpopulations (e.g., by age, gender)?} 
        \newline This information is present in the demographics responses from each participant, though the dataset is not subdivided corresponding to any of the demographics data.
        \item \textbf{Is it possible to identify individuals, either directly or indirectly (i.e., in combination with other data) from the dataset?} 
        \newline Yes. The dataset includes color data depicting all participants who consented to the public release of their color data.
        \item \textbf{Does the dataset contain data that might be considered sensitive in any way (e.g., data that reveals race or ethnic origins, sexual orientations, religious beliefs, political opinions or union memberships, or locations; financial or health data; biometric or genetic data; forms of government identification, such as social security numbers; criminal history)?}
        \newline No.
    \end{enumerate}
    \item \textbf{Collection Process} 
    \begin{enumerate}
        \item \textbf{How was the data associated with each instance acquired?} 
        \newline Images were collected using 4 Microsoft Azure Kinect cameras and 4 FLIR Point Grey Blackfly S cameras. See Paragraph~\ref{sec:DataCaptureSystem} for details about the sensors, and see Section 3 in the main paper for procedure details. 3D models were scanned using an Einscan-SP 3D-scanner, as detailed in Paragraph~\ref{sec:ObjectDataset}. 
        \item \textbf{What mechanisms or procedures were used to collect the data (e.g., hardware apparatuses or sensors, manual human curation, software programs, software APIs)?}  
        \newline For detail about the sensors used, see Section~\ref{sec:DataCaptureSystem}. For detail about the data collection procedure, see Section 3 in the main paper. All code to control sensors and manipulate data was written internally, excluding the Azure Kinect API and Spinnaker API which are used to control sensors.

        \item \textbf{Who was involved in the data collection process and how were they compensated)?} 
        \newline Students were recruited to administrate the data collection sessions. The student experimenters were compensated with course credit.
        \item \textbf{Over what timeframe was the data collected?} 
        \newline The data was collected between February 24, 2023 and April 5, 2023.
        \item \textbf{Were any ethical review processes conducted (e.g., by an institutional review board)?} 
        The project received approval from the institutional review board prior to data collection. 
        \item \textbf{Did you collect the data from the individuals in question directly, or obtain it via third parties or other sources?} 
        \newline Data was collected from the participants directly.
        \item \textbf{Were the individuals in question notified about the data collection?}
        \newline Yes. Participants were recruited voluntarily. The message used to recruit the participants can be found in Paragraph~\ref{text:participant-recruitment}. The speech read to participants that details the data collected can be found in Paragraph~\ref{text:participant-arrival}.
        \item \textbf{Did the individuals in question consent to the collection and use of their data?}
        \newline Yes, each participant signed an informed consent document where they consented to being videotaped and allowed the release of their color data and non-identifiable data. Two participants in this study did not consent to having their color data publicly released, and in these cases we withhold any of their identifiable data and release their non-identifiable data.
        \item \textbf{If consent was obtained, were the consenting individuals provided with a mechanism to revoke their consent in the future or for certain uses?}
        \newline Participants were reminded on multiple occasions that they may stop the study at any point if they wish. It was clarified that a participant may ask for their data to be deleted at any time, even after the conclusion of the study, by communicating with an experimenter and mentioning their 5-digit participant ID number.
        \item \textbf{Has an analysis of the potential impact of the dataset and its use on data subjects (e.g., a data protection impact analysis) been conducted?}   
        \newline Yes, for instance, a comprehensive assessment of risks has been performed and communicated to the participants via the informed consent form. Subjects also had the opportunity of requesting that identifiable color not be released via the informed consent form. Our dataset does not release color data for two participant pairs based on one participant in each pair having opted out of identifiable color data release.
    \end{enumerate}
    \item \textbf{Preprocessing, Cleaning and Labelling} 
    \begin{enumerate}
        \item \textbf{Was any preprocessing/cleaning/labeling of the data done (e.g., discretization or bucketing, tokenization, part-of-speech tagging, SIFT feature extraction, removal of instances, processing of missing values)?} 
        \newline Yes, we performed annotation and ran software on the images. See question 4(c).
        \item \textbf{Was the “raw” data saved in addition to the preprocessed/cleaned/labeled data (e.g., to support unanticipated future uses)?} 
        \newline Yes. The raw data is saved separately as videos and images.
        \item \textbf{Is the software that was used to preprocess/clean/label the data available?}  
        \newline Yes. SAM~\cite{kirillov2023segany}, OpenPose~\cite{cao2021openpose}, and Track Anything~\cite{yang2023track} are publicly available.
    \end{enumerate}
    \item \textbf{Uses} 
    \begin{enumerate}
        \item \textbf{Has the dataset been used for any tasks already?} 
        \newline No.
        \item \textbf{Is there a repository that links to any or all papers or systems that use the dataset?} 
        \newline As the dataset has not been publicly released yet, there are no papers that use the dataset at present.
        \item \textbf{What (other) tasks could the dataset be used for?}
        \newline The dataset could be used for training assistive robots for purposes such as in-home care for the elderly or providing help in the kitchen by retrieving a utensil or an ingredient when a person may not have the free hands or the time to do it on their own.
        \item \textbf{Is there anything about the composition of the dataset or the way it was collected and preprocessed/cleaned/labeled that might impact future uses?} 
        \newline The use of entirely non-invasive, markerless data collection techniques could impact the ability to obtain dense ground truth data, e.g. object pose, without substantial manual effort.
        \item \textbf{Are there tasks for which the dataset should not be used?}  
        \newline This dataset should not be used to cause a robot to intentionally give an object unsafely, e.g. extend the blade of a knife directly toward a human user.
    \end{enumerate}
    \item \textbf{Distribution}
    \begin{enumerate}
        \item \textbf{Will the dataset be distributed to third parties outside of the entity on behalf of which the dataset was created?} 
        \newline Yes. The dataset will be made publicly available upon acceptance.
        \item \textbf{How will the dataset will be distributed (e.g., tarball on website, API, GitHub)?} 
        \newline The dataset will be hosted on our local datacenter, and potentially Google Drive and GitHub. See Paragraph~\ref{text:access_preservation_plan} for further details.
        \item \textbf{When will the dataset be distributed?}
        \newline Upon acceptance.
        \item \textbf{Will the dataset be distributed under a copyright or other intellectual property (IP) license, and/or under applicable terms of use (ToU)?} 
        \newline Yes, see Paragraph~\ref{text:license} for details.
        \item \textbf{Have any third parties imposed IP-based or other restrictions on the data associated with the instances?}
        \newline No.
        \item \textbf{Do any export controls or other regulatory restrictions apply to the dataset or to individual instances?}  
        \newline No.
    \end{enumerate}
    \item \textbf{Maintenance} 
    \begin{enumerate}
        \item \textbf{Who will be supporting/hosting/maintaining the dataset?} 
        \newline The authors of this work will be hosting and maintaining the dataset. 
        \item \textbf{How can the owner/curator/manager of the dataset be contacted (e.g., email address)?}  
        \newline All authors can be contacted through the email addresses listed on the first page of the paper.
        \item \textbf{Is there an erratum?}
        \newline No.
        \item \textbf{Will the dataset be updated (e.g., to correct labeling errors, add new instances, delete instances)?} 
        \newline The dataset is likely to be expanded in the future with more ground truth and to be tailored to specific applications, e.g. 2D background replacement for more effective deep learning.
        \item \textbf{Are there applicable limits on the retention of the data associated with the instances (e.g., were the individuals in question told that their data would be retained for a fixed period of time and then deleted)?}
        \newline No limits have been placed on the data. The only information provided to participants is that they can choose to opt out of public release of identifiable color information.
        \item \textbf{Will older versions of the dataset continue to be supported/hosted/maintained?} 
        \newline Older versions of the dataset will only be expanded upon, not entirely replaced. 
        \item \textbf{If others want to extend/augment/build on/contribute to the dataset, is there a mechanism for them to do so?} 
        \newline Not currently, as the dataset is large and must be hosted on our private datacenter at present. Collaboration may be possible in the future with substantial compression.
    \end{enumerate}
\end{enumerate}

\section{Participant Forms and Messages}
\label{sec:participant-forms}
In this section, we provide further detail for contact with participants and participant form responses. 

\paragraph{Participant Recruitment Message.}
\label{text:participant-recruitment}
Participants were recruited voluntarily via communication at the institution where the study was conducted. The recruitment message was posted on online collaborative work spaces (* denotes information that is redacted to preserve anonymity):
\begin{addmargin}[2em]{2em}

{\small\texttt{\textbf{Subject:} Participants sought for research study on understanding human preferences for handover parameters for safe human-robot collaboration}

\texttt{You are receiving this request as part of a *-wide announcement on recruitment for this study. We seek participants for a research study on understanding human preferences for handover parameters for safe human-robot collaboration. We are looking for individuals aged 18 to 99, with no known upper limb disability or injury that interferes with curling, grasping, and lifting, and that have no injury to fingers on either hand.}

\texttt{The data collection will take no longer than 2 hours. Participants will complete a demographics questionnaire and take part in a set of experiments involving interacting with 68 objects. }

\texttt{Participants will be recorded using Azure Kinect cameras and Point Grey Blackfly S cameras to gather data on human body posture and contact regions on objects. The study will enable us to design algorithms for robots that are aware of human handover preferences, so as to ensure safe human-robot collaboration. }

\texttt{The safety of all participants in this study is of paramount importance. We request that all subjects take a COVID-19 test within the 3 days prior to their data collection session. If you are unable to take a test within this window, we will provide you with one.}

\texttt{To remain in compliance with CDC, state, and institutional safety regulations for COVID-19, participants should not have left the * County area up to 14 days prior to your session. Participants will be recruited if they do not exhibit the following symptoms and have not exhibited them for 14 days: fever or chills, cough, shortness of breath or difficulty breathing, fatigue, muscle or body aches, headache, new loss of taste or smell, sore throat, congestion or runny nose, nausea or vomiting, and diarrhea.} 

\texttt{To reach us for participation in this study, please email * * at *@*.*. If you wish to opt-out of follow-up emails, please respond with a note stating that you do not want to receive future emails about participating in the study. The * IRB approval number for this study is * and the contact information for the * IRB office is * via email, and (***) ***-**** via phone.}}
\end{addmargin}

\paragraph{Participant Arrival Message.}
\label{text:participant-arrival}
Upon arrival for their data collection session, participants were read the following by an experimenter:

\begin{addmargin}[2em]{2em}
\textit{Hello, my name is (research personnel) and we are conducting a research study on understanding human preferences for handover parameters for safe human-robot collaboration.} 

\textit{The purpose of this study is to understand the relationship between object form and function, and human preferences for handover parameters such as where you hold an object, what orientations and distances you prefer an object being handed to you, and at what point do you prefer that an object be released upon handover. Our study will help to design robots that are aware of human handover preferences, to ensure safe human-robot collaboration in home and work environments, for example, safe assistive robots to help older adults.}

\textit{Today you will first read the informed consent form, and then take part in an experiment where you will lift these 68 objects (shown), one by one, and hand them to your partner. At the start, one of you will be assigned the role of hander and the other will be assigned the role of receiver. The hander will be asked to give the object to the receiver, and then you both will fill out a response about the handover on these forms (clipboard). Please stow the clipboards next to your chairs while the capture is taking place. Half of the way through the session we'll ask you to switch the hander and receiver roles. For some of the handover interactions, you may be asked to wear blue nitrile gloves.}

\textit{As mentioned, you will follow each handover interaction with a response on the paper forms. With this response you are to indicate your level of comfort with the interaction that just occurred on a scale of 1 ("not comfortable at all") to 7 ("the most comfortable - a perfect handover"). You can think of a rating of 1 as representing a handover that was barely complete, where you may be forced to use an uncomfortable grasp, the timing is off, and/or the location or orientation of the object is not preferable. A comfort rating of a 7 should represent a handover that cannot be improved; everything was done naturally in your opinion.}

\textit{While handing objects back and forth, you will be recorded at all times using these 4 Azure Kinect depth sensors and Point Grey Blackfly S high speed color cameras that allow us to capture information about your body and hand skeleton while you perform the grasp and allow us to understand where and how you hold objects. The maximum weight of any object is no more than 8 lb which is about the same as a gallon of milk. Most objects are no more than 2-3 lb in weight, and they are all objects you may use in your home or office.}

\textit{Please remember not to squeeze any object too hard because some of the objects are fragile. Just try to grasp the objects naturally. We also ask that you cover over clothing that is similar in color to the blue gloves or the green curtains (if necessary, we have neutral-colored shirts available). Finally, try to avoid bumping the camera frame or moving the chairs or table during captures. We'll take a break half way through when we switch roles so you can stand up and stretch a little.}

\textit{For your safety, we are asking that you adhere to all safety and distancing regulations to minimize contamination.} 

\textit{Your name will be removed from our dataset so as to not be associated with your data. From this point forward, all of your data will be identified by your 5-digit ID number. Please hold onto this card for the remainder of the study or in case you would like to contact us in the future. If at any point you feel any discomfort and wish to stop the study, please let me know, and we will stop the study immediately. You have the right to opt out of this study at any time you choose, and your data will immediately be erased.}

\textit{If you have any questions, I am happy to answer them now.}
\end{addmargin}

\begin{figure}
    \centering
    \includegraphics[width=\linewidth]{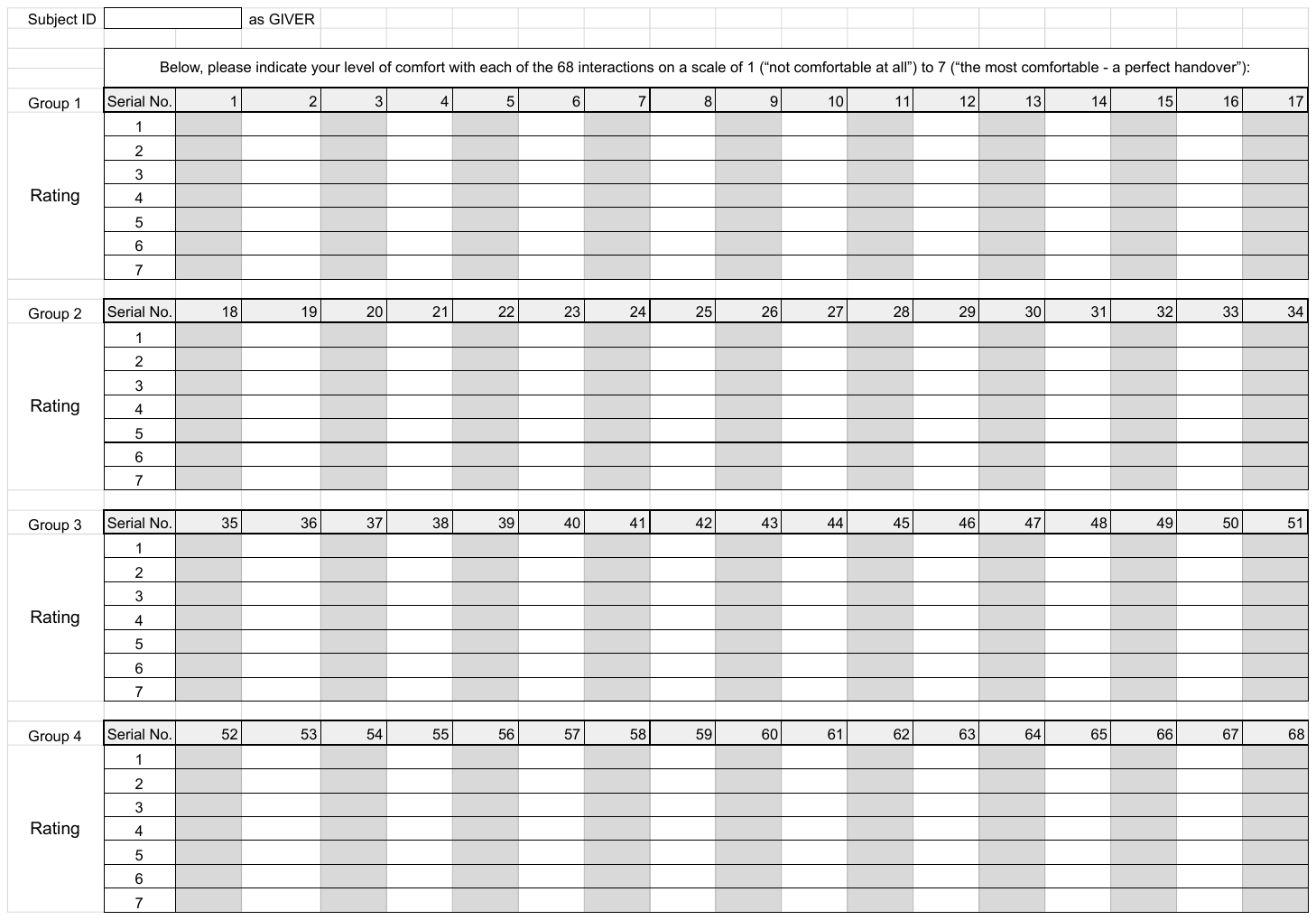}
    \caption{Template of the form completed by participants during a data collection session. The header changes depending on whether a participant is given the role of GIVER or RECEIVER.}
    \label{fig:sub_paper_form}
\end{figure}

\paragraph{Participant Paper Forms.} 
Participants completed the form in Figure~\ref{fig:sub_paper_form} to provide feedback on how comfortable they felt with each interaction. After every handover interaction, the participant marks their level of comfort in the column that corresponds to the interaction that happened immediately prior. The column in each group labeled with the numbers 1-7 displays the rating represented by each row, and the participant marks a single box in each column to provide their level of comfort with each interaction.

\begin{figure}
    \centering
    \includegraphics[width=\linewidth]{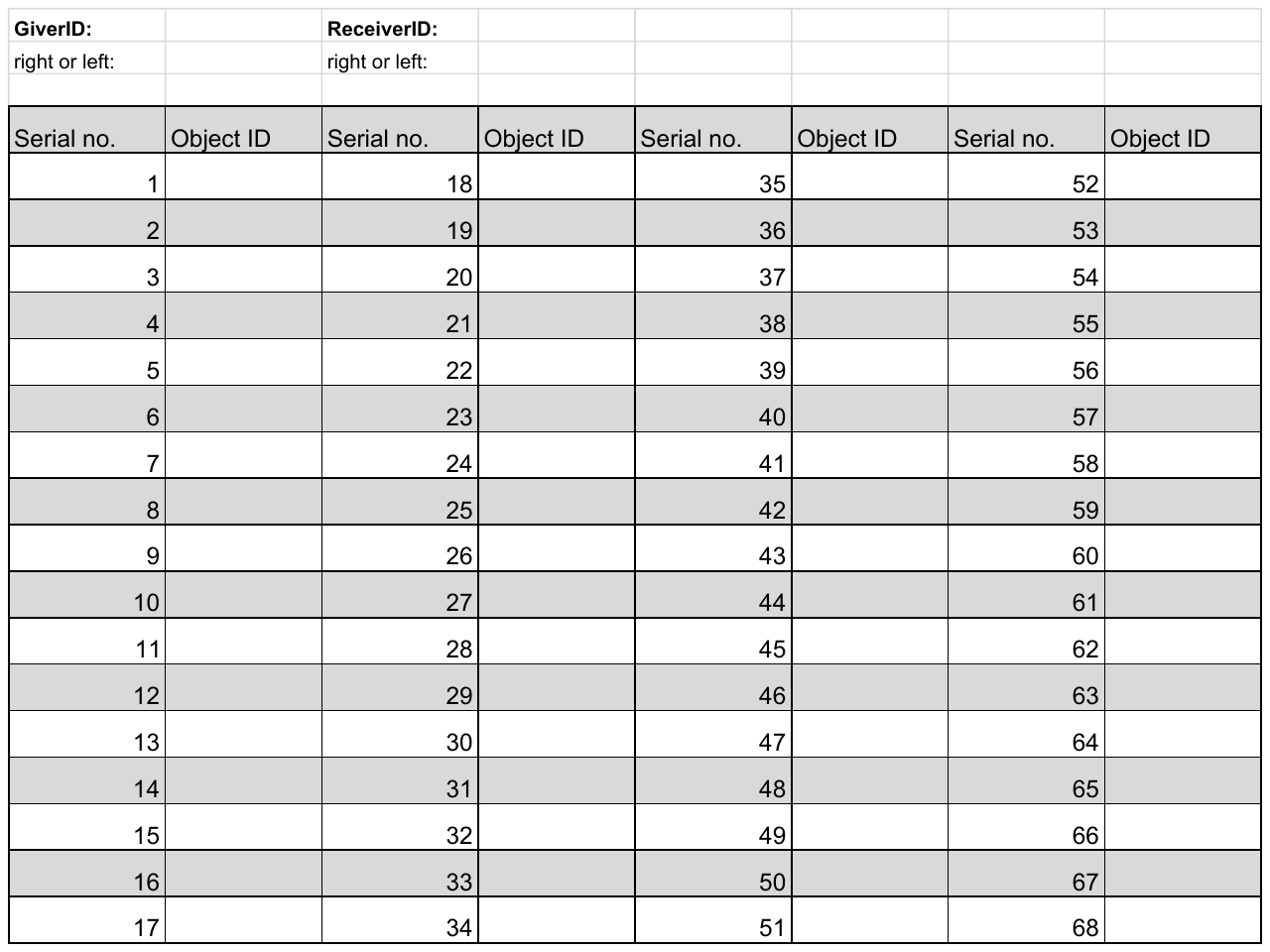}
    \caption{Template of the form completed by experimenters during a data collection session.}
    \label{fig:ex_paper_form}
\end{figure}

\paragraph{Experimenter Forms.}
Experimenters completed the form in Figure~\ref{fig:ex_paper_form} during the data collection session. The giver and receiver IDs are recorded along with their seating position. Since objects were randomly selected, experimenters recorded the object ID used for each interaction. The serial number columns represent the interaction index.

\begin{figure}
    \centering
    \includegraphics[width=\linewidth]{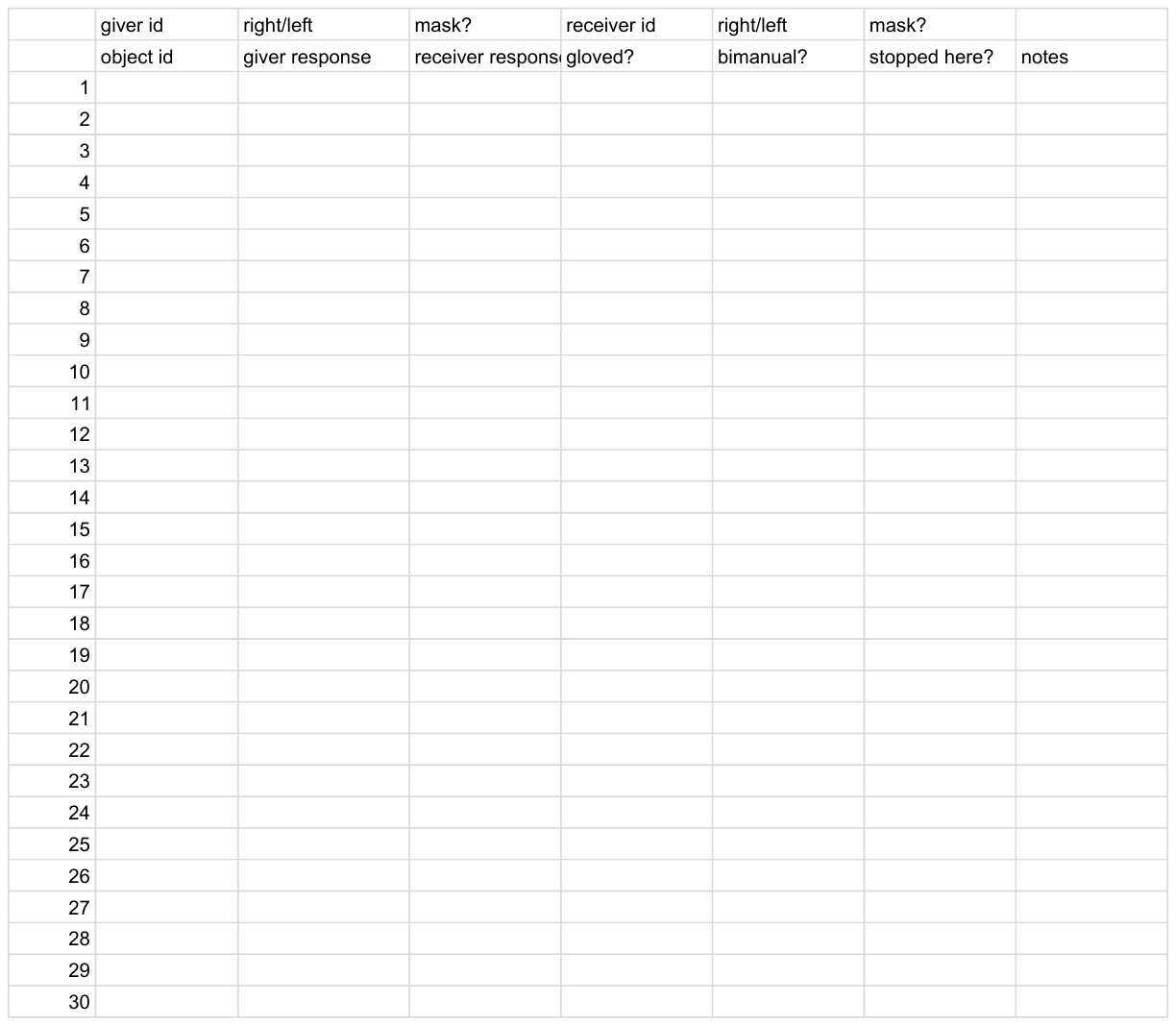}
    \caption{Template of the form completed by experimenters after a data collection session. Note that the form is truncated in length for display purposes, but the actual form extends to serial index 68.}
    \label{fig:paper_form}
\end{figure}

\paragraph{Paper Form Digitization Template.}
\label{text:paper-form-digitization}
All information written in the experimenter and participant paper forms is digitally entered into the form shown in Figure~\ref{fig:paper_form} by experimenters after each data collection session. To ensure the quality of the digitization by experimenters, code was used to validate the forms, which inspected object IDs, glove count, giver and receiver IDs, and comfort ratings.

\section{Data Capture System}
\label{sec:DataCaptureSystem}

In this section we provide additional detail about the capture system used for HOH.

\begin{figure}
    \centering
    \includegraphics[width=\linewidth]{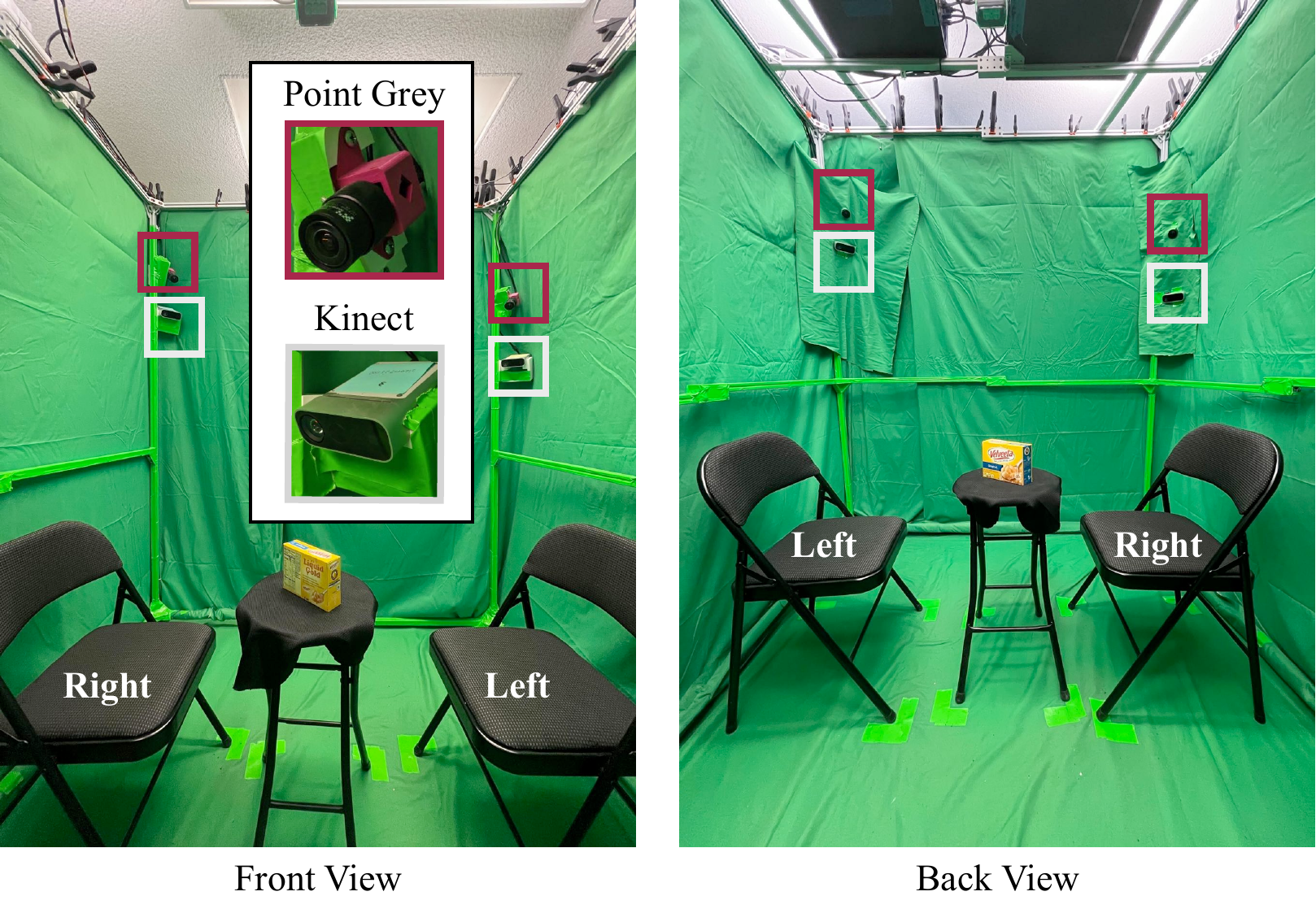}
    \caption{Capture setup showing placement of cameras and seating.}
    \label{fig:setup}
\end{figure}

\begin{figure}
    \centering
    \includegraphics[width=\linewidth]{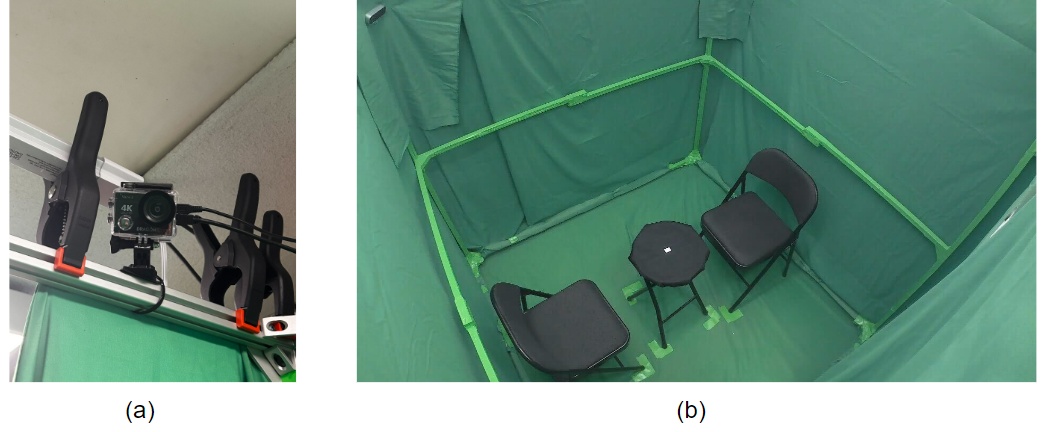}
    \caption{(a) DragonTouch camera that provides (b) a live feed into the capture environment.}
    \label{fig:dragontouch}
\end{figure}

\paragraph{Camera setup.}
Our data capture setup consists of a 1.7m $\times$ 1.7m $\times$ 2.0m T-slot frame rig with 4 Microsoft Azure Kinect RGB-D sensors and 4 FLIR Point Grey Blackfly S high-speed color cameras. The Kinect cameras are configured such that they record color (1920x1080 pixels) and depth (640x576 pixels) images at 30 frames per second (FPS). The Point Grey cameras have a 2.8-10 millimeter lens and record 60 FPS color (1440x1080 pixels) images. All extracted color images are stored as .jpg files. All extracted depth images are stored as 16-bit .png files. Kinects and Point Grey cameras are rigidly mounted on each of the corners of the capture system using custom-fabricated 3D-printed mounts at about 4 feet above the ground, directly pointed at the table. 1 Kinect/Point Grey pair is mounted on each vertical edge of the frame, 2 pairs at the front of the rig and 2 at the back as shown in Figure~\ref{fig:setup}. A table and two opposing chairs are located in the capture environment, and cameras are pointed to enable full capture of the handover space and the face, hands, and body posture of both participants. The Kinect depth sensors perform optimally with a target that is between 1.5 and 4 feet away from the sensor, and are mounted on the capture system corners to ensure that the center of the table is in the middle of that range. The Point Grey cameras have no such restriction, and are mounted above the Kinects on the corner columns, pointing directly at the table and configured to have as much of the scene in focus as possible. A Dragon Touch camera is mounted at the top of the system to provide a live video feed to the experimenters outside the capture space, which is not recorded, and is shown in Figure~\ref{fig:dragontouch}. All Kinects are administrated by a high-performance computer which commands the cameras and coordinates all recorded data and transfers it to the long term storage. All Point Grey cameras are connected to another similar computer with the same purpose. This configuration of sensors, control computers, and hardware has been tested over varied recording lengths of up to 15 minutes, and consistently yields frame drop amounts below 0.2\%.

\begin{figure}
    \centering
    \includegraphics[width=\linewidth]{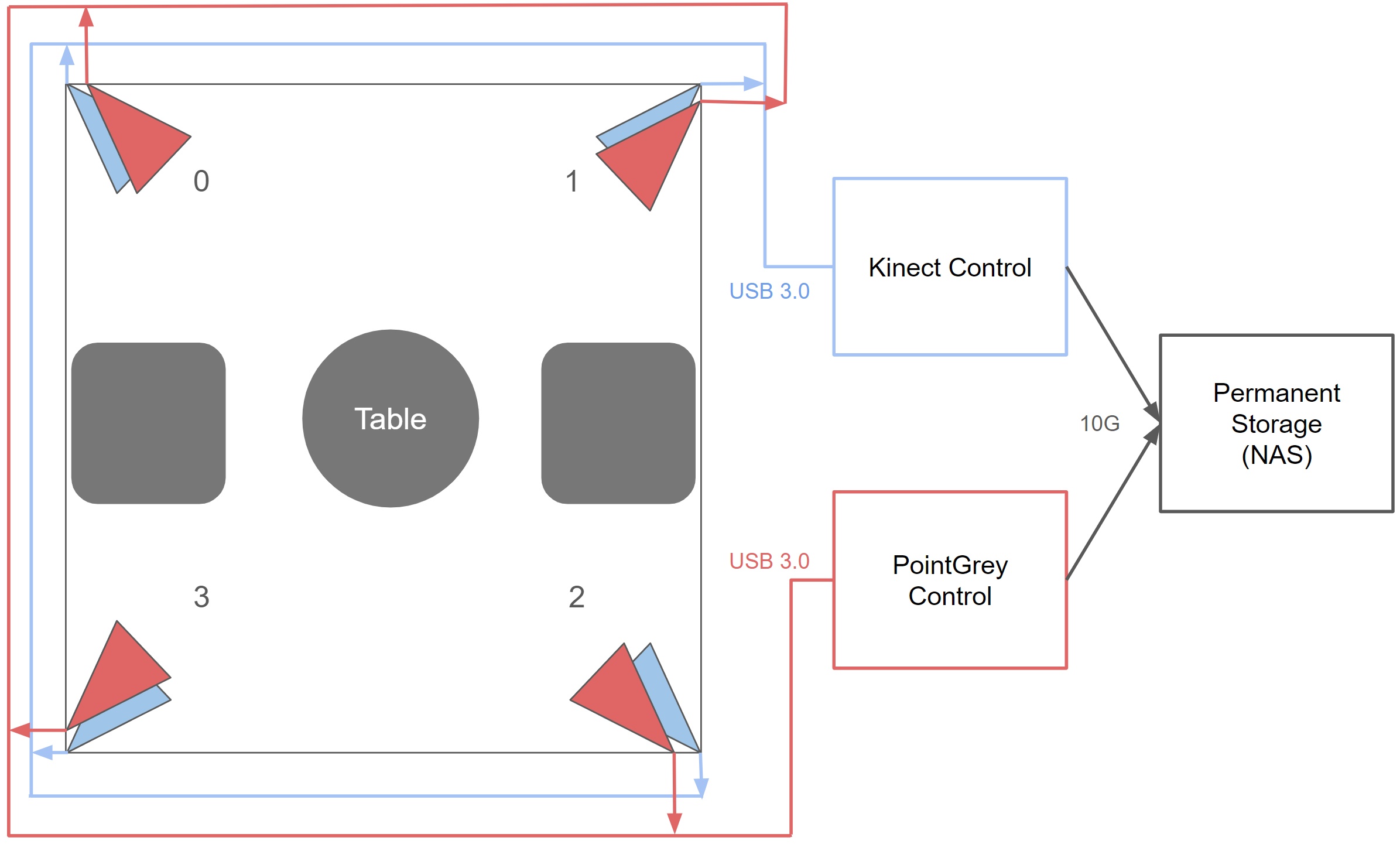}
    \caption{Networking diagram with Point Grey cameras in red and Kinect cameras in blue. Both camera types are connected to the respective control computers via USB 3.0. The control computers are connected to the NAS, where all data is stored.}
    \label{fig:networking}
\end{figure}

\paragraph{Networking.}
\label{text:networking}
The Kinect control computer is comprised of an Intel Core i9-9900k 8-core CPU and a GeForce RTX 2060 Super GPU in order to effectively support 4 Kinects recording simultaneously, and 2 NVMe and 3 solid state hard drives to enable writing of the large volume of recorded data. The Point Grey control computer uses a Ryzen Threadripper 1900x 8-core CPU and GeForce GTX 1080ti GPU, as the Point Grey sensors do not utilize the GPU. The Point Grey control computer also makes use of 3 NVMe and 2 solid state hard drives to accommodate the data writing in real time. Kinects and Point Grey cameras are connected via USB 3.0 to their control computers. Both sensor control computers make use of a separate PCIe USB expansion unit to allow for distributed use of motherboard lanes when processing data. Both sensor control computers are connected to an administrative DHCP server, hosted on a 576TB network-attached storage device (NAS). The NAS facilitates communication between computers and serves as long-term storage for data collected by the system. Connections to the NAS are 10-Gigabit in order to alleviate data transfer bottlenecks. Figure~\ref{fig:networking} illustrates the inter-device connections that form the capture system network.

\paragraph{Data Synchronization.}
Color, IR, and depth frames from the same Kinect are hardware synchronized, though frames from different devices are not synchronized to each other. We use the flash of a bright green light in each recording to identify a time step in the color images. This light flash is automatically detected using an image differencing technique that examines computes the average green intensity of a specific region of the images. This technique automatically detects the significant inter-frame green intensity change when the light deactivates. The frame where the light turns off for the last time is automatically detected on each color camera, and this frame is considered to be the first frame of the recording. The frame index of the new first frame is propagated to the depth and IR image streams from each Kinect color image stream. Synchronization was validated for every recording through manual viewing of an image tile containing images from each camera that show the 3-image sequence before, during, and after the synchronization light turned off. If the image tile was not strictly in this format, which happened in less than 15\% of cases, the synchronization offsets were manually corrected by adjusting the index of the new first frame to the point where the light actually shut off for each color image stream.

\begin{figure}
    \centering
    \includegraphics[width=\linewidth]{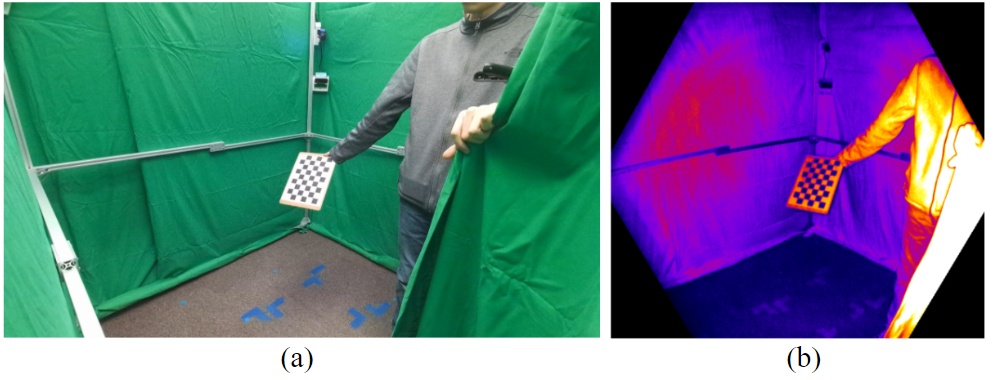}
    \caption{View of the calibration from the corner 0 (a) Kinect color and (b) Kinect infrared sensors.}
    \label{fig:calib_board}
\end{figure}
\paragraph{Sensor Calibration.}
\label{text:sensor_calib}
We use checkerboard camera calibration similar to the method of Zhang~\cite{zhang_calibration} to obtain sensor intrinsic parameters and extrinsic parameters that relate the poses of pairs of sensors. The checkerboard calibration target used was a black and white 6x9 checkerboard printed on paper. The target was fixed to a rigid back plate to ensure that it would always remain flat. The squares have a side length of 30 millimeters. There is significant contrast between the white and black squares on the calibration target visible in color and Kinect IR images. The Kinect IR images are used to calibrate the Kinect depth sensor. We calibrate each color sensor to the Kinect depth sensor on each corner. We calibrate pairs of depth sensors together to traverse between the capture system corners. This scheme allows for transformation of data from any sensor to any other sensor in the capture system through a composition of at most 3 extrinsic transformations. In each calibration operation, one person moves within the capture system while constantly adjusting the pose of the checkerboard in a manner that the sensor(s) being calibrated can see such that the checkerboard spans the entire field of view and space of poses that are possible within the camera view. This process must be completed thoroughly in order to achieve an accurate estimation of sensor intrinsic and extrinsic parameters. The intrinsic and extrinsic parameters were computed using the Stereo Camera Calibrator app in MATLAB~\cite{MATLAB}.

The data collection phase was split into two major participant groups, each interacting with a distinct set of 68 objects. Sensor calibration was performed prior to the recording of each of these groups, as this was a natural break in the recording process where calibration was possible. Calibration was performed multiple times to mitigate error from sensor drift over time. Each full-system calibration involved intrinsic parameter estimation for each individual sensor and extrinsic parameter estimation from multi-sensor calibration between pairs of sensors according to the above scheme. Over 200 images were used for each intrinsic calibration and over 1,000 were used for each extrinsic calibration in order to ensure less than 0.5 mean pixel error.

To obtain optimal point cloud scene reconstructions, we use Kinect depth sensor 0 as the reference coordinate frame and manually fine-tuned the extrinsic parameters to accurately transform the data from any depth sensor coordinate frame to the reference coordinate frame. This was done by applying the extrinsic transformations obtained from stereo calibration to transform the backprojected point clouds of depth sensor 1, 2, and 3 to the coordinate frame of depth sensor 0 of a single frame in one interaction in both groups. Using a tool designed with the vedo library~\cite{vedo_library} in Python, we manually rotated and translated the point clouds, aided by the Iterative Closest Point algorithm, to fine-tune the alignment with the reference coordinate point cloud. These fine-tuning transformations were saved and used when fusing the point clouds.

\section{Additional Details for Ground Truth Annotation and Data Processing}

In this section we provide additional details concerning the manual annotation processes and custom tools to obtain ground truth data originally mentioned in section 3 of the main paper. We also provide additional details concerning the implementation and execution of the post-processing mask tracking task introduced in section 3 of the main paper. 

\begin{figure}
    \centering
    \includegraphics[width=\linewidth]{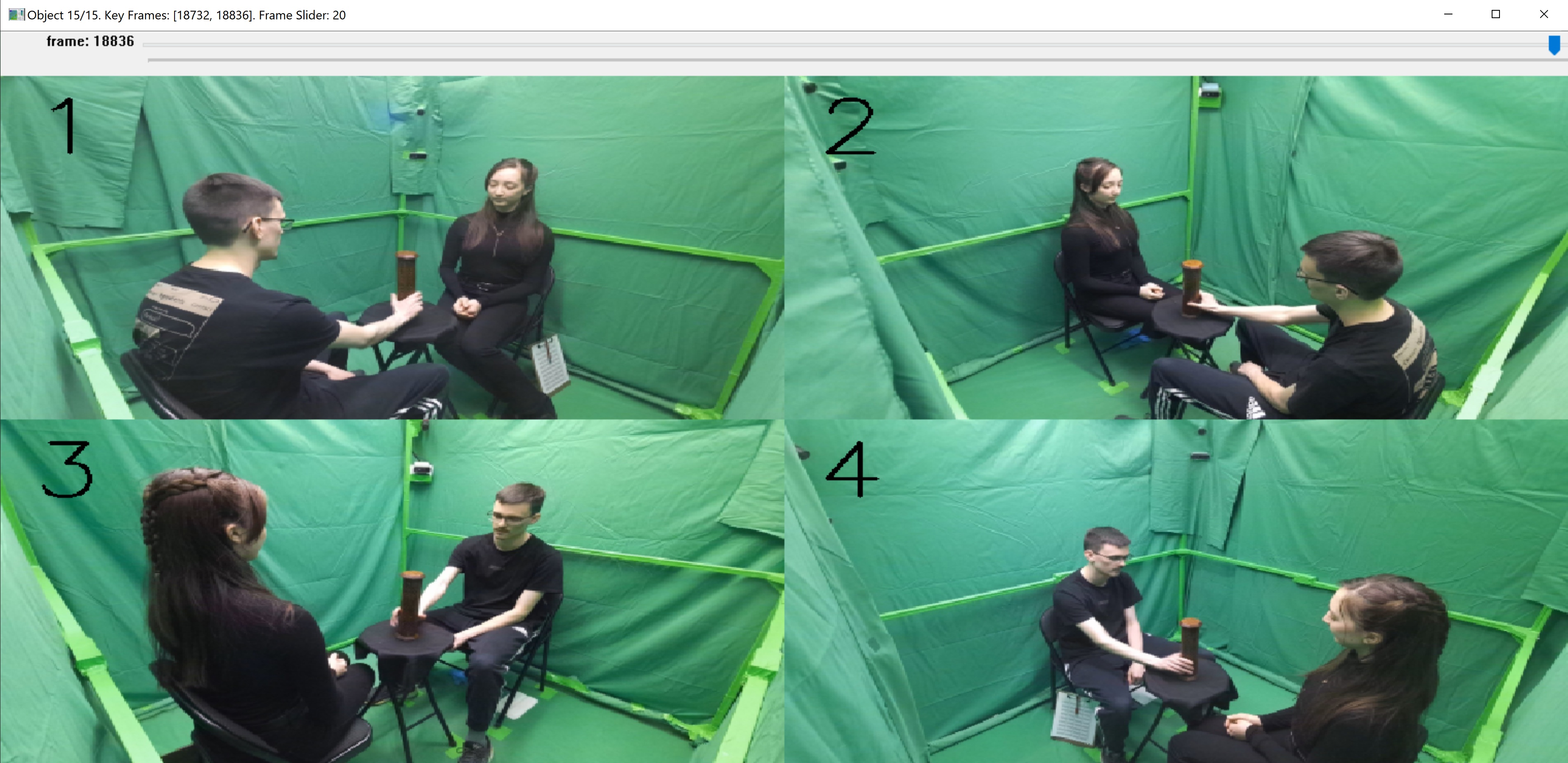}
    \caption{The annotation tool for selecting key events. Inputs are keypresses that correspond to temporal scrolling, frame selection, and saving.}
    \label{fig:kf_ann}
\end{figure}

\paragraph{Key Event Selection.}
Annotators used a tool created in-house to manually isolate 3 time points, referred to as key events, in each interaction that represent: first giver contact marking the grab portion of reach and grab phase called frame G, simultaneous giver and receiver grasp marking the middle region of the object transfer phase called frame T, and last receiver contact on the object marking the final part of the end of handover phase called frame R. The annotation tool, shown in \ref{fig:kf_ann}, uses the OpenCV~\cite{opencv_library} Python library to show the images and receive user input. The annotator selects a key event using keypresses, and they can scroll temporally through all images in the recording to select key events for all interactions. Once all key events are chosen, they are saved in .json files.

\begin{figure}
    \centering
    \includegraphics[width=\linewidth]{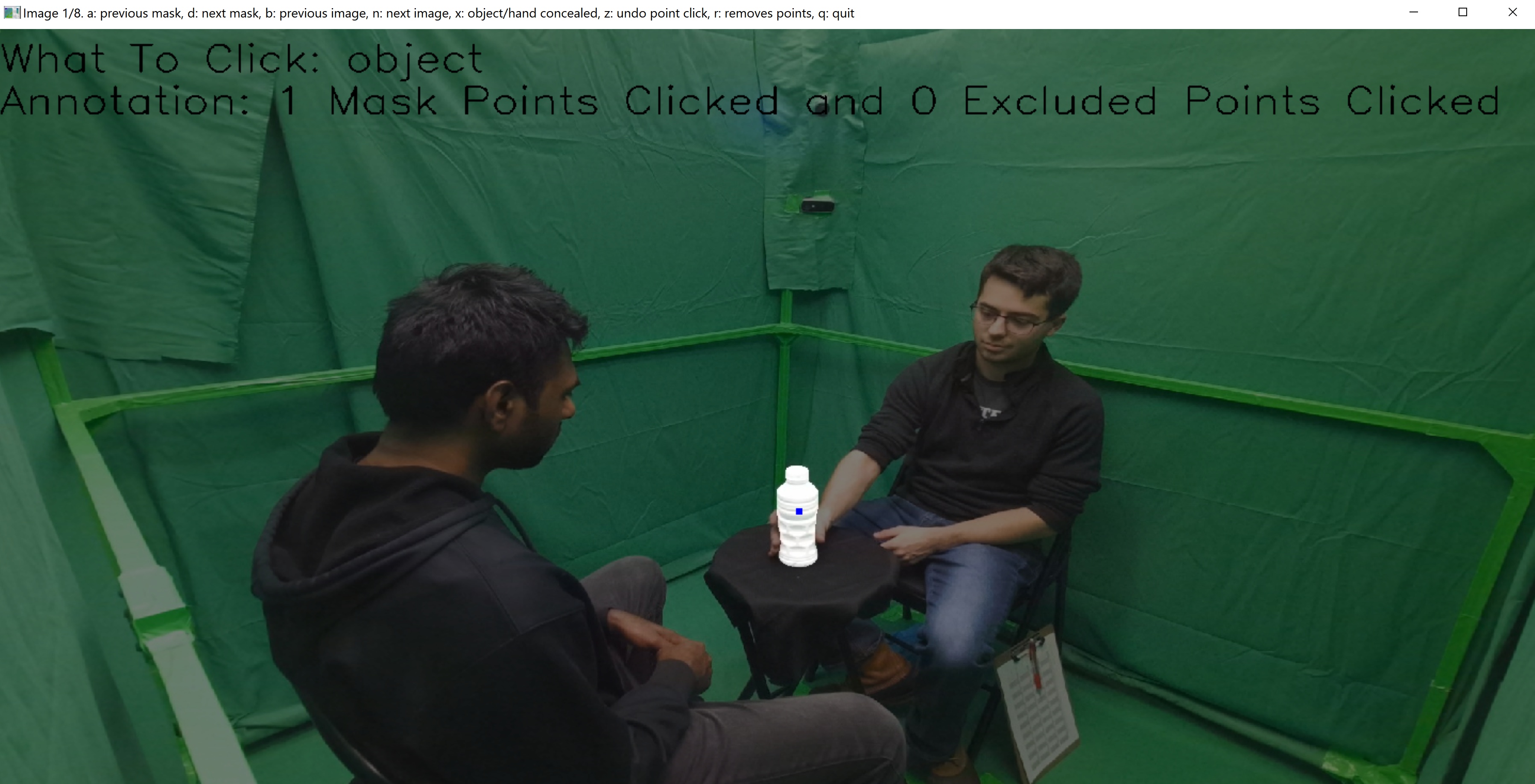}
    \caption{The annotation tool for selecting masks from the full image segmentation. Inputs are mouse clicks and keypresses which correspond to mask selection and saving, respectively.}
    \label{fig:seg_ann}
\end{figure}

\paragraph{Segmentation Mask Selection.}
For each key event for all 4 Kinect color cameras for each interaction, annotators use a custom tool, shown in \ref{fig:seg_ann} to select 3 masks from the set of approximately 60 scene segmentation masks: that of giver's hand, the object, and the receiver's hand. Note that the receiver's hand is not present in frame G and the giver's hand is not present in frame R, and all 3 targets are present in frame T. Using this tool, the annotator clicks on the giver hand, object, or receiver hand in the image to select the corresponding mask, which is highlighted by the program. If the mask is correct, the annotator continues on to select the next mask in the process. If the mask is incorrect, the annotator clicks more foreground and background points, using left and right clicking respectively, to refine the clicked mask. If no background points are marked, the clicked masks are considered correct. If both foreground and background points are marked, the image is fed back to SAM, specifically the SamPredictor initialized with the default model checkpoint, to generate a mask of only the hand or object, which is refined by the annotator input points to hone in on the correct shape. If the hand or object is occluded from the camera, it is marked as concealed. In any case, the selected giver hand, object, and receiver hand masks are heavily validated through a visual review of each color image with the selected mask overlaid onto it. In this validation step, the selected mask is either labeled correct or incorrect by a different annotator than the one who originally selected the mask for that frame. The approximately 15\% of masks that were labeled as incorrect were later fixed through the use of an interactive tool. All points clicked and mask indices located at those points are saved in .json files.

\begin{figure}
    \centering
    \includegraphics[width=\linewidth]{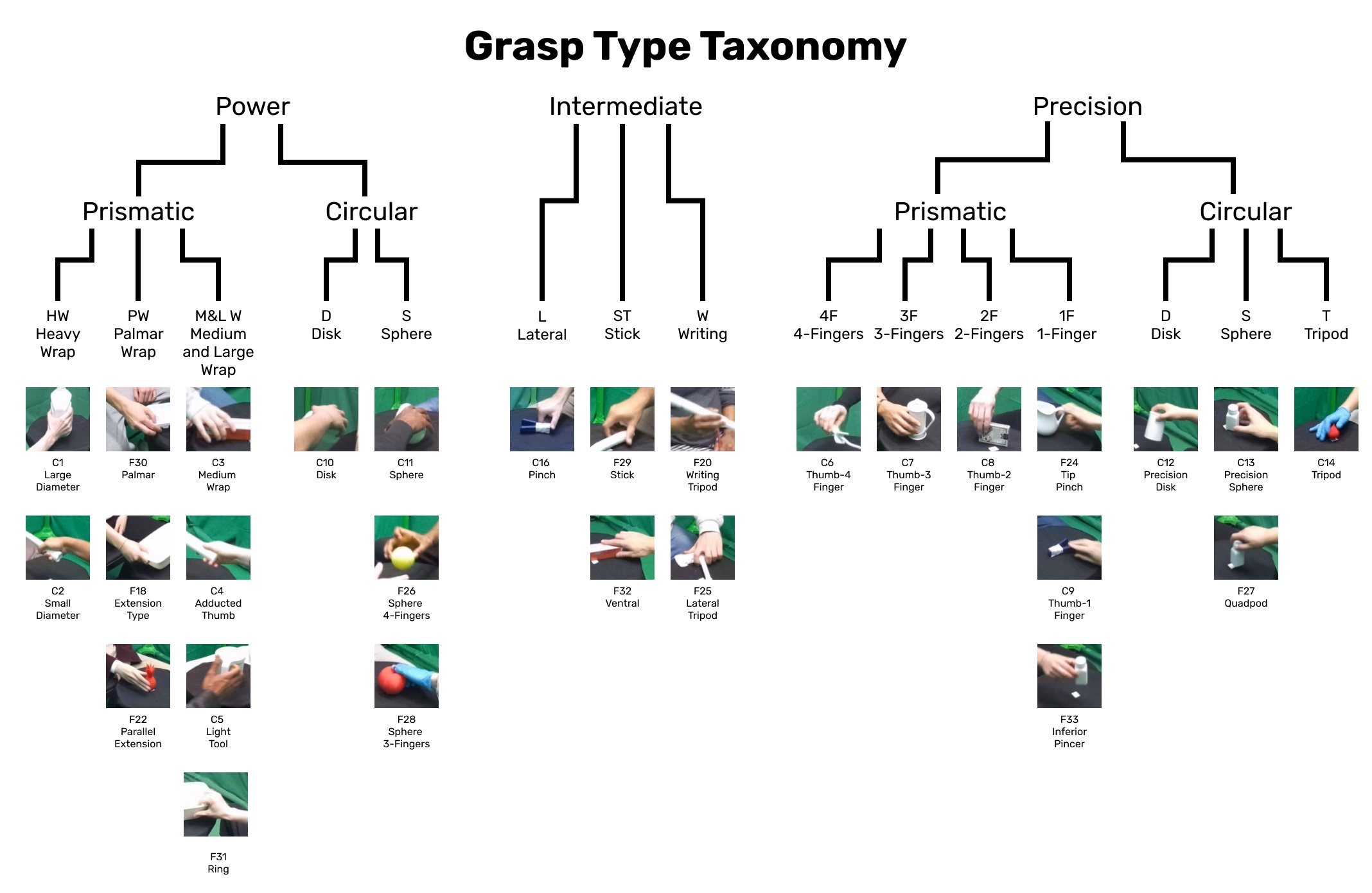}
    \caption{Grasp Type Taxonomy with image examples for all 28 grasp types. The grasp name is under each image.}
    \label{fig:grasps}
\end{figure}

\paragraph{Grasp Taxonomy.}
As discussed in Section 4 of the main paper, we categorized all object grasps as per the grasp taxonomy discussed in Cini et al.~\cite{cini2019choice}. Figure~\ref{fig:grasps} shows the breakdown of all 28 grasps with image examples. Power, the use of a power grip, Precision, the use of precision handling, and Intermediate, the use of both power and precision, are the 3 main grasp classifications. Both Power and Precision are broken down into Prismatic and Circular grasp types while Intermediate is broken down into Lateral, Stick, and Writing. Power Prismatic is further broken down into Heavy Wrap, Palmar Wrap, and Medium and Large Wrap. Power Circular has 2 categories, one for disk shaped-objects and the other for sphere-shaped objects. Precision Prismatic, on the other hand, has 4 categories based on the number of fingers used to grasp the object. Prismatic Circular also has an additional Tripod category along with the Disk and Sphere categories. Each lowest-level category has 1 to 4 grasp types that are named with the letter C or F and a number.

\begin{figure}
    \centering
    \includegraphics[width=\linewidth]{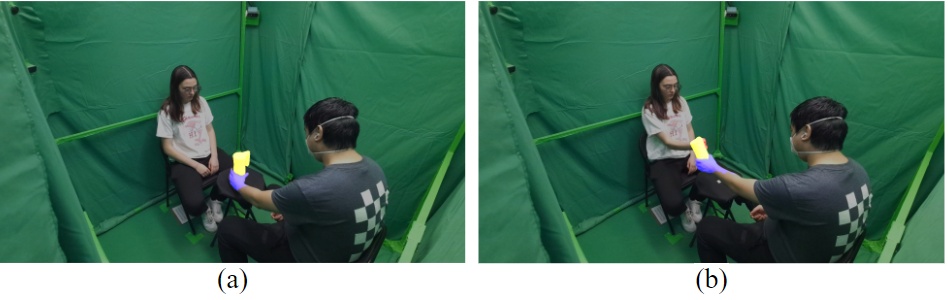}
    \caption{Hand and object mask tracking with the giver's hand in blue, the object in yellow, and the receiver's hand in red. The giver hand mask is increasing up the giver's arm from (a) Frame G, to (b) Frame T.}
    \label{fig:masks}
\end{figure}

\paragraph{Mask Tracking.}
The mask tracking approach is robust to enable intermediate frame tracks of hands or objects that may be occluded in a particular key event annotated mask. Track-Anything~\cite{yang2023track} requires an input mask and a set of images, e.g. for the giver hand track the input would be the Frame G giver hand mask and a list of successive frames from G through R, and for the object track the input would be the Frame G object mask and a list of successive frames from G through R. This approach is also effective in reverse-temporal order, e.g. for the receiver hand track where the input would be the Frame R receiver hand mask and a list of successive frames from R through G. In the event that a giver hand, object, or receiver hand is occluded in a ground truth annotated key event frame, we pass a different key event frame annotation to the tracking program. The likelihood of a particular entity being occluded in all annotated key event frames is low, as the entity and potential sources of occlusion move over time. A potential source for mask tracking error is due to the tendency of a track to slowly expand over successive frames, especially for hand masks when the participant was not wearing long sleeves, as shown in Figure~\ref{fig:masks}. All tracked masks for objects, giver hands, and receiver hands for single interactions are compiled in forward-temporal order, i.e. G to R order, and saved in a compressed .npz file.

\paragraph{Network and Training Details}
\label{sec:training_details}

For \texttt{o2gg} and \texttt{g2rg}, the hyperparameters of their base network PoinTr~\cite{yu2021pointr} were used as default, i.e., 300 epochs and learn rate of 0.0005, and a batch size of 24 was used as automatically determined by the PoinTr implementation. We used the PoinTr implementation as is. For \texttt{g2rt}, the learning rate was kept to the default of Informer~\cite{zhou2021informer} at 0.0001, and the batch size was increased to 64. The number of epochs were set to 1,000. Since we addressed spatial trajectory generation, only positional encodings were used for Informer, and temporal embeddings were eliminated. For \texttt{o2or}, batch size of 24 and learn rate of 0.001 were used. Due to the smaller network size, the training error declined rapidly, as such training epochs were set to 80. For \texttt{o2or} the original PointNet~\cite{qi2016pointnet} encoder was used to generate a 1,024 global feature vector, which was fed to a (1,024,128,64,4) multi-layer perceptron that yielded the x, y, z, and w components of the rotation quaternion.

\section{Computing and Training Details}
\label{sec:ComputingAndTrainingDetails}

\paragraph{Computing Resources and Annotation Time Estimates.}
Computing resources were used for the following non-experimental purposes. The extraction and synchronization processes were CPU-based, and ran for approximately 240 hours spread across 3 computers with AMD Ryzen 2700X CPUs. The full-scene Segment Anything Model (SAM)~\cite{kirillov2023segany} segmentation for all Kinect color images ran for approximately 3,000 hours spread across 9 NVIDIA 3090 and 2 NVIDIA 3090Ti GPUs. Intermediate frame entity mask tracking ran for approximately 450 hours spread across 5 NVIDIA 3090 GPUs. Point cloud processing were CPU-intensive, and ran for approximately 1,500 hours across 20 computers with AMD Ryzen 2700X CPUs. Annotators spent approximately 580 hours performing all annotations for the dataset.

\paragraph{Computing Resources for Experimental Results}

We trained and tested the neural networks for \texttt{o2gg} and \texttt{g2rg} using our own server with four (4) NVIDIA M40 GPUs, two (2) Intel Xeon E5-2640 v4 CPUs, and 128GB of RAM. We trained and tested the neural networks for \texttt{o2or} using two of our own servers. The first server had two (2) NVIDIA 3090 GPUs, two (2) Intel Xeon E5-2640 v4 CPUs, and 256GB of RAM. The second server was identical to the first, but had 128GB of RAM. We trained and tested the neural networks for \texttt{g2rt} using our own server with one (1) NVIDIA 3090, one (1) Intel i5-10600K CPU, and 64GB of RAM.

\section{Additional Outputs for Experiments}
\label{sec:additionalResults}

\begin{figure}
    \centering
    \includegraphics[width=\linewidth]{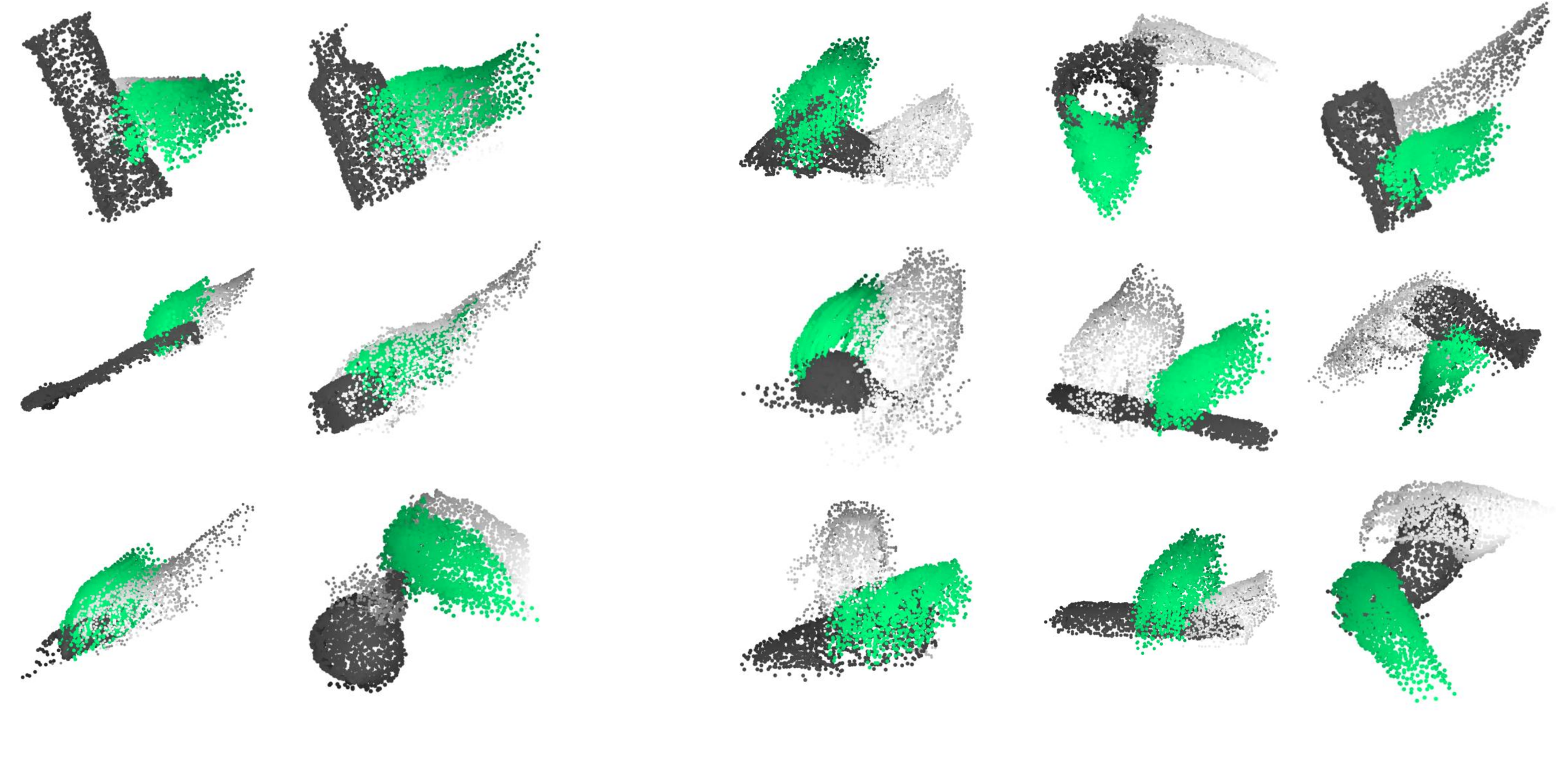}
    \caption{Additional results of \texttt{o2gg} using complete data. Predicted grasp in green versus GT grasp in light gray on input object in dark gray. Examples shown where grasp is close to GT (left) and grasp though deviating from GT is plausible (right).}
    \label{fig:suppcompleteO2GG}
\end{figure}

\begin{figure}
    \centering
    \includegraphics[width=\linewidth]{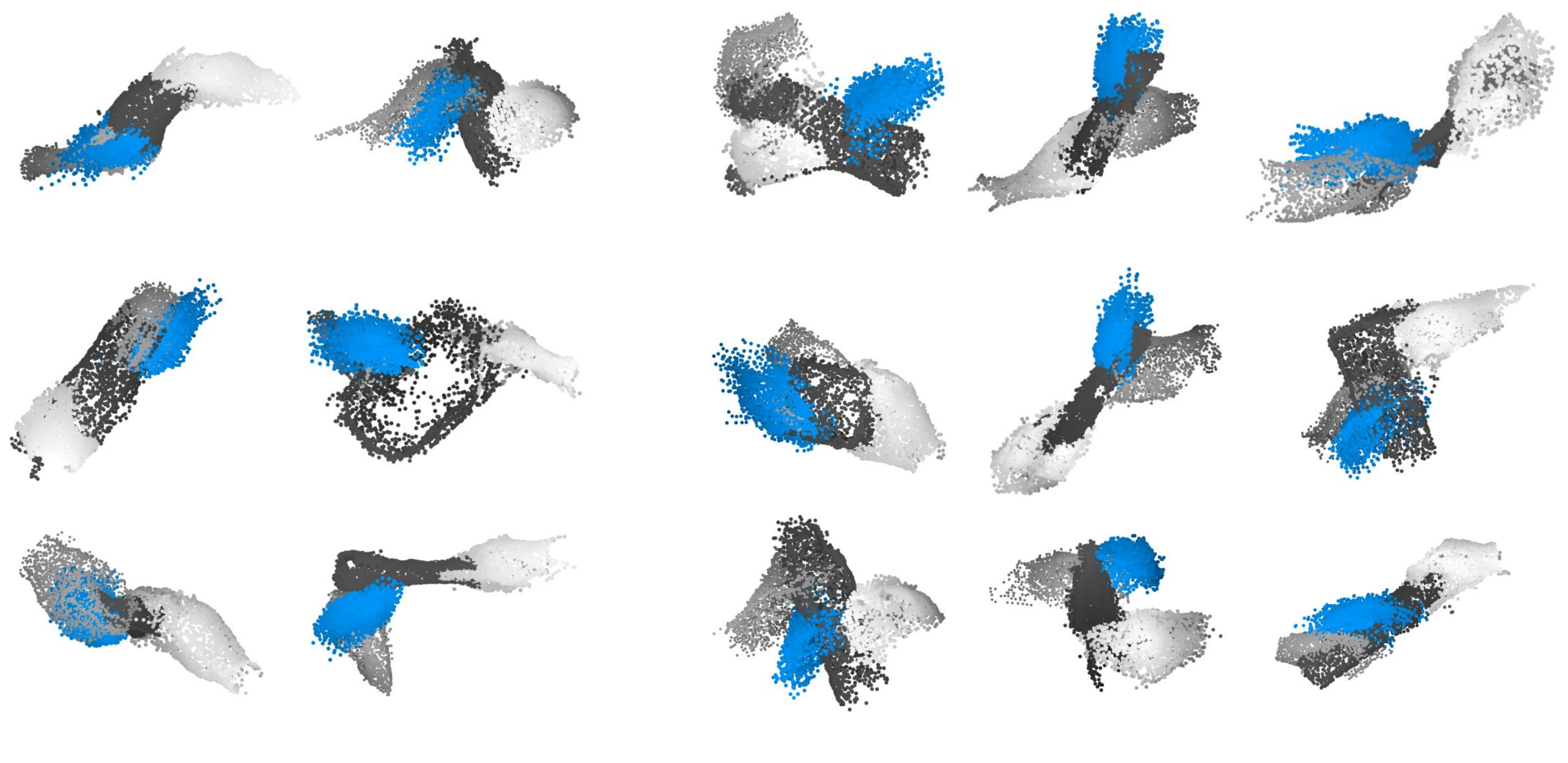}
    \caption{Additional results of \texttt{g2rg} using complete data. Input object and giver are in dark and light gray. Predicted receiver grasp in blue versus GT receiver grasp in medium gray. Examples shown where grasp is close to GT (left) and grasp though deviating from GT is plausible (right).}
    \label{fig:suppcompleteG2RG}
\end{figure}

\begin{figure}
    \centering
    \includegraphics[width=\linewidth]{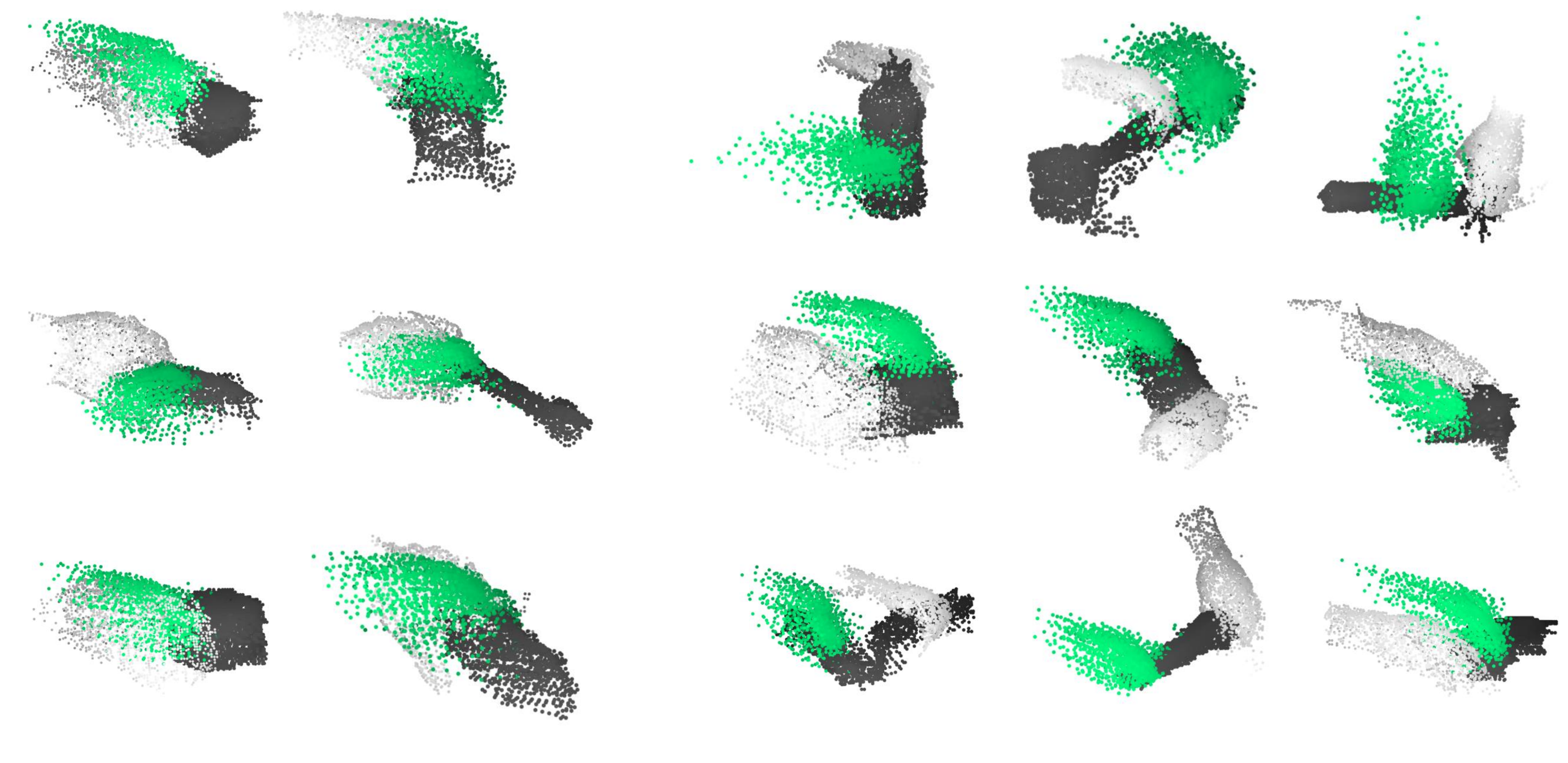}
    \caption{Additional results of \texttt{o2gg} using partial data. Predicted grasp in green versus GT grasp in light gray on input object in dark gray. Examples shown where grasp is close to GT (left) and grasp though deviating from GT is plausible (right).}
    \label{fig:supppartialO2GG}
\end{figure}

\begin{figure}
    \centering
    \includegraphics[width=\linewidth]{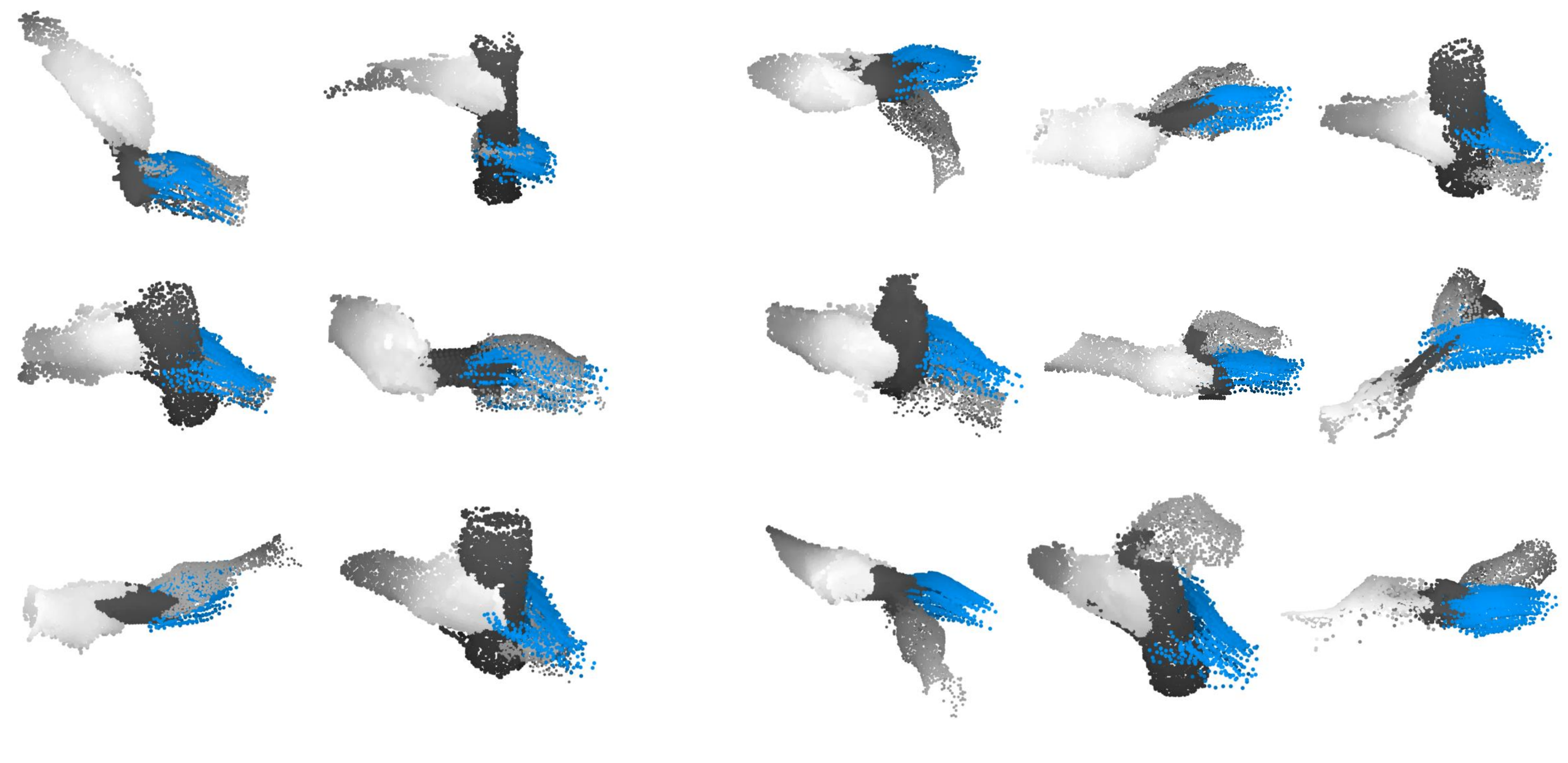}
    \caption{Additional results of \texttt{g2rg} using partial data. Input object and giver are in dark and light gray. Predicted receiver grasp in blue versus GT receiver grasp in medium gray. Examples shown where grasp is close to GT (left) and grasp though deviating from GT is plausible (right).}
    \label{fig:supppartialG2RG}
\end{figure}

\begin{figure}
    \centering
    \includegraphics[width=\linewidth]{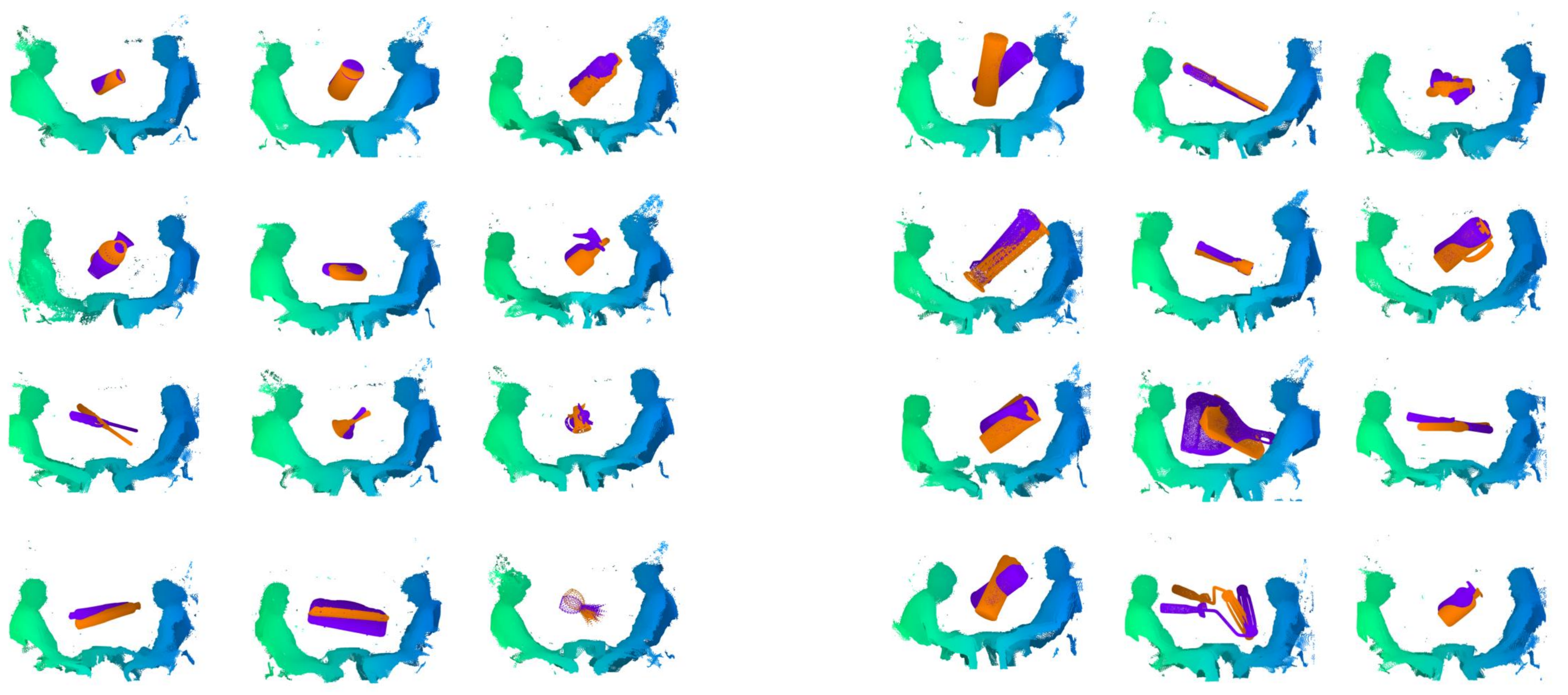}
    \caption{Additional results of \texttt{o2or} using complete data. Predicted orientation in purple versus GT orientation in orange, with examples shown where orientation is close to GT (left) and orientation though deviating from GT is plausible (right).}
    \label{fig:suppcompleteOR}
\end{figure}

\begin{figure}
    \centering
    \includegraphics[width=\linewidth]{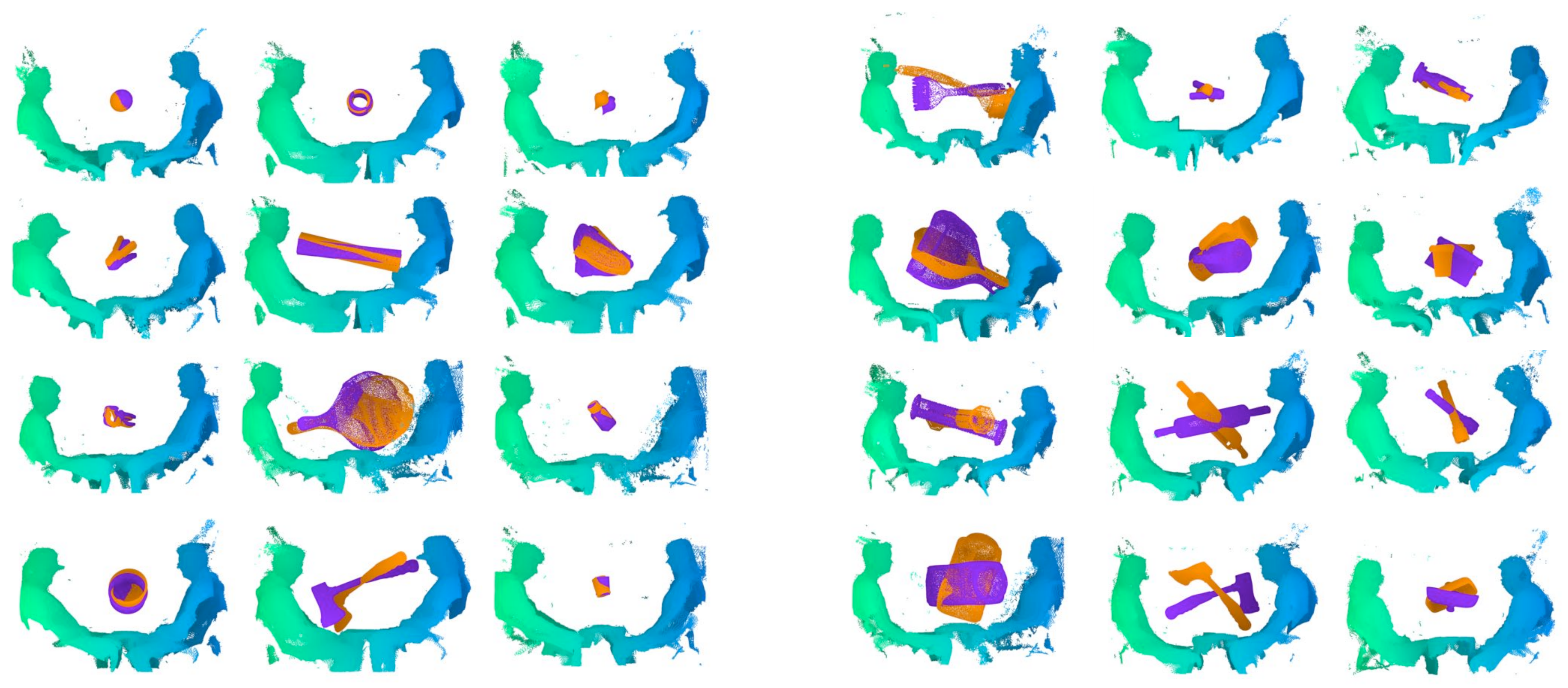}
    \caption{Additional results of \texttt{o2or} using partial data. Predicted orientation in purple versus GT orientation in orange, with examples shown where orientation is close to GT (left) and orientation though deviating from GT is plausible (right).}
    \label{fig:supppartialOR}
\end{figure}

Figures~\ref{fig:suppcompleteO2GG} through \ref{fig:supppartialOR} show further visual results of running \texttt{o2gg}, \texttt{g2rg}, and \texttt{o2or} on complete and partial data. The figures show examples of results that are close to GT on the left, and plausible outputs further from GT on the right.

\section{Creation of Object Dataset}
\label{sec:ObjectDataset}
We collected a set of 136 objects for use in this study. All objects are approximately table-scale, meaning that they could sit unsupported on a table surface. All objects are less than 2.5kg. All objects have at least one dimension that is less than 6 inches to enable unimanual human grasping.

116 objects were store-bought and the remaining 20 were 3D-printed from CAD models of miniatures and perishable items such as fruit. A variety of everyday use items, including 52 of our store-bought objects, have shiny or dark surfaces on which the Kinect's infrared time-of-flight sensor malperforms, yielding poor depth. We coated most of the 52 objects with white matte spray paint, and duct taped a small subset for which spray paint failed to stick. 

\begin{figure}[t!]
    \centering
    \includegraphics[width=\linewidth]{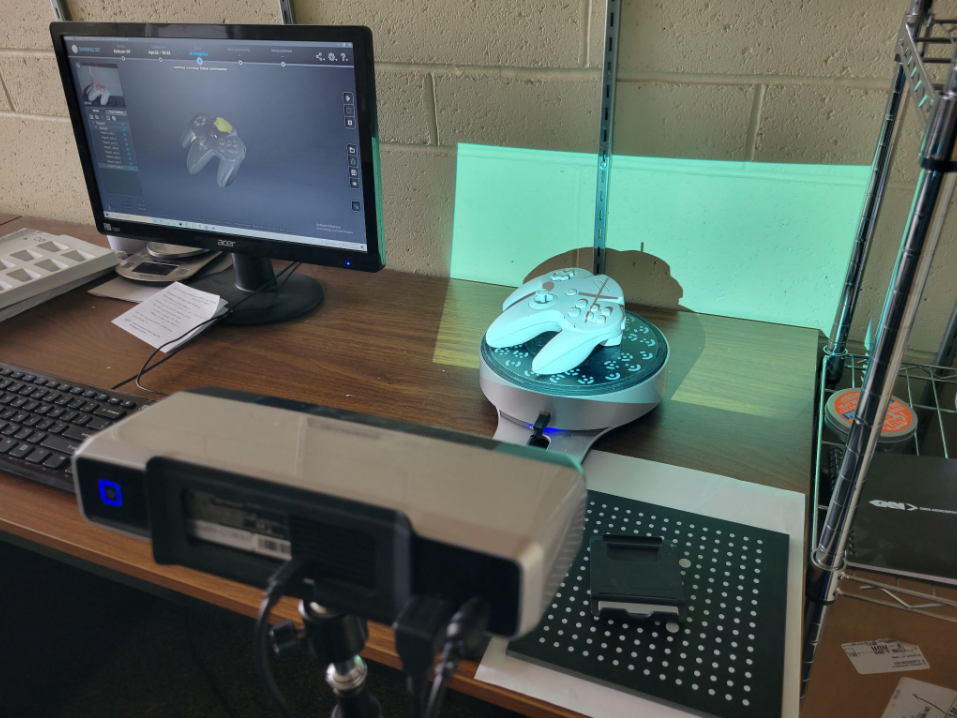}
    \caption{The Einscan-SP 3D scanner (bottom center) with object on turntable being scanned (center right) and EXScan S software compositing the scan (top left).}
    \label{fig:Einscan}
\end{figure}

\begin{figure}[h!]
    \centering
    \includegraphics[width=\linewidth]{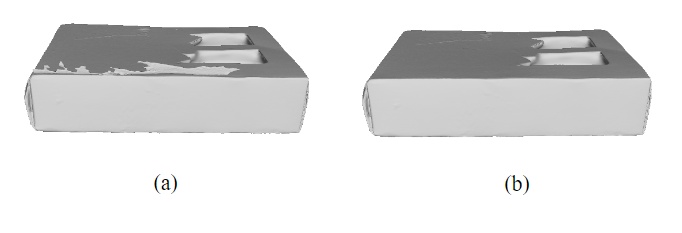}
    \caption{(a) Original mesh from scanner and (b) manually cleaned mesh.}
    \label{fig:orig_cleaned}
\end{figure}

We use an Einscan-SP 3D scanner with EXScan S to obtain high-fidelity meshes for the 116 store-bought objects as shown in Figure~\ref{fig:Einscan}. Meshes are bounding-box centered at the origin in their canonical orientation. Meshes are manually cleaned in Autodesk Netfabb to remove any additional parts or scan artifacts that are not present in the real object. Example original and cleaned meshes are shown in Figure~\ref{fig:orig_cleaned}. The cleaned mesh likely has holes, and is made watertight according to the approach of Stutz and Geiger~\cite{stutz2018learning}. We rotated each mesh to a manually determined standard orientation. For ease of use, the high-fidelity meshes are uniformly simplified down to 40,000 faces using quadratic decimation. We store metadata using the waterproof meshes prior to quadratic decimation. We weigh physical objects on a kitchen scale to obtain mass.

\begin{figure}
    \centering
    \includegraphics[width=\linewidth]{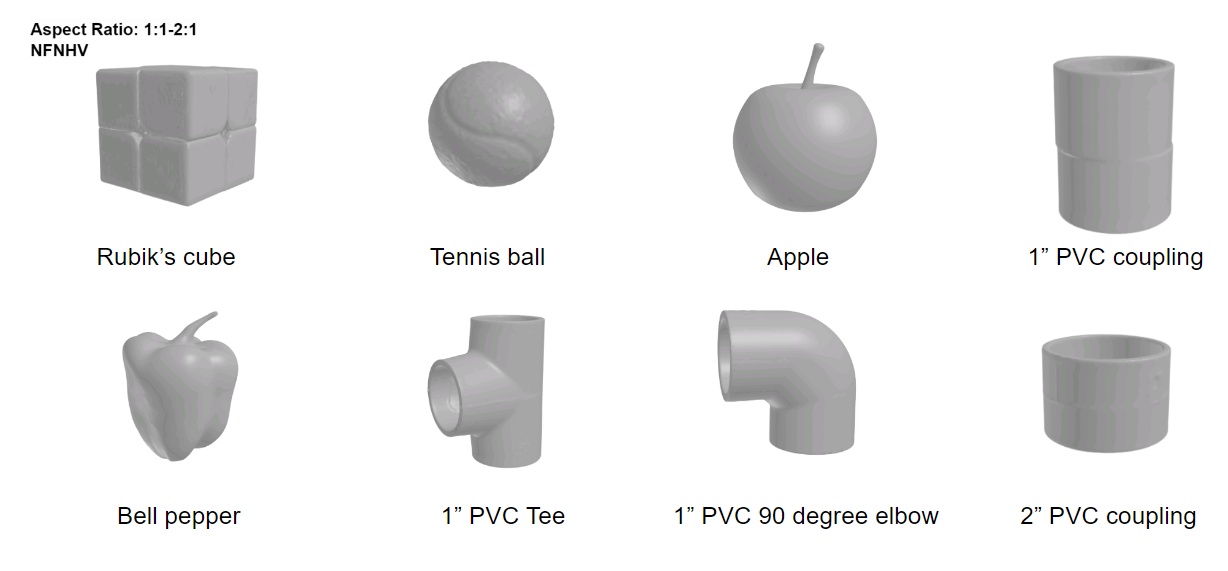}
    \caption{Objects from bin 1 which have the aspect ratio 1:1-2:1 and are NFNHV }
    \label{fig:rendergroup1}
\end{figure}

\begin{figure}
    \centering
    \includegraphics[width=\linewidth]{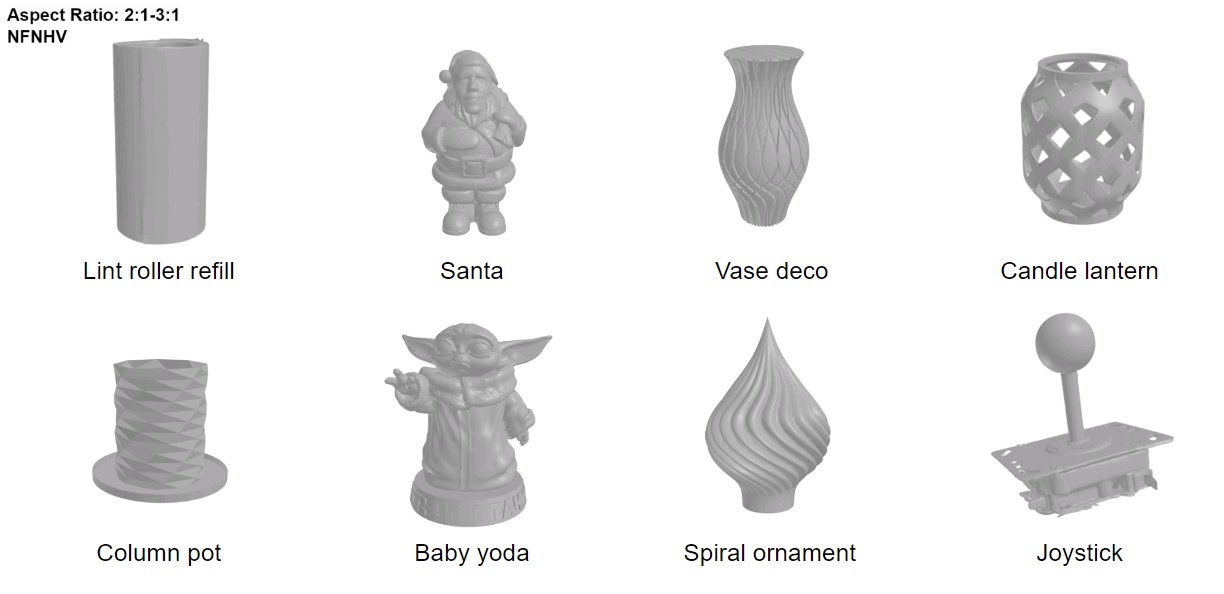}
    \caption{Objects from bin 2 which have the aspect ratio 2:1-3:1 and are NFNHV}
    \label{fig:rendergroup2}
\end{figure}

\begin{figure}
    \centering
    \includegraphics[width=\linewidth]{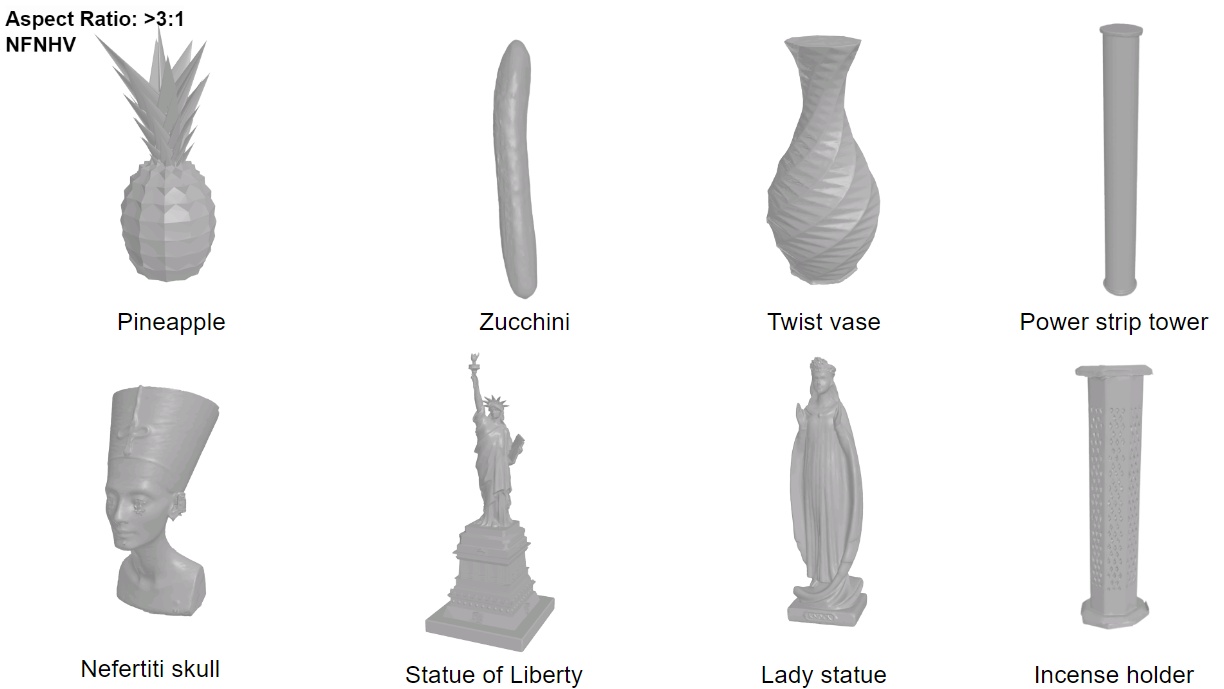}
    \caption{Objects from bin 3 which have the aspect ratio >3:1 and are NFNHV}
    \label{fig:rendergroup3}
\end{figure}

\begin{figure}
    \centering
    \includegraphics[width=\linewidth]{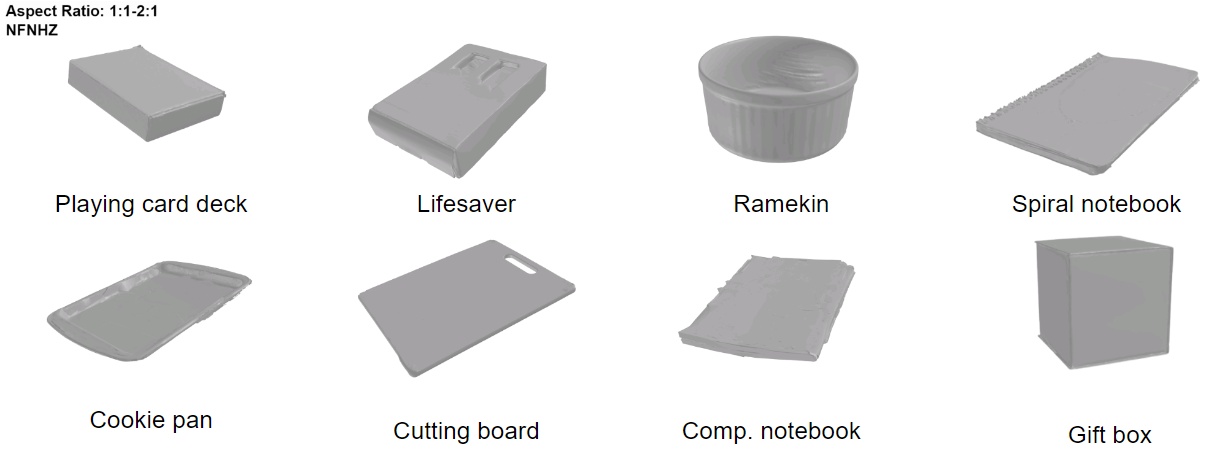}
    \caption{Objects from bin 4 which have the aspect ratio 1:1-2:1 and are NFNHZ}
    \label{fig:rendergroup4}
\end{figure}

\begin{figure}
    \centering
    \includegraphics[width=\linewidth]{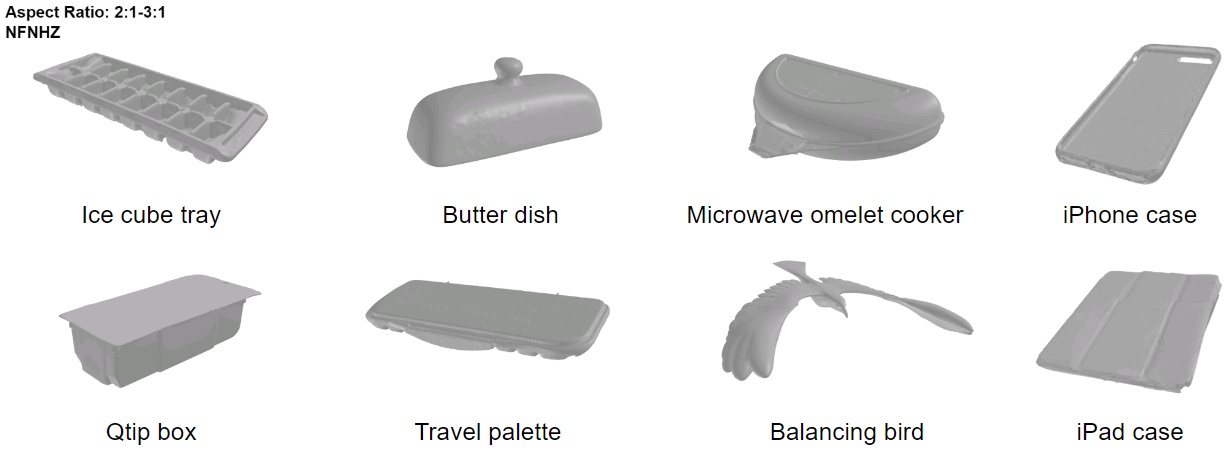}
    \caption{Objects from bin 5 which have the aspect ratio 2:1-3:1 and are NFNHZ}
    \label{fig:rendergroup5}
\end{figure}

\begin{figure}
    \centering
    \includegraphics[width=\linewidth]{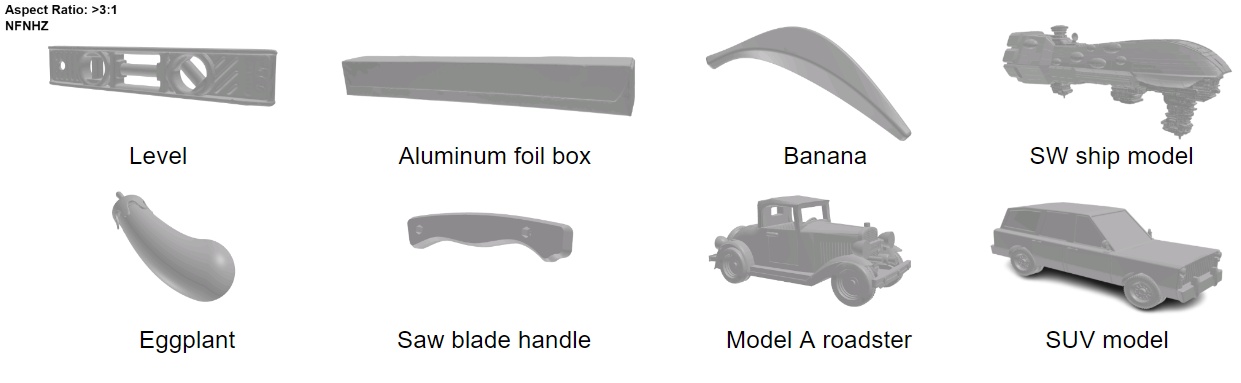}
    \caption{Objects from bin 6 which have the aspect ratio >3:1 and are NFNHZ}
    \label{fig:rendergroup6}
\end{figure}

\begin{figure}
    \centering
    \includegraphics[width=\linewidth]{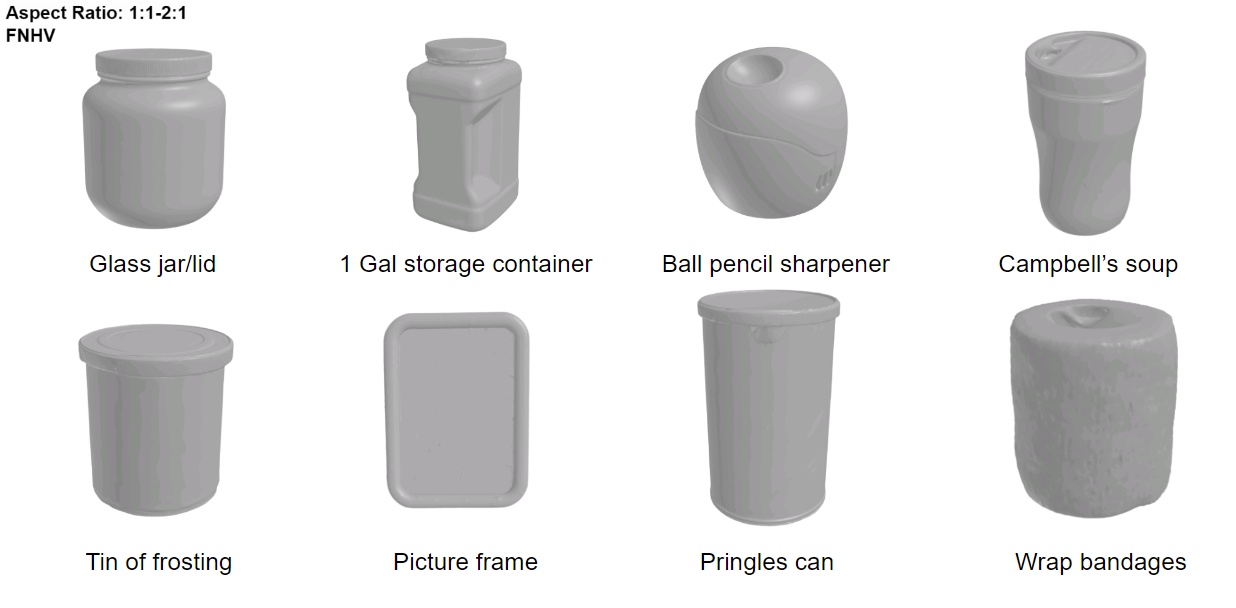}
    \caption{Objects from bin 7 which have the aspect ratio 1:1-2:1 and are FNHV}
    \label{fig:rendergroup7}
\end{figure}

\begin{figure}
    \centering
    \includegraphics[width=\linewidth]{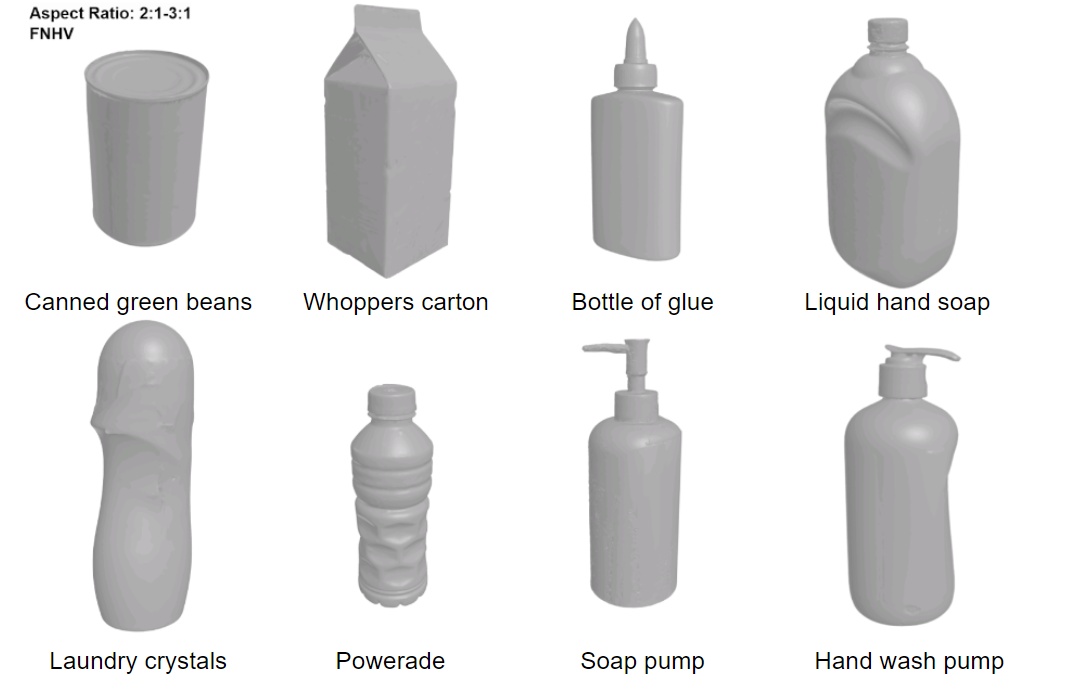}
    \caption{Objects from bin 8 which have the aspect ratio 2:1-3:1 and are FNHV}
    \label{fig:rendergroup8}
\end{figure}

\begin{figure}
    \centering
    \includegraphics[width=\linewidth]{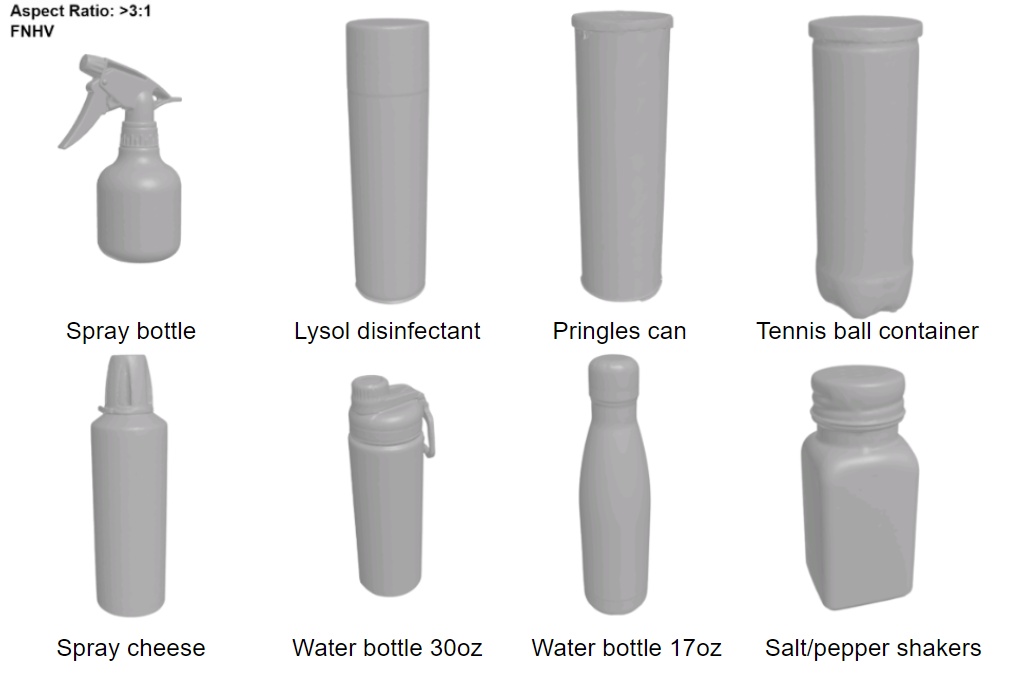}
    \caption{Objects from bin 9 which have the aspect ratio >3:1 and are FNHV}
    \label{fig:rendergroup9}
\end{figure}

\begin{figure}
    \centering
    \includegraphics[width=\linewidth]{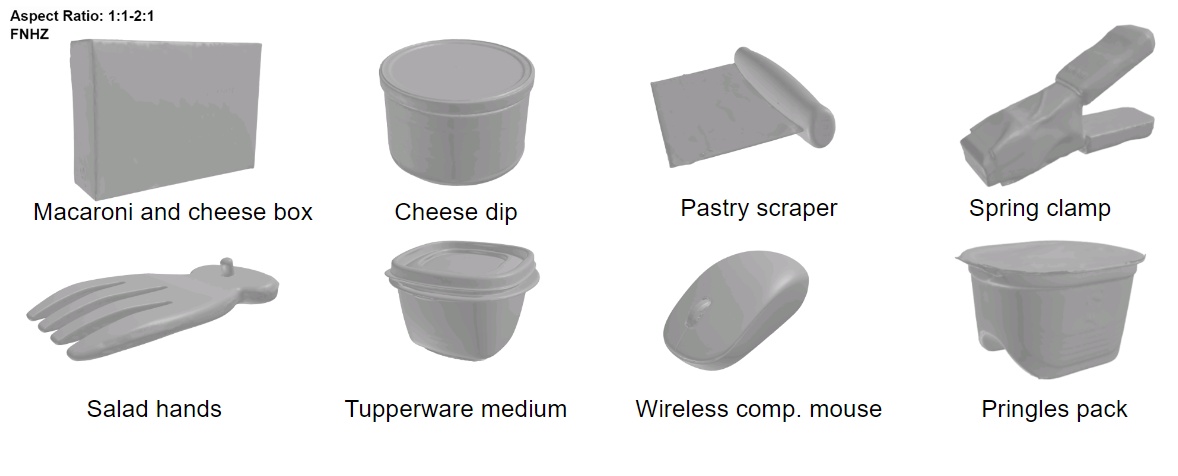}
    \caption{Objects from bin 10 which have the aspect ratio 1:1-2:1 and are FNHZ}
    \label{fig:rendergroup10}
\end{figure}

\begin{figure}
    \centering
    \includegraphics[width=\linewidth]{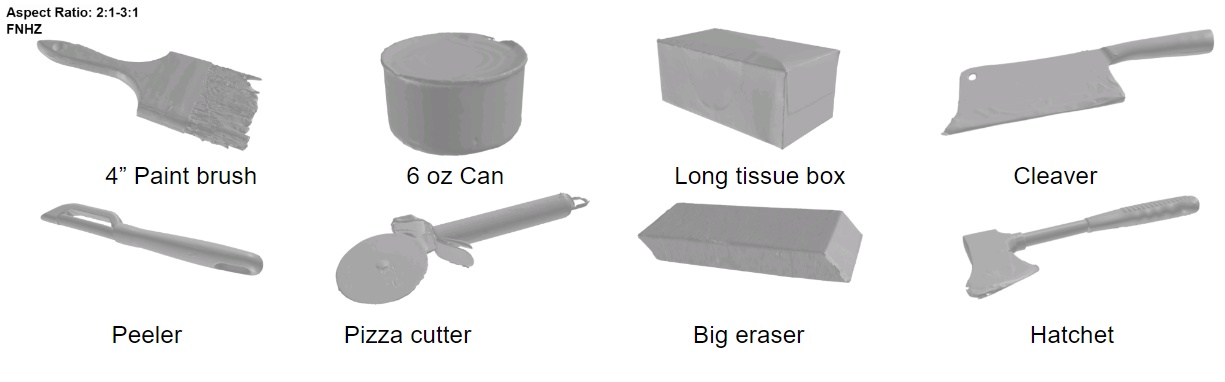}
    \caption{Objects from bin 11 which have the aspect ratio 2:1-3:1 and are FNHZ}
    \label{fig:rendergroup11}
\end{figure}

\begin{figure}
    \centering
    \includegraphics[width=\linewidth]{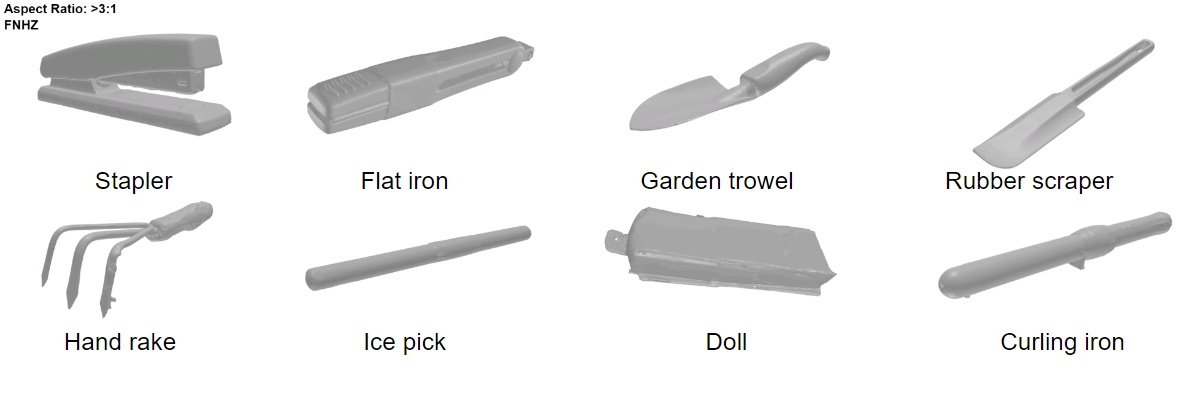}
    \caption{Objects from bin 12 which have the aspect ratio >3:1 and are FNHZ}
    \label{fig:rendergroup12}
\end{figure}

\begin{figure}
    \centering
    \includegraphics[width=\linewidth]{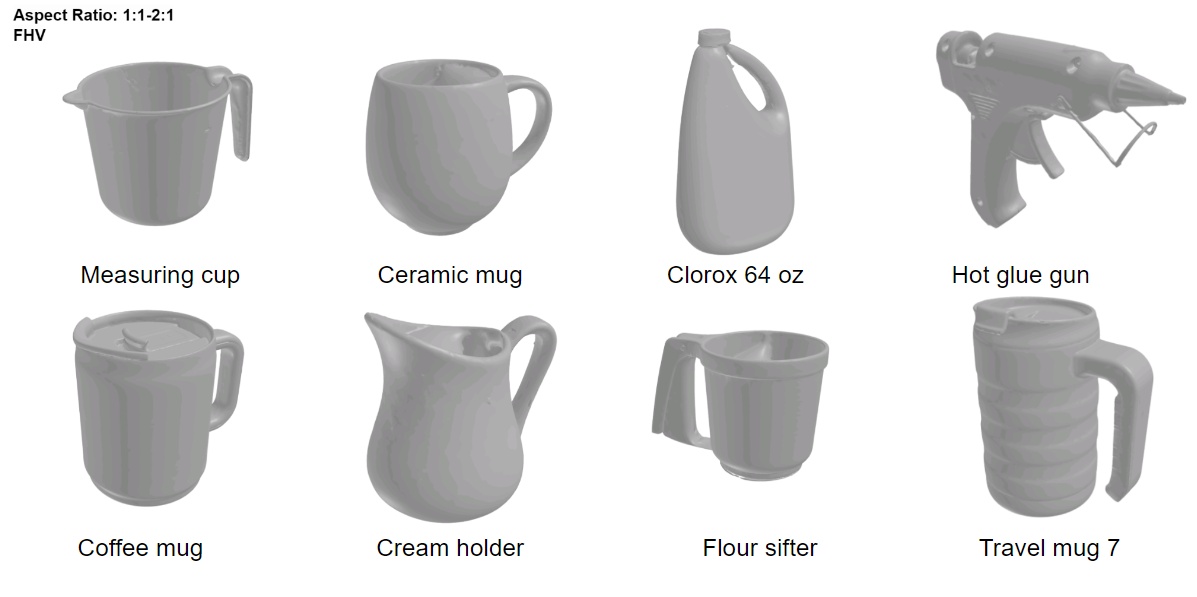}
    \caption{Objects from bin 13 which have the aspect ratio 1:1-2:1 and are FHV}
    \label{fig:rendergroup13}
\end{figure}

\begin{figure}
    \centering
    \includegraphics[width=\linewidth]{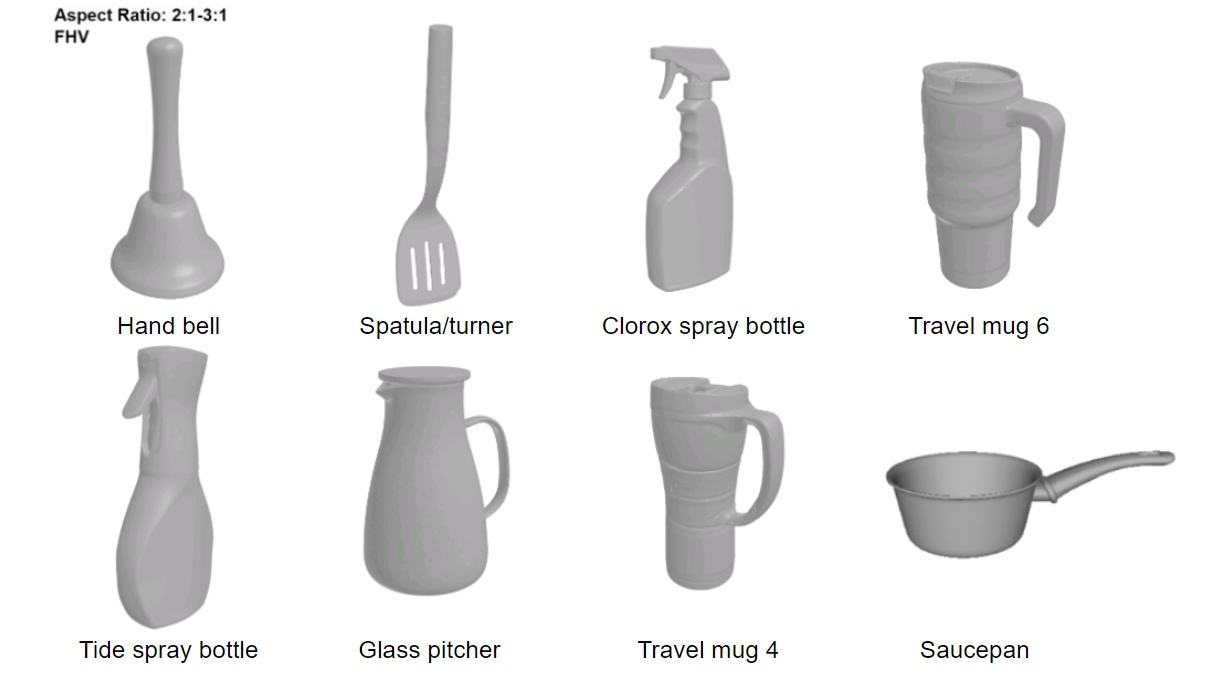}
    \caption{Objects from bin 14 which have the aspect ratio 2:1-3:1 and are FHV}
    \label{fig:rendergroup14}
\end{figure}

\begin{figure}
    \centering
    \includegraphics[width=\linewidth]{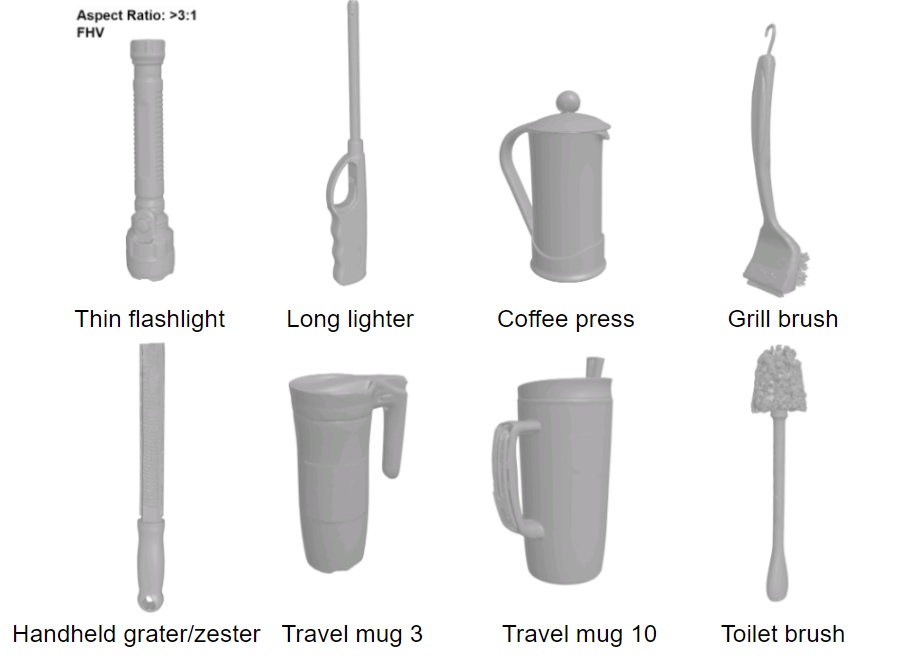}
    \caption{Objects from bin 15 which have the aspect ratio >3:1 and are FHV}
    \label{fig:rendergroup15}
\end{figure}

\begin{figure}
    \centering
    \includegraphics[width=\linewidth]{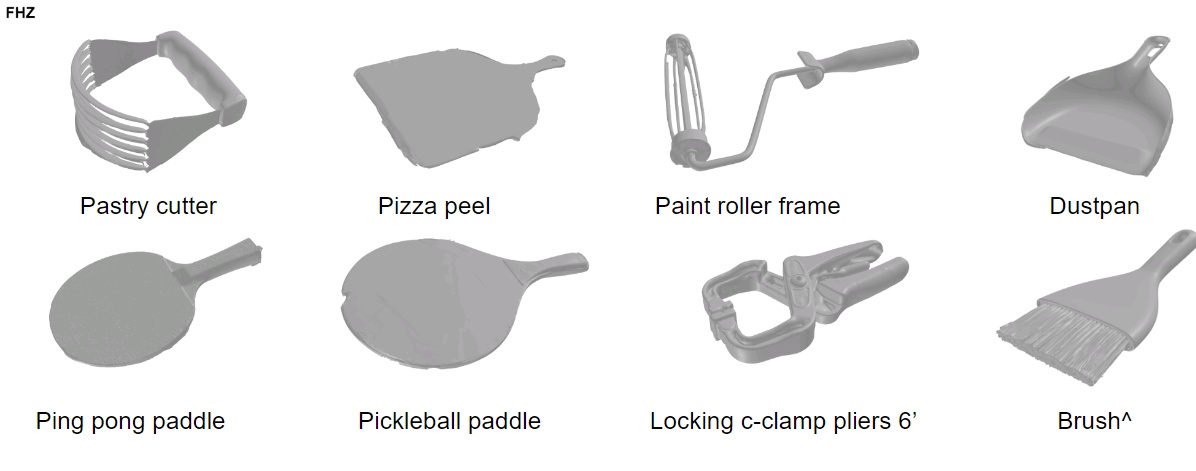}
    \caption{Objects from bin 16 which are FHZ}
    \label{fig:rendergroup16}
\end{figure}

\begin{figure}
    \centering
    \includegraphics[width=\linewidth]{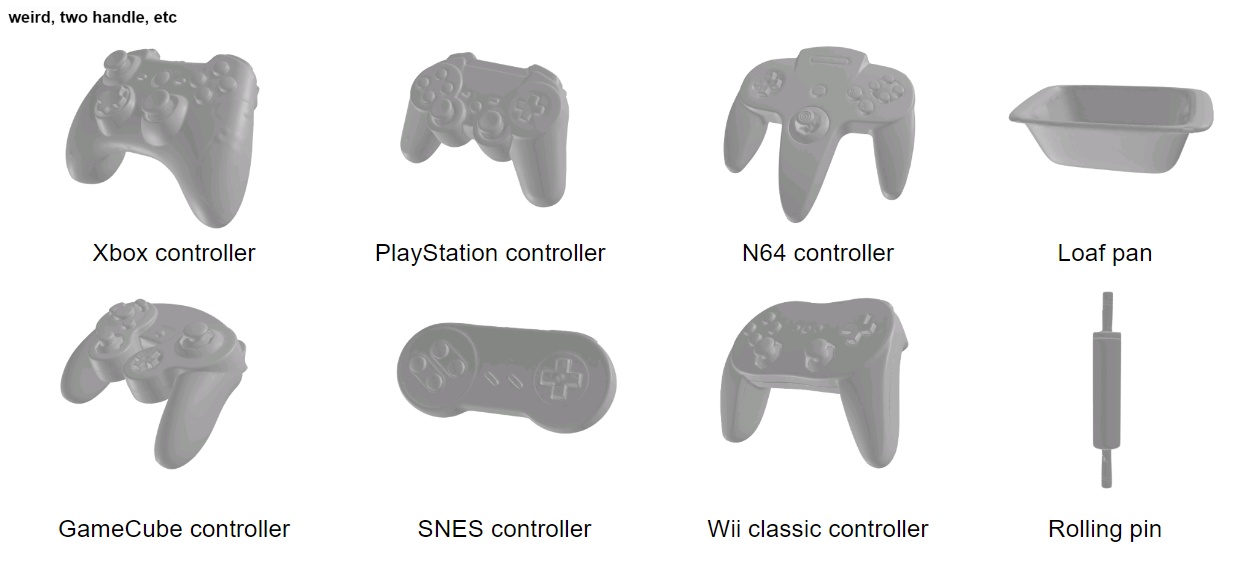}
    \caption{Objects from bin 17 which are classified as "other"}
    \label{fig:rendergroup17}
\end{figure}

We provide metadata information for all 136 objects used in our work in Table~\ref{tab:object_metadata}. Metadata for objects 116 and 120 is from the original mesh posted on Thingiverse as shown in Table~\ref{tab:thingiverse_attribution}. Figures~\ref{fig:rendergroup1} to \ref{fig:rendergroup17} show renders of the 3D models of the 136 objects, categorized in the 17 aspect-ratio and functionality categories in our work.

\setlength{\tabcolsep}{2pt}
{\small\begin{longtable}{|l|l|l|l|l|l|l|}
\caption{Metadata for each object mesh used in HOH.}
\label{tab:object_metadata}\\
\hline
ID & Description & Mass (g) & Vertices  & Faces  & Type  & Aspect Ratio
\endfirsthead
\endhead
\hline
\hline
100  & Rubik's Cube small                   & 75         & 1,449,255 & 2,506,594 & NFNHV        & 1:1-2:1           \\
102  & Tennis ball                          & 67         & 1,249,999 & 2,499,994 & NFNHV        & 1:1-2:1           \\
106  & 1" PVC Tee                           & 67         & 1,235,598 & 2,471,414 & NFNHV        & 1:1-2:1           \\
107  & 1" PVC 90 degree elbow               & 53         & 1,249,999 & 2,499,998 & NFNHV        & 1:1-2:1           \\
109  & 1" PVC Coupling                      & 31         & 1,249,999 & 2,499,998 & NFNHV        & 1:1-2:1           \\
110  & 2" PVC Coupling                      & 59         & 1,249,995 & 2,500,014 & NFNHV        & 1:1-2:1           \\
114  & Lint roller refill                   & 80         & 1,241,735 & 2,483,672 & NFNHV        & 2:1-3:1           \\
123  & Joystick                             & 231        & 1,250,007 & 2,501,696 & NFNHV        & 2:1-3:1           \\
T001 & Doll                                 & 165        & 854,375   & 1,709,430 & FNHZ              & \textgreater{}3:1 \\
132  & Lady statue                          & 176        & 741,212   & 1,482,424 & NFNHV        & \textgreater{}3:1 \\
134  & Power strip tower                    & 591        & 374,362   & 748,792   & NFNHV        & \textgreater{}3:1 \\
135  & Incense holder                       & 277        & 1,234,100 & 2,468,686 & NFNHV        & \textgreater{}3:1 \\
200  & Playing Card Deck 4pk                & 103        & 1,248,897 & 2,497,828 & NFNHZ        & 1:1-2:1           \\
201  & Lifesaver Candy Box                  & 242        & 1,228,021 & 2,456,092 & NFNHZ        & 1:1-2:1           \\
202  & Ramekin                              & 268        & 1,249,999 & 2,499,994 & NFNHZ        & 1:1-2:1           \\
203  & Cookie pan                           & 973        & 2,034,211 & 4,732,076 & NFNHZ        & 1:1-2:1           \\
204  & Cutting board                        & 762        & 1,249,999 & 2,499,998 & NFNHZ        & 1:1-2:1           \\
206  & Comp. notebook                       & 322        & 1,282,894 & 2,715,282 & NFNHZ        & 1:1-2:1           \\
207  & Spiral notebook                      & 49         & 1,249,818 & 2,502,330 & NFNHZ        & 1:1-2:1           \\
211  & Gift box                             & 103        & 1,249,995 & 2,500,030 & NFNHZ        & 1:1-2:1           \\
213  & Butter dish                          & 216        & 1,250,000 & 2,499,996 & NFNHZ        & 2:1-3:1           \\
216  & Microwave omelet cooker              & 114        & 1,199,080 & 2,398,176 & NFNHZ        & 2:1-3:1           \\
217  & Qtip box (2pk)                       & 296        & 1,186,889 & 2,373,786 & NFNHZ        & 2:1-3:1           \\
220  & Travel palette                       & 230        & 1,043,628 & 2,087,264 & NFNHZ        & 2:1-3:1           \\
222  & iPhone Case                          & 29         & 1,095,691 & 2,191,526 & NFNHZ        & 2:1-3:1           \\
223  & iPad case                            & 305        & 1,246,934 & 2,521,020 & NFNHZ        & 2:1-3:1           \\
224  & Level                                & 52         & 580,234   & 1,160,500 & NFNHZ        & \textgreater{}3:1 \\
225  & Aluminum foil box                    & 181        & 926,667   & 1,853,330 & NFNHZ        & \textgreater{}3:1 \\
301  & Glass jar/lid                        & 738        & 1,249,999 & 2,499,994 & FNHV              & 1:1-2:1           \\
303  & 1Gal storage container               & 173        & 1,249,990 & 2,500,008 & FNHV              & 1:1-2:1           \\
304  & Ball pencil sharpener                & 154        & 249,156   & 498,308   & FNHV              & 1:1-2:1           \\
305  & Tin of frosting                      & 386        & 1,249,999 & 2,499,998 & FNHV              & 1:1-2:1           \\
308  & Picture frame 11x14                  & 764        & 1,250,013 & 2,500,070 & FNHV              & 1:1-2:1           \\
309  & Pringles can (short) (12pk)          & 104        & 1,249,964 & 2,500,044 & FNHV              & 1:1-2:1           \\
310  & Campbell's soup 10.75oz (8pk)        & 351        & 1,209,073 & 2,418,368 & FNHV              & 1:1-2:1           \\
311  & Wrap bandages 2" 6pk                 & 19         & 1,249,997 & 2,500,006 & FNHV              & 1:1-2:1           \\
314  & Canned Green Beans                   & 473        & 1,249,882 & 2,501,408 & FNHV              & 2:1-3:1           \\
315  & Whoppers carton                      & 382        & 1,249,995 & 2,499,998 & FNHV              & 2:1-3:1           \\
316  & Bottle of glue                       & 179        & 848,235   & 1,696,478 & FNHV              & 2:1-3:1           \\
317  & Laundry crystals                     & 513        & 936,835   & 1,873,678 & FNHV              & 2:1-3:1           \\
318  & Powerade                             & 632        & 1,068,025 & 2,136,090 & FNHV              & 2:1-3:1           \\
320  & Soap pump                            & 102        & 1,061,839 & 2,153,150 & FNHV              & 2:1-3:1           \\
322  & Liquid hand soap 50oz                & 2044       & 1,250,029 & 2,500,164 & FNHV              & 2:1-3:1           \\
323  & Hand wash pump                       & 415        & 1,249,999 & 2,499,998 & FNHV              & 2:1-3:1           \\
326  & Spray bottle                         & 54         & 1,095,518 & 2,191,040 & FNHV              & \textgreater{}3:1 \\
328  & Lysol disinfectant                   & 685        & 922,156   & 1,844,316 & FNHV              & \textgreater{}3:1 \\
329  & Pringles can                         & 198        & 1,127,367 & 2,254,946 & FNHV              & \textgreater{}3:1 \\
330  & Spray cheese                         & 293        & 1,106,393 & 2,213,170 & FNHV              & \textgreater{}3:1 \\
332  & Water bottle 20oz                    & 334        & 1,249,352 & 2,499,096 & FNHV              & \textgreater{}3:1 \\
333  & Water bottle 17oz                    & 300        & 734,182   & 1,468,372 & FNHV              & \textgreater{}3:1 \\
334  & Tennis ball container                & 47         & 1,249,692 & 2,500,012 & FNHV              & \textgreater{}3:1 \\
335  & Salt/pepper shakers                  & 107        & 1,195,377 & 2,390,772 & FNHV              & \textgreater{}3:1 \\
F001 & Macaroni \& Cheese Box (3pk)         & 369        & 1,249,997 & 2,499,998 & FNHZ              & 1:1-2:1           \\
402  & Cheese dip                           & 282        & 1,237,936 & 2,476,016 & FNHZ              & 1:1-2:1           \\
403  & Pastry scraper                       & 204        & 150,009   & 2,300,020 & FNHZ              & 1:1-2:1           \\
404  & Salad hands                          & 107        & 475,852   & 951,728   & FNHZ              & 1:1-2:1           \\
406  & Tupperware medium                    & 65         & 1,249,982 & 2,500,098 & FNHZ              & 1:1-2:1           \\
408  & Wireless comp. mouse                 & 51         & 1,250,000 & 2,499,996 & FNHZ              & 1:1-2:1           \\
410  & Spring clamp                         & 67         & 1,102,878 & 2,205,778 & FNHZ              & 1:1-2:1           \\
411  & Pringles pack (small) 18pk           & 29         & 1,237,226 & 2,474,582 & FNHZ              & 1:1-2:1           \\
412  & 4" paint brush                       & 83         & 579,790   & 1,161,096 & FNHZ              & 2:1-3:1           \\
415  & 6oz can                              & 180        & 1,240,341 & 2,480,686 & FNHZ              & 2:1-3:1           \\
416  & Long tissue box                      & 299        & 1,249,999 & 2,499,998 & FNHZ              & 2:1-3:1           \\
417  & Peeler                               & 37         & 358,598   & 720,790   & FNHZ              & 2:1-3:1           \\
418  & Pizza cutter                         & 55         & 793,896   & 1,588,056 & FNHZ              & 2:1-3:1           \\
419  & Garden trowel                        & 187        & 390,889   & 781,782   & FNHZ              & 2:1-3:1           \\
421  & Cleaver                              & 399        & 380,195   & 760,131   & FNHZ              & 2:1-3:1           \\
423  & Hatchet                              & 950        & 636,305   & 1,272,694 & FNHZ              & 2:1-3:1           \\
424  & Stapler                              & 150        & 967,171   & 1,934,418 & FNHZ              & \textgreater{}3:1 \\
426  & Flat iron                            & 161        & 481,708   & 963,424   & FNHZ              & \textgreater{}3:1 \\
O010 & Big Eraser                           & 214        & 1,237,999 & 2,476,030 & FNHZ              & \textgreater{}3:1 \\
430  & Hand rake                            & 228        & 418,414   & 836,844   & FNHZ              & \textgreater{}3:1 \\
432  & Ice pick                             & 93         & 324,325   & 648,680   & FNHZ              & \textgreater{}3:1 \\
C003 & Ice Cube Tray                        & 103        & 1,241,945 & 2,484,010 & NFNHZ        & 2:1-3:1           \\
434  & Rubber scraper                       & 39         & 293,714   & 587,428   & FNHZ              & \textgreater{}3:1 \\
435  & Curling iron                         & 196        & 370,451   & 740,898   & FNHZ              & \textgreater{}3:1 \\
500  & Measuring cup                        & 79         & 1,249,995 & 2,500,006 & FHV                   & 1:1-2:1           \\
502  & Ceramic mug (2pk)                    & 358        & 1,249,999 & 2,499,998 & FHV                   & 1:1-2:1           \\
503  & Clorox 64oz                          & 2069       & 1,249,994 & 2,500,000 & FHV                   & 1:1-2:1           \\
505  & Coffee mug (small, handle)           & 233        & 1,249,994 & 2,500,008 & FHV                   & 1:1-2:1           \\
507  & Cream holder                         & 535        & 1,250,016 & 2,500,092 & FHV                   & 1:1-2:1           \\
508  & Flour sifter                         & 139        & 1,229,875 & 2,459,966 & FHV                   & 1:1-2:1           \\
509  & Hot glue gun                         & 86         & 1,008,984 & 2,017,984 & FHV                   & 1:1-2:1           \\
511  & Travel mug 7 (handle)                & 378        & 1,158,350 & 2,316,908 & FHV                   & 1:1-2:1           \\
513  & Saucepan 2                           & 617        & 1,118,018 & 2,236,046 & FHV                   & 2:1-3:1           \\
514  & Hand bell                            & 74         & 769,846   & 1,540,082 & FHV                   & 2:1-3:1           \\
516  & Spatula/turner                       & 146        & 316,995   & 610,792   & FHV                   & 2:1-3:1           \\
517  & Clorox spray bottle                  & 739        & 978,053   & 1,956,130 & FHV                   & 2:1-3:1           \\
518  & Tide spray bottle                    & 500        & 1,034,111 & 2,068,234 & FHV                   & 2:1-3:1           \\
519  & Glass pitcher 60oz                   & 614        & 1,249,999 & 2,499,998 & FHV                   & 2:1-3:1           \\
520  & Travel mug 4 (handle)                & 425        & 1,282,759 & 2,500,002 & FHV                   & 2:1-3:1           \\
522  & Travel mug 6 (handle)                & 449        & 1,280,815 & 2,499,998 & FHV                   & 2:1-3:1           \\
525  & Thin flash light                     & 55         & 514,793   & 1,029,594 & FHV                   & \textgreater{}3:1 \\
526  & Long lighter (6pk)                   & 35         & 199,958   & 399,916   & FHV                   & \textgreater{}3:1 \\
527  & Coffee press                         & 234        & 1,242,822 & 2,485,672 & FHV                   & \textgreater{}3:1 \\
528  & Handheld grater/zester               & 93         & 871,379   & 1,846,140 & FHV                   & \textgreater{}3:1 \\
530  & Travel mug 3 (handle)                & 392        & 1,249,994 & 2,500,000 & FHV                   & \textgreater{}3:1 \\
531  & Travel mug 10 (handle)               & 374        & 1,249,915 & 2,500,098 & FHV                   & \textgreater{}3:1 \\
533  & Toilet brush                         & 166        & 515,317   & 1,031,586 & FHV                   & \textgreater{}3:1 \\
534  & Grill brush                          & 313        & 743,919   & 1,488,358 & FHV                   & \textgreater{}3:1 \\
601  & Pastry cutter                        & 168        & 1,232,141 & 2,465,350 & FHZ                   & 1:1-2:1           \\
602  & Pizza peel                           & 375        & 993,058   & 1,987,112 & FHZ                   & 1:1-2:1           \\
603  & Paint roller frame                   & 212        & 552,552   & 1,105,252 & FHZ                   & 1:1-2:1           \\
605  & Ping pong paddle (2)                 & 132        & 785,606   & 1,571,540 & FHZ                   & 1:1-2:1           \\
606  & Pickleball paddle (2)                & 242        & 1,258,015 & 2,525,668 & FHZ                   & 1:1-2:1           \\
607  & Locking c-clamp pliers 6"            & 51         & 906,725   & 1,813,462 & FHZ                   & 1:1-2:1           \\
610  & Dustpan                              & 120        & 1,246,308 & 2,495,334 & FHZ                   & 1:1-2:1           \\
611  & Brush \textasciicircum{}             & 180        & 1,240,321 & 2,480,836 & FHZ                   & 1:1-2:1           \\
701  & Xbox controller                      & 206        & 1,250,000 & 2,499,996 & other & 1:1-2:1           \\
702  & Playstation controller               & 229        & 1,235,324 & 2,470,656 & other & 1:1-2:1           \\
703  & N64 controller                       & 169        & 1,249,999 & 2,500,018 & other & 1:1-2:1           \\
704  & Gamecube controller                  & 156        & 1,248,340 & 2,496,698 & other & 1:1-2:1           \\
706  & SNES controller (2pk)                & 67         & 935,443   & 1,870,902 & other & 1:1-2:1           \\
708  & Wii classic controller (2pk)         & 114        & 1,250,031 & 2,500,106 & other & 1:1-2:1           \\
713  & Loaf pan                             & 916        & 1,250,111 & 2,500,444 & other & 1:1-2:1           \\
714  & Rolling pin                          & 573        & 761,806   & 1,525,708 & other & 1:1-2:1           \\
104  & Apple                                & 88         & 5,644     & 11,285    & NFNHV        & 1:1-2:1           \\
105  & Bell pepper                          & 87         & 21,094    & 42,184    & NFNHV        & 1:1-2:1           \\
115  & Santa                                & 41         & 246,876   & 493,744   & NFNHV        & 2:1-3:1           \\
116  & Deco vase                            & 146        & 103,896   & 207,788   & NFNHV        & 2:1-3:1           \\
118  & Column pot                           & 54         & 574       & 1,144     & NFNHV        & 2:1-3:1           \\
120  & Baby yoda statue                     & 56         & 778,292   & 1,556,670 & NFNHV        & 2:1-3:1           \\
121  & Short spiral ornament                & 21         & 44,752    & 89,500    & NFNHV        & 2:1-3:1           \\
122  & Candle lantern                       & 112        & 86,431    & 173,082   & NFNHV        & 2:1-3:1           \\
124  & Pineapple                            & 74         & 8,108     & 16,278    & NFNHV        & \textgreater{}3:1 \\
126  & Zucchini                             & 42         & 340,012   & 680,020   & NFNHV        & \textgreater{}3:1 \\
127  & Twist vase                           & 35         & 1,077,060 & 2,514,160 & NFNHV        & \textgreater{}3:1 \\
128  & Nefertiti head                       & 90         & 1,249,999 & 2,499,998 & NFNHV        & \textgreater{}3:1 \\
129  & Statue of liberty                    & 64         & 101,710   & 203,471   & NFNHV        & \textgreater{}3:1 \\
221  & Balancing bird                       & 26         & 317,622   & 635,480   & NFNHZ        & 2:1-3:1           \\
228  & Banana                               & 53         & 5,625     & 11,246    & NFNHZ        & \textgreater{}3:1 \\
232  & Eggplant                             & 192        & 76,634    & 153,260   & NFNHZ        & \textgreater{}3:1 \\
233  & Saw blade handle                     & 19         & 2,505     & 5,014     & NFNHZ        & \textgreater{}3:1 \\
234  & Model A Roadster                     & 51         & 58,191    & 118,146   & NFNHZ        & \textgreater{}3:1 \\
235  & Star wars ship model                 & 42         & 95,825    & 191,704   & NFNHZ        & \textgreater{}3:1 \\
236  & SUV model                            & 37         & 26,012    & 52,036    & NFNHZ        & \textgreater{}3:1\\\hline

\end{longtable}}


{\small
\bibliographystyle{plain}
\bibliography{supp}







}